\documentclass[journal]{IEEEtran} 
\ifCLASSOPTIONcompsoc
  \usepackage[nocompress]{cite}
\else
  \usepackage{cite}
\fi

\ifCLASSINFOpdf
\else
\fi

\usepackage[normalem]{ulem}
\usepackage{bm}
\usepackage{amsmath}
\usepackage{amssymb}
\usepackage{balance}
\usepackage{amsmath}
\usepackage{diagbox}
\usepackage{adjustbox}
\usepackage{subfigure}
\usepackage{makecell}
\usepackage{mathrsfs}
\usepackage{xcolor}

\usepackage{float}
\usepackage{pifont}
\newcommand{\xmark}{\ding{55}}
\usepackage[T1]{fontenc}
\usepackage{stfloats}

\usepackage[pagebackref=true,breaklinks=true,letterpaper=true,colorlinks,bookmarks=false]{hyperref}

\DeclareMathOperator*{\argmin}{arg\,min}
\begin{document}
\title{Deep Magnification-Flexible Upsampling over \\3D Point Clouds}

\author{Yue Qian, Junhui Hou, \textit{Senior Member}, \textit{IEEE},  Sam Kwong, \textit{Fellow}, \textit{IEEE}, and Ying He, \textit{Member}, \textit{IEEE}
    \thanks{This work was supported in part by the Natural Science Foundation of China under Grants 61871342, and in part by the Hong Kong Research Grants Council under Grant 9042955 (CityU 11202320). (\textit{Corresponding Author: Junhui Hou})} 
    \thanks{Y. Qian, J. Hou, and S. Kwong are with the Department of Computer Science, City University of Hong Kong, Hong Kong. Email: yueqian4-c@my.cityu.edu.hk; jh.hou@cityu.edu.hk;cssamk@cityu.edu.hk}
    \thanks{Y. He is with the School of Computer Science and Engineering, Nanyang Technological University, Singapore, 639798. Email: yhe@ntu.edu.sg}}

\IEEEtitleabstractindextext{%
\begin{abstract}
This paper addresses the problem of generating dense point clouds from given sparse point clouds to model the underlying geometric structures of objects/scenes. 
To tackle this challenging issue, we propose a novel end-to-end learning-based framework.
Specifically, by taking advantage of the linear approximation theorem, we first formulate the problem explicitly, which boils down to determining the interpolation weights and high-order approximation errors. Then, we design a \textit{lightweight} neural network to adaptively learn unified and sorted interpolation weights as well as the high-order refinements, by analyzing the local geometry of the input point cloud.  
The proposed method can be interpreted by the explicit formulation, and thus is more memory-efficient than existing ones. In sharp contrast to the existing methods that work only for a pre-defined and fixed upsampling factor, the proposed framework only requires a single neural network with one-time training to handle various upsampling factors within a typical range, which is highly desired in real-world applications. In addition, we propose a simple yet effective training strategy to drive such a flexible ability. In addition, our method can handle non-uniformly distributed and noisy data well. 
Extensive experiments on both synthetic and real-world data demonstrate the superiority of the proposed method over state-of-the-art methods both quantitatively and qualitatively. The code will be publicly available at \url{https://github.com/ninaqy/Flexible-PU}.

\end{abstract}

}

\maketitle

\IEEEdisplaynontitleabstractindextext

\IEEEpeerreviewmaketitle

\section{Introduction}
\label{sec:introduction}

\begin{figure}[t]
\centering
\includegraphics[width=0.5\textwidth]{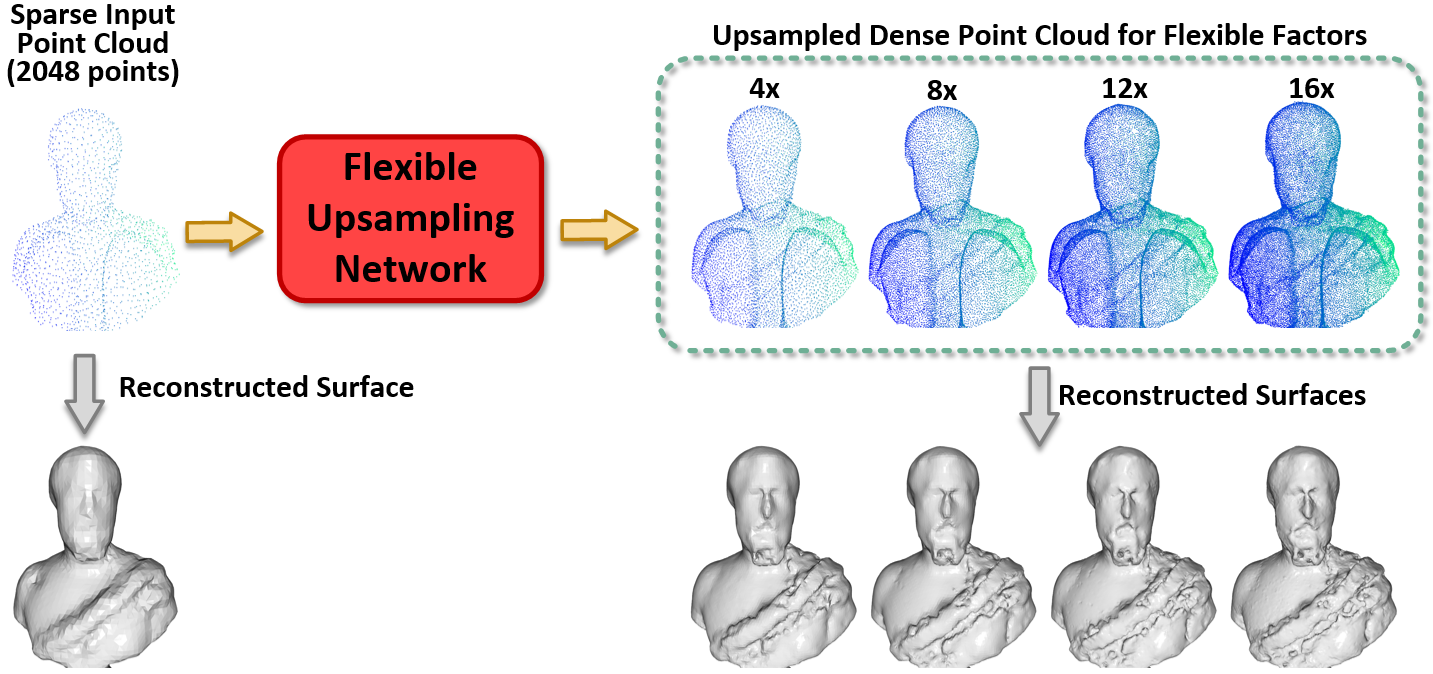}
\caption{The proposed method can upsample a sparse point cloud with a flexible factor not greater than the maximum upsampling factor after one-time training. In this example, 
the maximum upsampling factor is equal to $16$.
Here we only show upsampling results for factors $4\times$, $8\times$, $12\times$ and $16\times$, and observe the proposed method can generate meaningful geometric details. Moreover, the generated details are richer and closer to the ground truth ones with the factor increasing. }
\label{fig:upsample_all}
\end{figure}

\IEEEPARstart{O}{wing} to the flexibility and efficiency in representing objects/scenes of complex geometry and topology, 
point clouds are widely used  
in immersive telepresence \cite{orts2016holoportation}, 
3D city reconstruction \cite{lafarge2012creating}, \cite{musialski2013survey}, cultural heritage reconstruction \cite{xu2014tridimensional}, \cite{bolognesi2015testing}, geophysical information systems \cite{paine2016shoreline}, \cite{nie2016estimating}, autonomous driving \cite{chen2017multi}, \cite{li20173d},  
and  virtual/augmented reality \cite{held20123d}, \cite{santana2017multimodal}.
Despite of great progress of 3D sensing technology \cite{hakala2012full}, \cite{kimoto2014development} in recent years, it is still costly and time-consuming to acquire dense point clouds for representing shapes with rich geometric details, which are highly desired in downstream applications. 
Instead of relying on hardware improvement, we are interested in developing a computational method that is able to upsample a given sparse, low-resolution point cloud to a dense one that faithfully represents the underlying surface (see Fig.~\ref{fig:upsample_all}). Since the upsampling problem is often thought as 
a 3D counterpart of image super-resolution \cite{lai2017deep}, \cite{zhang2018residual}, intuitively one may consider borrowing powerful techniques from image processing community. However, due to the unordered and irregular nature of point clouds, such an extension is far from trivial, especially when the underlying surface has complex geometry and/or topology. Besides,  the two types of data are essentially different i.e., 3D point clouds represent explicit geometry information of objects/scenes, while 2D images only record the
reflected light intensities (i.e., color) by objects/scenes, which hinders the straightforward extension of well-developed image super-resolution techniques to some extent. 

There are roughly two categories of methods for point cloud upsampling: optimization-based methods \cite{alexa2003computing}, \cite{lipman2007parameterization}, \cite{huang2009consolidation}, \cite{preiner2014continuous}, \cite{huang2013edge},\cite{wu2015deep} and deep learning-based methods \cite{yu2018pu}, \cite{yu2018ec}, \cite{yifan2019patch},  \cite{li2019pu}, \cite{qian2020pugeo}. The former usually fits local geometry and works well for smooth surfaces with less features. However, these methods struggle with multi-scale structure preservation. The latter adopts trained neural networks to 
adaptively learn structures from data, and outperforms optimization-based methods to a significant extent. 
However, the existing deep learning-based methods 
either take little consideration of the geometric properties of 3D point clouds 
or local neighborhood information, which limits their performance.

Upsampling raw point clouds with various upsampling factors is common in point cloud processing. For example, the input point clouds captured by different sensors may have different resolutions. Therefore, the user may have to upsample each of them with different factors to obtain the desired resolution. 
The user may also determine the upsampling factor based on resource constraints, such as display, computing power, and transmission bandwidth. 
Besides,  
the desired point cloud resolution varies with application scenarios. For example, a high-resolution point cloud is highly expected for surface reconstruction, while a moderately sparse one is tolerated for object detection. However, existing methods are designed for a \textit{fixed} upsampling factor. To handle upsampling with varying factors, one has to build multiple networks and train each of them with a pre-defined factor, which increases both the model complexity and the training time significantly.
Thus, a single network that is trained only once and can support flexible upsampling factors is highly desired in real-world applications. 

In this paper, we propose a novel end-to-end learning-based magnification-flexible upsampling method for 3D point clouds, 
which is capable of upsampling an input point cloud with flexible factors 
after one-time training. 
Motivated by the fact that tangent plane is the best local linear approximation of a curved surface, we generate a new sample by an affine combination of neighboring points projected onto the tangent plane. 
Technically, given a query point, 
the proposed framework first creates new points in its local neighborhood as the affine combination of its neighboring points, where unified and sorted interpolation weights are adaptively learned by analyzing the local geometry structure, instead of being predefined. These newly generated points are distributed in the convex hull of the neighbouring points. 
Then, the coordinates of the coarse points are further refined to approach the underlying surface via a self-attention-based refinement module. Different from the existing methods, the proposed method upsamples point clouds with the local neighborhood information explicitly involved in an interpretable manner.
Besides, we propose a simple yet effective training strategy to drive the learning of the flexibility of our network. 
Through extensive experiments and evaluations on both synthetic and real-world data, we demonstrate that the proposed framework can consistently outperform state-of-the-art methods for upsampling factors from $4\times$ to $16\times$ in terms of commonly-used quantitative metrics. Qualitatively, 3D meshes reconstructed from the densified points of the proposed method contain richer geometric details than those of state-of-the-art approaches.  
More importantly, owing to our unique and explicit formulation towards the 3D point cloud upsampling problem, the proposed method is much more memory-efficient and more interpretable than existing methods.

The rest of this paper is organized as follows. Section~\ref{sec:related work} reviews existing methods  on point cloud upsampling. Section~\ref{sec:formulation} formulates the point cloud upsampling problem in an explicit manner by using the linear approximation theorem. Section~\ref{sec:proposed method} presents the proposed framework, followed by experimental results and comparisons in Section~\ref{sec:exp}. Finally, Section~\ref{sec:con} concludes the paper.

\section{Related Work}
\label{sec:related work}
\subsection{Optimization-based Methods} 
Over the past decade, a number of optimization-based methods for point cloud upsampling/resampling have been proposed. For example, Alexa \textit{et al.} \cite{alexa2003computing} upsampled points 
by referring the Voronoi diagram, which requires the surface smoothness assumption and computes on the moving least squares surface.
Based on a locally optimal projection operator (LOP), Lipman \textit{et al.}  \cite{lipman2007parameterization} developed a parametrization-free method for point resampling and surface reconstruction. Subsequently, Huang \textit{et al.} \cite{huang2009consolidation} and  Preiner \textit{et al.} \cite{preiner2014continuous} proposed weighted LOP and continuous LOP, respectively. Specifically, the weighted LOP iteratively consolidates point clouds by means of normal estimation, and thus is robust to noise and outliers. The continuous LOP can perform fast surface reconstruction by adopting a Gaussian mixture model. However, LOP-based methods assume that points are sampled from smooth surfaces, which degrades upsampling quality towards sharp edges and corners. To effectively preserve the sharp features, Huang \textit{et al.} \cite{huang2013edge} presented an edge-aware (EAR) approach, which first resamples points away from edges with reference to given normal information, then progressively upsamples points to approach the edge singularities. However, the performance of EAR heavily depends on the given normal information and parameter tuning. By introducing the concept of deep point, Wu \textit{et al.}~\cite{wu2015deep} proposed a method to jointly perform point cloud completion and consolidation under the guidance of extracted Meso-skeletons. 
The method can successfully recover regions with holes; however, it is sensitive to noise and outliers. Dinesh \textit{et al}.~\cite{dinesh20193d} proposed a graph signal optimization method, 
which minimizes the total variation of estimated normals by partitioning the point clouds into two disjoint sets and optimizes the corresponding coordinates by the alternating method of multipliers iteratively.

\subsection{Deep Learning-based Methods} 
The great success of deep learning in image/video processing and analysis encourages both academia and industrial to explore the potential of deep learning on 3D point cloud processing and analysis. However, the unordered and irregular characteristics of point clouds make it non-trivial.   
Qi \textit{et al}. \cite{qi2017pointnet} pioneered PointNet, the first deep learning-based platform that can directly process the raw 3D point cloud data. The shared multi-layer perceptron (MLP) per point and the symmetric max-pooling operation help PointNet to cope with the irregular and unordered characteristics of point clouds. Afterwards, there are emerging works striving to extract more meaningful and discriminative features with awareness of local and global information. For example, PointNet++~\cite{qi2017pointnet++} exploits the local geometry structure by aggregating features of neighbouring points. 
DGCNN \cite{wang2018dynamic} considers dynamic neighbors based on the feature distance. PointCNN~\cite{li2018pointcnn} permutes order of points in a local region to apply shared convolution for all candidate points. 
These deep learning-based methods have achieved promising results in point cloud classification and segmentation. Moreover, they are also adopted as backbones to extract high dimensional features in other point cloud processing tasks, such as detection~\cite{shi2019pointrcnn, qi2019deep,qi2018frustum}, registration~\cite{yew20183dfeat,lu2019deepvcp,wang2019deep}, and reconstruction~\cite{deng2018ppf,yang2018foldingnet, achlioptas2018learning}.

Recently, Yu \textit{et al.} \cite{yu2018pu} proposed the first deep learning algorithm for point cloud upsampling, called PU-Net, which employs PointNet++  to extract features and the expands the features by multi-branch MLPs.  
Although PU-Net outperforms the previous optimization-based approaches, it overlooks the spatial relations among the points severely and cannot produce dense point clouds with high quality. The follow-up work EC-Net \cite{yu2018ec}, adopts a joint loss of 
point-to-edge distance to preserve sharp edges. However, EC-Net requires training data with annotated edge and surface information, which is tedious to obtain in practice.
Inspired by the cascaded structure in image super-resolution, Wang \textit{et al.} \cite{yifan2019patch} proposed 3PU-Net which can progressively upsample an input to a relatively large upsampling factor, say 16$\times$. 
However, 3PU-Net does not model the local geometry well. Also, it requires a careful step-by-step training. 
In addition, the way of appending a 1D code to expand features limits each subnet upsamples an input by a factor $2$, and thus 3PU-Net only supports the overall upsampling factor in powers of $2$.
By introducing an additional discriminator, Li \textit{et al.} \cite{li2019pu} developed an adversarial framework, called PU-GAN, to boost the quality of upsampled point clouds. 
Qian \textit{et al}.~\cite{qian2020pugeo} proposed the first geometry-centric network PUGeo-Net. 
It first approximates the augmented Jacobian matrix of a local parameterization and then performs refinement along the normal direction of the estimated tangent plane.  
The above existing deep learning-based methods have to be separately trained 
for each upsampling factor, which restricts their flexibility in practice. 

Note that in parallel to our work, some contemporaneous works for point cloud upsampling have emerged recently. For example,
Qian \textit{et al}.~\cite{qian2021pu} introduced PU-GCN, which uses a multi-scale graph convolutional network to encode the local information of a typical point from its neighborhoods. Li \textit{et al}.~\cite{li2021point} disentangled the generator into two cascaded sub-networks, with the latter one applying local and global refinement for each point. Inspired by Meta-SR~\cite{hu2019meta} for image super-resolution, Ye \textit{et al}.~\cite{ye2021meta} proposed Meta-PU, which adopts residual graph convolution blocks to adjust weights for different upsampling factors dynamically. Note that for each upsampling factor, Meta-PU first generates $R_{max}\times$ points and then adopts farthest point sampling (FPS)~\cite{eldar1997farthest} to downsample the resulting dense point cloud to the desired ratio. However, FPS, as a post-processing step, is very time-consuming, especially when dealing with dense point clouds.

\subsection{Self-attention Mechanism} 

The self-attention mechanism~\cite{bahdanau2014neural,lin2017structured} was first proposed for neural machine translation to align long sentences. The self-attention layer transforms  input vectors to three kinds of vectors, namely Query, Key and Value vectors. 
The output of the self-attention layer is the weighted sum of the Value, where the weights are obtained by a compatibility function taking Query and Key as inputs. Based on self-attention, Vaswani \textit{et al.}~\cite{vaswani2017attention} proposed Transformer, which involves multi-head attention to aggregate the input, 
and the following Transformer-based frameworks~\cite{devlin2018bert,dai2019transformer,yang2019xlnet} achieve great success in 
natural language processing. Moreover, the self-attention and Transformer mechanism have also inspired many tasks in computer vision, 
such as image classification~\cite{hu2019local,ramachandran2019stand,zhao2020exploring,dosovitskiy2020image}, image generation~\cite{zhang2019self}, and object detection~\cite{carion2020end}. Recently, self-attention has also been used for 3D point cloud processing~\cite{yang2019modeling, guo2020pct, zhao2020point}, as the self-attention and Transformer mechanism process data discretely, making them naturally suitable for point cloud data with irregular structures. Note that the existing works ~\cite{yang2019modeling, guo2020pct, zhao2020point} mainly adopt self-attention to enhance feature extraction 
in high-level tasks like classification.
\begin{figure}[t]
\centering
\includegraphics[width=0.4\textwidth]{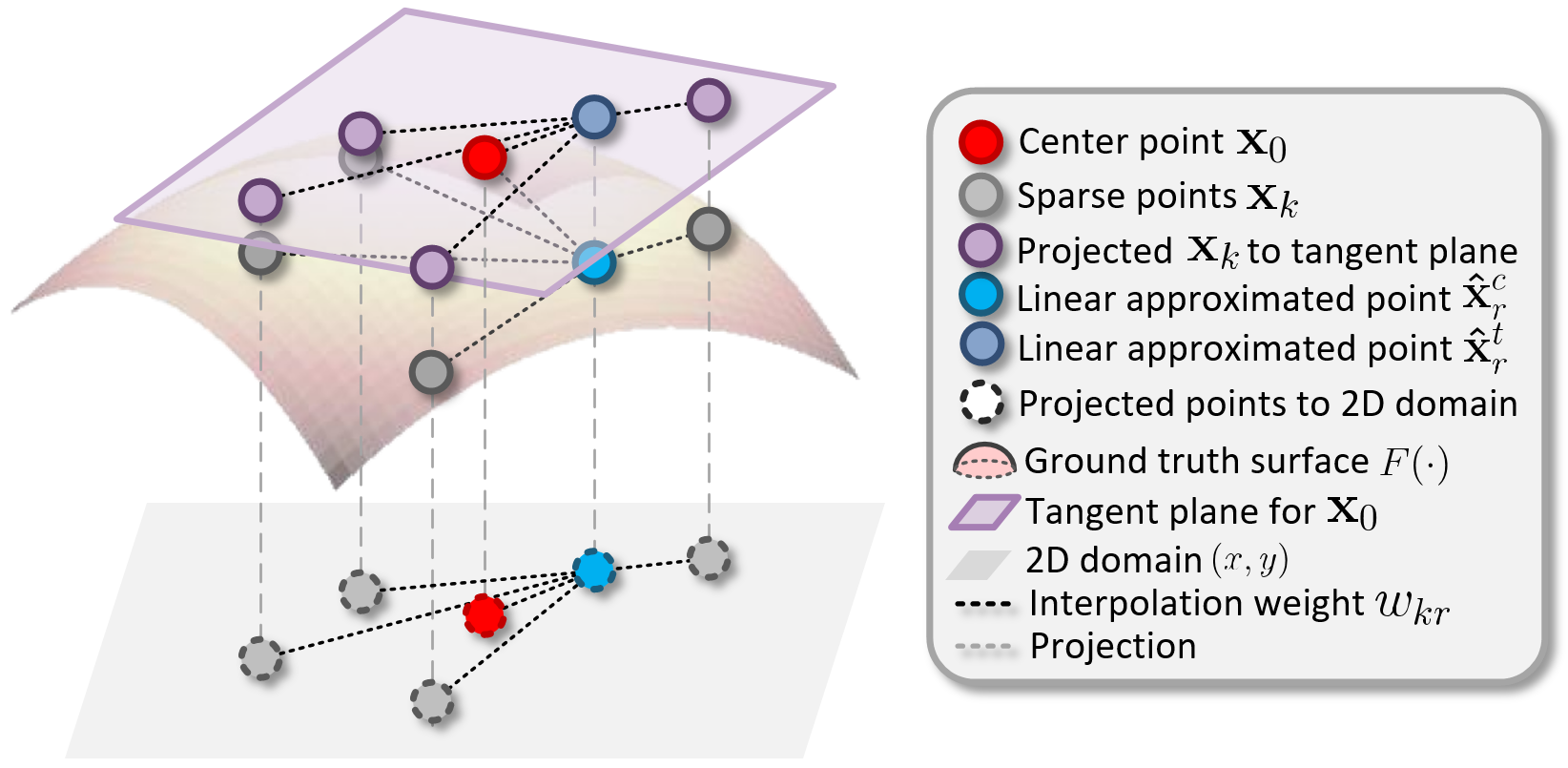}
\caption{The local neighborhood of a 3D surface around point $\mathbf{x}_i$ (red) can be approximated by the tangent plane at $\mathbf{x}_i$. Therefore, the upsampled point $\mathbf{p}_i^r$ can also be approximated by an affine combination of neighboring points projected onto the tangent plane. To avoid calculation of tangent plane, $\mathbf{p}_i^r$ can also be estimated by the linear interpolation directly from sparse neighborhood points.}
\label{fig:surface}
\end{figure}

 \begin{figure*}[t]
\centering
\includegraphics[width=1\textwidth]{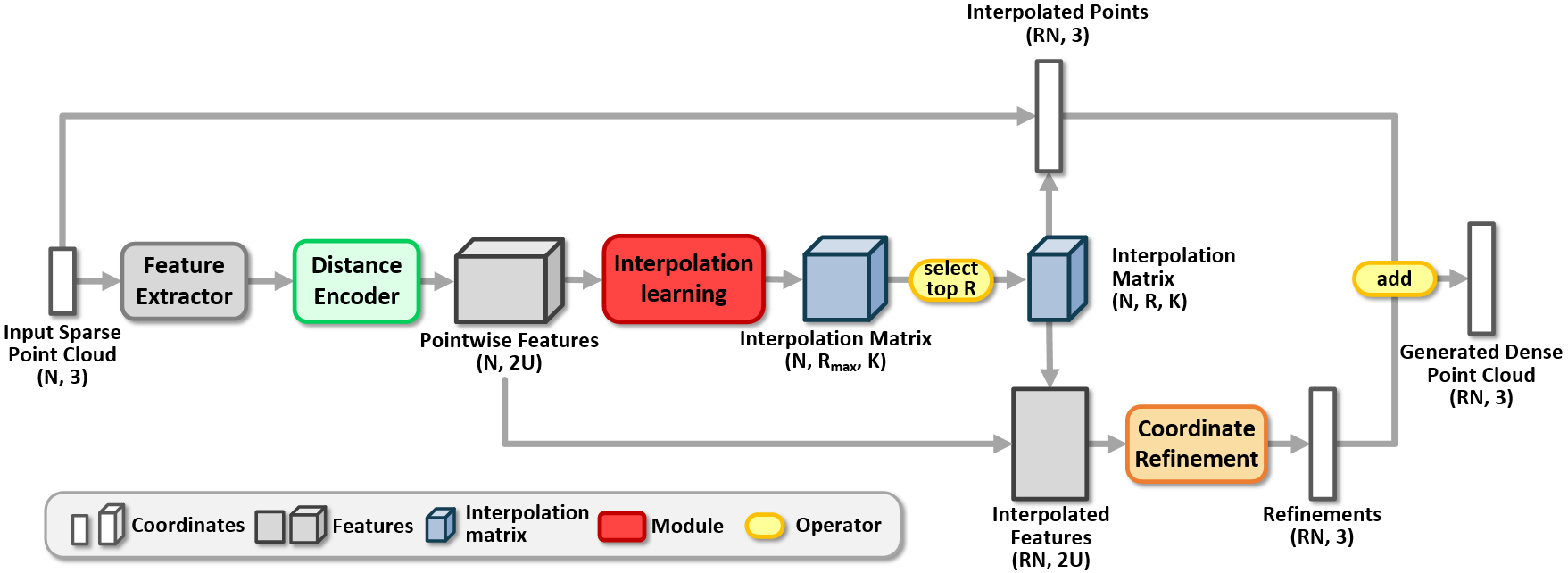}
\caption{The flowchart of the proposed method. Given a sparse point cloud with $N$ points, it first learns a $U$-dimensional feature for each point (i.e., the feature learning module) and also embeds the relative position information of $K$ nearest neighbouring (KNN) points into the $U$-dimensional features (i.e., the distance encoder module). 
Then the two types of high-dimensional features are concatenated to regress unified and sorted interpolation weights (i.e., the weight learning module), which coarsely interpolate the input sparse point cloud into a dense one. 
The coarse point cloud is finally refined via a self-attention-based refinement module, which regresses an offset for each point from the corresponding feature (see Fig. \ref{fig:refine} for the detailed network architecture). 
After one-time end-to-end training, the proposed method is capable of handling flexible upsampling factors not greater than the maximum factor $R_{max}$.}
\label{fig:flow}
\end{figure*}

\section{Problem Formulation}\label{sec:formulation} 
Denote by $\mathcal{X}=\{\mathbf{x}_i|\mathbf{x}_i\in\mathbb{R}^{3}\}_{i=1}^M$ a sparse point cloud with $M$ points and a user-specified scaling factor $R$. Our objective is to generate a dense point cloud $\mathcal{P}_R=\{\mathbf{p}_{i}^r|\mathbf{p}_i^{r}\in\mathbb{R}^{3}\}_{i, r=1}^{M, R}$ with $\mathbf{p}_i^r$ being the $r$-th upsampled point from $\mathbf{x}_i$, which contains more geometric details to approximate the underlying surface. Specifically, for each point of $\mathcal{X}$, we interpolate a certain number of nearest neighboring points located in its neighborhood to generate additional $R$ points.

In order to formulate the problem, we first consider a local neighborhood of point $\mathbf{x}_i=(x_i,y_i,z_i)$ and its $K$ nearest neighboring (KNN) points $\left\{\mathbf{x}_i^k=(x_i^k, y_i^k, z_i^k)\right\}_{k=1}^K$. We assume the surface is locally smooth at $\mathbf{x}_i$ so that it can be represented by a smooth implicit function $f(\cdot)$, i.e.,
\begin{equation}
    f(x,y,z)=0.
\end{equation}
Therefore, the neighboring points satisfy $f(x_i^k, y_i^k, z_i^k)=0$. 
If the partial derivative $\frac{\partial f}{\partial z}$ does not vanish, the local surface can be expressed explicitly as a height function $F:\mathbb{R}^2\rightarrow\mathbb{R}$ using the implicit function theorem~\cite{hildebrand1962advanced}, i.e.,
\begin{equation}
    z=F(x,y).
\end{equation}
Using Taylor expansion, we can locally approximate the surface at $(x_i, y_i)$ as 
\begin{equation}
\begin{split}
    z(x,y)=&F(x_i,y_i)+\nabla F(x_i,y_i)^\mathsf{T}\cdot(x-x_i, y-y_i)\\
    &+O\left((x-x_i, y-y_i)^2\right),
    \end{split}
\end{equation}
where $O\left((x-x_i, y-y_i)^2\right)$ contains the high-order terms of vector $(x-x_i, y-y_i)$. To generate more points $\left\{\mathbf{p}_i^r=(\hat{x}_i^r, \hat{y}_i^r, \hat{z}_i^r)\right\}_{r=1}^R$ locating on the surface, we adopt a 3-step strategy. 

First, we express the $x$- and $y$-coordinates of each new point (which are also the parameters of the parametric function $z(\cdot,\cdot)$) as the affine  combination of known points:
\begin{equation}
    (\hat{x}_i^r, \hat{y}_i^r)=\sum_{k=1}^K w_i^{k,r} (x_i^k, y_i^k),
        \label{equ:linear}
\end{equation}
where the weights $w^{k,r}$ are non-negative and satisfy partition of unity
\begin{equation}
  \sum_{k=1}^K w_i^{k,r}=1, \forall r.
          \label{equ:sum}
\end{equation}
Second, we define a linear function $H:\mathbb{R}^2\rightarrow\mathbb{R}$
\begin{displaymath}
H(x,y)\triangleq F(x_i,y_i)+\nabla F(x_i,y_i)^\mathsf{T}\cdot(x-x_i, y-y_i),
\end{displaymath}
as the first-order approximation of $z(x,y)$.
Geometrically speaking, $H(x,y)$ represents the tangent plane of $z(x,y)$ at $(x_i,y_i)$.
Therefore, the $z$-coordinate $\hat{z}_r$ can be approximated as 
\begin{eqnarray}
\nonumber \hat{z}_i^r&\approx H(\hat{x}_i^r, \hat{y}_i^r)=H\left(\sum_{k=1}^K w_i^{k,r} (x_i^k, y_i^k)\right)\\&=\sum_{k=1}^K w_i^{k,r} H(x_i^k, y_i^k)
    \approx \sum_{k=1}^K w_i^{k,r} z_i^k.
\end{eqnarray}
Define $\mathbf{\widehat{p}}_i^r\triangleq \left(\hat{x}_i^r, \hat{y}_i^r,\sum_{k=1}^K w_i^{k,r} H(x_i^k, y_i^k)\right)$ and $\mathbf{\widetilde{p}}_i^r\triangleq \left(\hat{x}_i^r, \hat{y}_i^r,\sum_{k=1}^K w_i^{k,r} z_i^k\right)$. 
Note that both $\mathbf{\widehat{p}}_i^r$ and $\mathbf{\widetilde{p}}_i^r$ are \textit{linear approximations} for $\mathbf{p}_i^r$ (see Fig.~\ref{fig:surface}).
Since each $H(x_k,y_k)$ is on the tangent plane, the combined point $\mathbf{\widehat{p}}_i^r$ is also on the tangent plane. 
In contrast, $\mathbf{\widetilde{p}}_i^r$ is a linear combination of sample points $\{\mathbf{x}_i^k\}$, therefore it is in its convex hull. 

Third, as the linear approximation $\mathbf{\widehat{p}}_i^r$ requires tangent plane estimation and the projection of neighborhood points which are non-trivial over point clouds,
for simplicity we approximate $\mathbf{p}_i^r$ by using $\mathbf{\widetilde{p}}_i^r$, i.e. the direct interpolation of  $\mathbf{x}_k$, 
together with an approximation error $\mathbf{e}_i^r=(\delta x, \delta y, \delta z)$, i.e.,
\begin{equation}
    \mathbf{p}_i^r = \mathbf{\widetilde{p}}_i^r +\mathbf{e}_i^r. 
    \label{equ:explicity form}
\end{equation}

In summary, combining Eqs.   \eqref{equ:linear}, \eqref{equ:sum}, \eqref{equ:explicity form}, and the definition of $\mathbf{\widetilde{p}}_i^r$, we can formulate the upsampling procedure from a sparse point cloud $\{\mathbf{x}_i\}$ to a dense point cloud $\{\mathbf{p}_i^r\}$ as
\begin{equation}
    \mathbf{p}_i^r = \sum_{k=1}^K w_i^{k,r}\mathbf{x}_i^k  +\mathbf{e}_i^r, \text{ where }   \sum_{k=1}^K w_i^{k,r}=1, w_i^{k,r}\geq 0, \forall r.
    \label{equ:final}
\end{equation}
Hence,
the problem of interpolating a 3D point cloud is boiled down to determining the interpolation
weights $\left\{w_i^{k,r}\right\}$ and the high-order approximation errors $\left\{\mathbf{e}_i^r\right\}$\footnote{Note that the accurate number of nearest neighboring points (i.e., $K$) is not necessary, because the values of the weights have taken this issue into account, i.e., a point is not a neighbour when $\omega^{k,r}=0$. }.

\section{Proposed Method}
\label{sec:proposed method}
\subsection{Overview}
Motivated by the explicit formation in Section \ref{sec:formulation}, we propose a novel data-driven framework to realize 3D point cloud upsampling in an end-to-end fashion, in which the interpolation weights and the approximation error in Eq. (\ref{equ:final}) are adaptively learned for each point of $\mathcal{X}$, by analyzing its local geometry property. As illustrated in  Fig.~\ref{fig:flow}, the proposed framework is a \textit{lightweight} neural network, which is mainly composed of three phases, i.e., local feature embedding, learning interpolation weights, and coordinate refinement. Specifically, it first embeds an input point cloud into a high-dimensional feature space point-by-point via local geometry-aware feature extraction. 
Then, it learns interpolation weights by regressing the resulting high-dimensional features, which are utilized to interpolate nearest neighbouring points, leading to a coarse upsampled point cloud. 
 Finally, it adopts the self-attention mechanism to estimate the approximation errors to refine the coordinates of the coarse point cloud.  
For an input point cloud with $M$ points, we extract patches containing $N$ points, and then apply the proposed method for upsampling in a patch-by-patch manner.

Note that in contrast to existing deep learning-based 3D point cloud upsampling methods that support only a pre-defined and fixed upsampling factor, making them less practical for real-world applications, the proposed framework is able to achieve magnification-flexible upsampling, i.e., it can handle flexible factors after one-time training. Such a flexibility is credited to the unique principle of our framework, which allows us to learn \textit{unified} and sorted interpolation weights. That is, the network is initialized with the maximum factor $R_{\max}$, and  the interpolation with a random $R$ ($R\leq R_{max}$) is performed in each iteration 
during training, i.e., the top-$R$ groups of estimated weights are selected for the $R\times$ upsampling, such that the learned groups of interpolation weights are naturally sorted. Therefore, during inference the top-$R$ groups of estimated interpolation weights could be selected for a specific factor. 

In what follows, we will detail the proposed framework phase by phase.

\subsection{Geometry-aware Local Feature Embedding}
In this phase,  each 3D point $\mathbf{x}_i$ of $\mathcal{X}$ is projected onto a high-dimensional feature space, denoted by $\mathbf{c}_i\in\mathbb{R}^{U}$.
Particularly, 
we adopt the dynamic graph CNN (DGCNN)~\cite{wang2018dynamic} to realize such a process.  Unlike previous deep feature representation methods for point clouds~\cite{qi2017pointnet, qi2017pointnet, li2018pointcnn} which are applied to individual points or a fixed graph constructed with the distance between coordinates, DGCNN defines the local neighborhood based on the distance between features obtained in the preceding layer. Specifically, 
denote by $\mathcal{E}\subset\mathcal{X}\times\mathcal{X}$  the edges calculated by $k$-nearest neighbors, then the initial directed graph $\mathcal{G} = (\mathcal{X}, \mathcal{E})$ is updated dynamically from one layer to another layer, based on the feature distance. In addition, it involves dense connections to aggregate multiple levels of features. Though using the local neighborhood in feature space, the learned feature representation $\mathbf{c}_i$ encodes both local and non-local information, while still keeping the permutation invariant property.

Moreover, we adopt a distance encoder~\cite{hu2020randla} to explicitly embed the relative position between points. Such an explicit embedding augments the corresponding point features to be aware of their neighborhood information. 
Let $\mathcal{S}_i^K=\{\mathbf{x}_i^k\}_{k=1}^K$ be the set of $K$ nearest neighbouring points of $\mathbf{x}_i$ in the sense of the Euclidean distance, and accordingly the associated high-dimensional features of the $K$ points obtained by DGCNN are  denoted by $\{\mathbf{c}_i^k\}_{k=1}^K$.    
The distance encoder employs an MLP to obtain a high-dimensional feature $\mathbf{r}_i^k\in\mathbb{R}^U$ for each neighbouring point, i.e.,  
\begin{equation}
    \mathbf{r}_i^k=\mathsf{MLP}\left(\mathbf{x}_i\oplus\mathbf{x}_i^k\oplus(\mathbf{x}_i-\mathbf{x}_i^k)\oplus\|\mathbf{x}_i-\mathbf{x}_i^k\|_2\right),
\end{equation}
where $\oplus$ is the concatenation operator,  $\|\cdot\|_2$ is the $\ell_2$ norm of a vector, and $\mathsf{MLP}(\cdot)$ denotes the MLP process. The encoded relative distance feature is  further concatenated to the feature $\mathbf{c}_i^k$ by DGCNN to form $\mathbf{\widetilde{c}}_i^k\in\mathbb{R}^{2U}$:
\begin{equation}
    \mathbf{\widetilde{c}}_i^k=\mathbf{c}_i^k\oplus \mathbf{r}_i^k.
\end{equation}
With the explicit encoding of local coordinate information, the high-dimensional feature can capture local geometric patterns.

\subsection{Learning Unified and Sorted Interpolation Weights}
\label{sec:weights}
As aforementioned, given a upsampling factor $R$, we aim to generate $R$ points for each neighbouring region of input point $\mathbf{x}_i$. 
As analyzed in Section~\ref{sec:formulation}, the coarse prediction $\mathbf{\widetilde{p}}_i^r$ can be obtained as the affine combination of the $K$ nearest neighboring points, i.e.,  
\begin{align}
    &\mathbf{\widetilde{p}}_i^r=\sum_{k=1}^K w_i^{k,r}\mathbf{x}_i^k \nonumber \\
    \text{subject to }&\mathbf{x}_i^k\in\mathcal{S}_i^K, ~\sum_{k=1}^K w_i^{k,r}=1, \text{and}~w_i^{k,r}\geq 0. 
    \label{eqn:sum}
\end{align}
To this end, we learn the interpolation weights using a weight learning module, which consists of MLPs applied to the point-wise feature $\mathbf{\widetilde{c}}_i^k$.
To achieve magnification-flexible interpolation, 
unified and sorted interpolation weights $\mathbb{W}_i^k=\left[\widetilde{w}_i^{k,1}, \widetilde{w}_i^{k,2}, \dots, \widetilde{w}_i^{k,R_{max}}\right]\in\mathbb{R}^{R_{max}}$ 
are learned, i.e., the output size of the weight learning module is initialized to the number of weights for the upsampling with a maximum factor $R_{\max}$, and the learning of such unified weights is modeled as 
\begin{equation}
\mathbb{W}_i^k=\mathsf{MLPs}\left(\mathbf{\widetilde{c}}_i^k\right).
\end{equation}
Then, for a specific upsampling factor $R$, the top-$R$ weights in a canonical order of $\mathbb{W}_i^k$ are selected as the corresponding interpolation weights, i.e.,
\begin{equation}
\mathbf{\widetilde{w}}_i^k = \left[\widetilde{w}_i^{k,1}, \widetilde{w}_i^{k,2}, \cdots, \widetilde{w}_i^{k,R}\right] \subseteq \mathbb{W}_i^k.
\end{equation}
Such a flexible manner is enabled by our unique formulation of the upsampling problem and  our special training strategy, i.e., in each iteration of the training process, upsampling with a randomly selected scale factor is performed, and the corresponding weights are updated, so that the learned weights are naturally sorted (see Section IV-E for details).  

Moreover, to meet the partition of unity constraint in Eq.~(\ref{eqn:sum}), we normalize the weights using a softmax layer, i.e., 
\begin{equation}
    w_i^{k,r}=\frac{e^{\widetilde{w}_i^{k,r}}}{\sum_{k=1}^K e^{\widetilde{w}_i^{l,r}}}.
\end{equation}
As the high-dimensional features explicitly encode relative distance information and local geometry details, it is expected that the weights, which can encourage the interpolated points to fit the underlying surface well, will be predicted.

\subsection{Self-attention-based Coordinate Refinement}
As formulated in Section \ref{sec:formulation}, the generated point $\mathbf{\widetilde{p}}_i^r$ via directly interpolating neighbouring points is distributed in the convex hull of the neighbouring points, but not necessary on the underlying curved surface. Therefore, we need an approximation error $\mathbf{e}_i^r$ to compensate the loss. In this subsection, we adaptively learn such approximation errors to refine the coarse predictions.
\begin{figure}[H]
\centering
\includegraphics[width=0.5\textwidth]{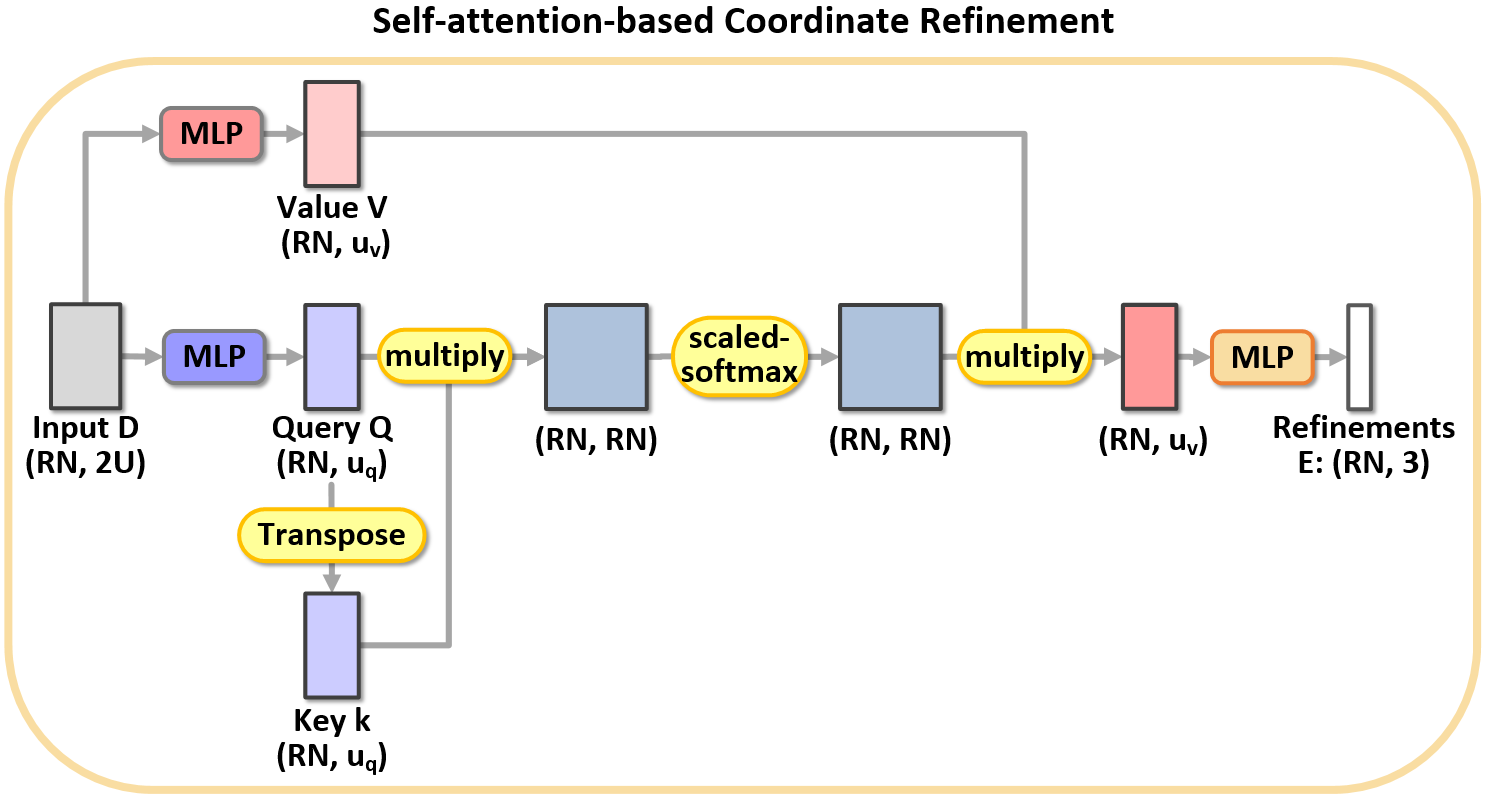}
\caption{The flowchart of the self-attention-based coordinate refinement module.}
\label{fig:refine}
\end{figure}

Similar to Eq. (\ref{eqn:sum}), we can also obtain the corresponding feature 
$\mathbf{d}_i^r$\footnote{To simplify the notation, we also use  $\{\mathbf{d}_l\}_{l=1}^{NR}$ to represent the features.} for each interpolated point 
$\mathbf{\widetilde{p}}_i^r$ as 
\begin{equation}
    \mathbf{d}_i^r=\sum_{k=1}^K w_i^{k,r}\mathbf{\widetilde{c}}_i^k.
    \label{eqn:feature}
\end{equation}

As illustrated in Fig.~\ref{fig:refine}, we adopt the self-attention-based mechanism 
to estimate the high-order term in Eq. (\ref{equ:final})  from 
features $\{\mathbf{d}_l\}_{l=1}^{NR}$.  
Specifically, 
we first employ MLPs to generate the \textit{Query} vectors $\{\mathbf{q}_l\}_{l=1}^{NR}$ and \textit{Value} vectors $\{\mathbf{v}_l\}_{l=1}^{NR}$, i.e., 
\begin{equation}
\mathbf{q}_l = \mathsf{MLP}(\mathbf{d}_l), \mathbf{v}_l = \mathsf{MLP}(\mathbf{d}_l).
\end{equation}
Let $\mathbf{Q}=[\mathbf{q}_1;\cdots;\mathbf{q}_{NR}]\in\mathbb{R}^{NR\times u_q}$, $\mathbf{V}\in\mathbb{R}^{NR\times u_v}$,  
and $\mathbf{K}=\mathbf{Q}$,
 and the output of the self-attention layer is obtained as
\begin{equation}
\mathbf{\widetilde{D}}=\mathsf{Attention}(\mathbf{Q},\mathbf{K},\mathbf{V})=\mathsf{Softmax}\left(\frac{\mathbf{Q}\mathbf{K}^\mathsf{T}}{\sqrt{u_q}}\right)\mathbf{V},
\end{equation}
where $\mathsf{Softmax}(\cdot)$ refers to the softmax function, and $\widetilde{\mathbf{D}}=[\mathbf{\widetilde{d}}_1;\cdots;\mathbf{\widetilde{d}}_{NR}]\in\mathbb{R}^{NR\times u_v}$, which is then utilized to estimate the approximation error via an MLP: 
\begin{equation}
\mathbf{e}_i^r = \mathsf{MLP}\left(\mathbf{\widetilde{d}}_i^r\right),
\end{equation}
Finally, the refined point can be obtained as  
\begin{equation}
    \mathbf{p}_i^r=\mathbf{\widetilde{p}}_i^r+\mathbf{e}_i^r. 
\end{equation}

\subsection{Loss Function and Training Strategy}\label{sec:loss}
Let $\mathcal{\widetilde{P}}_R=\{\mathbf{\widetilde{p}}_i^r\}$ be the coarse prediction,   $\mathcal{P}_R=\{\mathbf{p}_i^r\}$ be the refined prediction, and 
$\mathcal{Y}_R=\{\mathbf{y}_l\}_{l=1}^{NR}$ be the ground-truth dense point cloud of the $R\times$ upsampling, associated with the normal attribute denoted by $\mathcal{N}_R=\{\mathbf{\overline{n}}_l\}$.  To train the proposed method end-to-end, we design a loss function composed of the reconstruction error of the coarse prediction, the reconstruction error of the refined prediction, a projection distance loss, and a uniform loss. 

To be specific, we adopt the Chamfer distance (CD) to measure the reconstruction errors, i.e., 
 \begin{align}
& L_{coarse}=\mathsf{CD}(\mathcal{\widetilde{P}}_R, \mathcal{Y}_R)  \\
& L_{refine}=\mathsf{CD}(\mathcal{P}_R, \mathcal{Y}_R),  
\end{align} 
where 
 \begin{equation}
 \mathsf{CD}(\mathcal{X}, \mathcal{Y})= \frac{1}{M}\left(\sum_{\mathbf{x}\subset \mathcal{X}}||\mathbf{x}-\phi_\mathcal{Y}(\mathbf{x})||_2+\sum_{\mathbf{y}\subset \mathcal{Y}}||\mathbf{y}-\psi_\mathcal{X}(\mathbf{y})||_2\right)  \nonumber
\end{equation} 
with $M$ being the number of points in $\mathcal{X}$ and  $\mathcal{Y}$, $\phi_\mathcal{Y}(\mathbf{x})=\argmin_{\mathbf{y}\subset \mathcal{Y}}\|\mathbf{x}-\mathbf{y}\|_2$, and $\psi_\mathcal{X}(\mathbf{y})=\argmin_{\mathbf{x}\subset \mathcal{X}}\|\mathbf{x}-\mathbf{y}\|_2$. 

We define the projection distance between upsampled point clouds and ground-truth ones as 
\begin{equation}
    L_{pro}=\frac{1}{RN}\sum_{\mathbf{y}_l\subset \mathcal{Y}_R}\left| \mathbf{\overline{n}}_l\cdot (\mathbf{y}_l-\psi_{\mathcal{P}_R}(\mathbf{y}_l))\right|.
\end{equation}
We also adopt the uniform loss~\cite{li2019pu} to ensure the uniformity of generated points, defined as
\begin{equation}
     L_{uni}=\sum_{j=1}^M U_{\text{imbalance}}(S_j)\cdot U_{\text{clutter}}(S_j),
\end{equation}
where $S_j$ is the point cloud subset of  $\mathcal{P}_R$, with the centroid being one of the $M$ seed points picked by farthest point sampling (FPS)~\cite{eldar1997farthest};  $U_{\text{imbalance}}(\cdot)$ measures the deviation of number of points in each subset; and  $U_{\text{clutter}}(\cdot)$ indicates the deviation of relative distances of points in each subset. 

We define the overall loss function for training the proposed method for upsampling  with a fixed factor $R$ as 
\begin{equation}
    L^R= \alpha L_{refine}+\beta L_{coarse} + \gamma L_{pro}+\zeta L_{uni},
    \label{eqn:totalloss}
\end{equation}
where $\alpha$, $\beta$, $\gamma$  and $\zeta$ are four positive parameters. 
Note that we do not require the ground-truth normals during testing.  

Our goal is 
to train the proposed method in a \textit{flexible} manner, i.e., after one-time training, our method is able to deal with flexible scaling factors during inference. A na\"{i}ve way is to use 
the summation of the loss function in Eq. (\ref{eqn:totalloss}) for various factors as the loss function, i.e., $L=\sum_j \lambda_j L^{R_j}$ where $\lambda_j>0$ is the weight to balance different factors. However, it is difficult to tune the parameters $\lambda_j$. In our implementation, we randomly select a factor to optimize in each iteration of the training process. Experimental results demonstrate the effectiveness and efficiency of the simple training strategy.
\begin{table*}[t!]
\centering
\caption{Quantitative comparisons of different methods with various scaling factors. We uniformly scaled the models into a unit cube, so the distance metrics are unitless. Here, the results refer to the average of 39 testing point clouds, and the best ones are highlighted in bold. For all metrics, the lower the better.}
\begin{tabular}{c|c||c |c c c c c}\Xhline{5\arrayrulewidth}
$R$ & Method &\makebox[3em]{Network}  & \makebox[3em]{CD } & \makebox[3em]{HD }  &\makebox[3em]{JSD } & \makebox[3.5em]{P2F mean} & \makebox[3.5em]{P2F std} \\
       &  &\makebox[3em]{size} & \makebox[3em]{$(10^{-2})$} & \makebox[3em]{$(10^{-2})$}  &\makebox[3em]{$(10^{-2})$} & \makebox[3em]{$(10^{-3})$} & \makebox[3em]{$(10^{-3})$}  \\
\Xhline{2\arrayrulewidth}
4$\times$ & EAR \cite{huang2013edge} & 13.6 MB & 3.034   & 5.567  & 9.037  & 7.801 &  8.794  \\
& PU-Net \cite{yu2018pu} & 9.4 MB  & 2.151 & 4.276 & 2.345 & 3.705 & 3.233  \\
&3PU-Net\cite{yifan2019patch}& 92.5 MB &  1.922 & 2.894  & 1.643 & 1.883 & 1.895 \\
& PU-GAN \cite{li2019pu}& 7.1 MB  & 1.915 & 3.608 & 1.727 & 1.920 & 2.931  \\
& PU-GAN-G \cite{li2019pu}& 7.1 MB  & 2.315 & 8.143 & 3.562 & 5.068  & 6.736  \\
& PUGeo-Net \cite{qian2020pugeo}& 27.1 MB  & 1.870 & \bf{2.776} & 1.800 & 1.568  & 1.578  \\
& Proposed & \bf{4.7 MB} &   \bf{1.772} & 2.804 & \bf{1.550} & \bf{1.382} & 1.579  \\
& Proposed (flexible) & 4.7 MB &   1.814 & 3.233 & 1.903 & 1.524 & \bf{1.490}\\\Xhline{2\arrayrulewidth}
8$\times$ &PU-Net \cite{yu2018pu} & 14.0 MB  & 1.684 & 5.507 & 1.763 & 3.714 & 3.321 \\
&3PU-Net \cite{yifan2019patch}& 92.5 MB  & 1.501 & 3.051 & 1.230 & 2.044 & 2.155   \\
& PU-GAN-G \cite{li2019pu}&  8.6 MB  & 1.482  & 3.110 & 1.209  & 2.403 & 2.215 \\
& PUGeo-Net \cite{qian2020pugeo}& 27.1 MB  & 1.424 & \bf{2.677} & 1.208 & 1.926  & 1.996  \\
& Proposed & \bf{4.7 MB} &  \bf{1.414} & 2.929 & \bf{1.153} & \bf{1.722} & \bf{1.763}  \\
& Proposed (flexible) & \bf{4.7 MB} &  1.419 & 3.327 & 1.262 & 1.817 & 1.899  \\\Xhline{2\arrayrulewidth}
12$\times$ & PU-Net \cite{yu2018pu} & 19.0 MB & 1.506 & 5.119 & 1.762 & 3.878 & 3.313  \\
&3PU-Net\cite{yifan2019patch}& - & - & - & - & - & -\\
& PU-GAN-G \cite{li2019pu}&  10.1 MB &1.360  & 4.094 & 1.522 & 3.024 & 2.644   \\
& PUGeo-Net \cite{qian2020pugeo}& 27.1 MB  & 1.250 & \bf{2.571} & 1.116 & 2.077  & 2.042  \\
& Proposed & \bf{4.7 MB} &\bf{1.208} & 3.200 & \bf{1.040}  & \bf{1.725} & \bf{1.795}  \\
& Proposed (flexible) & 4.7 MB & 1.216 & 3.335 & 1.080  & 1.814 & 1.883 \\\Xhline{2\arrayrulewidth}
16$\times$ & PU-Net \cite{yu2018pu} & 23.0 MB & 1.343 & 4.895 & 1.512 & 3.557 & 3.089  \\
&3PU-Net\cite{yifan2019patch} & 92.5 MB & 1.177 & 3.150 & 0.960 & 2.222 & 2.396  \\
& PU-GAN-G \cite{li2019pu}& 11.5 MB &  1.218 & 3.983 & 1.132 & 2.866 & 2.704  \\
& PUGeo-Net \cite{qian2020pugeo}& 27.1 MB  & 1.111 & \bf{2.957} & 1.023 & 1.770  & 2.036  \\
& Proposed & \bf{4.7 MB}  &  \bf{1.068} & 3.021 & \bf{0.870} & \bf{1.612} & \bf{1.844}  \\
& Proposed (flexible) & 4.7 MB &  1.088 & 3.550 & 1.001 & 1.754 & 1.808  \\
\Xhline{5\arrayrulewidth} 
\end{tabular}
\label{table:compare} 
\end{table*}

\section{Experimental results}
\label{sec:exp}

\subsection{Experiment Settings 
}\label{sec:setting}
\subsubsection{Dataset} We adopted the same training dataset as~\cite{li2019pu}, consisting of 120 3D mesh models. We applied Poisson disk sampling~\cite{corsini2012efficient} to each mesh model to generate sparse point clouds with 5000 points, and various ground-truth dense point clouds for different factors.
For the experiments on uniform data, 
 we cropped input point clouds into patches of $N=256$ points via  (FPS) for training. For the experiments on non-uniform data, patches of $N=256$ points were randomly sampled from uniformly distributed patches with $1024$ points for training.
Finally, we applied data augmentation techniques, including random scaling, rotation and point perturbation, to increase data diversity. 

\subsubsection{Implementation details} We empirically set the hyper-parameters of the loss function in Eq. (\ref{eqn:totalloss}) $\alpha=100$, $\beta=30$, $\gamma=100$ and $\zeta=1$, the parameter $K=32$ in KNN, and the maximal upsampling factor $R_{max}=16$. 
We used the Adam optimizer~\cite{kingma2015adam} with the learning rate 0.001. We trained the network with the mini-batch of size $8$ for $400$ epochs.  
When training the proposed method in the flexible scenario, we selected the upsampling factor $R$ from 4, 8, 12, and 16    
respectively with the probability 0.1, 0.2, 0.3, and 0.4 in each training iteration,   
and any integer upsampling factor $R\in[4, 16]$
can be conducted during inference.
We implemented the proposed framework in TensorFlow.
Besides, we also trained the proposed method for single upsampling factors.

\subsubsection{Compared methods} We compared the proposed method with optimization-based EAR\footnote{The code is publicly available at https://vcc.tech/research/2013/EAR} \cite{huang2013edge}, and five representative deep learning-based methods, which are PU-Net \cite{yu2018pu}, 3PU-Net \cite{yifan2019patch}, PU-GAN~\cite{li2019pu}, PU-GAN-G~\cite{li2019pu}, and PUGeo-Net \cite{qian2020pugeo}. 
The detailed settings are listed as follows.
\begin{itemize}
\item We tested EAR with the default hyper-parameters.  
Since EAR cannot set the exact number of the to-be-reconstructed point cloud, we generated dense point clouds by EAR 
with a slightly larger number of points, and then downsampled them to the desired number of points.
\item For 3PU-Net and PUGeo-Net, we adopted their official implementations and retrained  them on the same dataset and with the same data augmentations as ours for fair comparisons. Note that the architecture of 3PU-Net limits its upsampling factor only to be a power of 2.
\item 
Since PU-GAN utilized the same training dataset as ours, we directly used the 
pre-trained model released by the authors for testing 
to ensure correctness\footnote{Note that in the original paper PU-GAN was only trained and tested for the upsampling task with $R = 4$. For other upsampling factors, we failed to achieve satisfactory results by retraining PU-GAN. It is not unexpected as it is well-known that training a GAN framework is non-trivial. Therefore, we only provided the upsampling results of PU-GAN when $R = 4$.}.
\item Besides, to directly examine the ability of the generator, 
we also compared with the generator of PU-GAN, denoted as PU-GAN-G. We used the official code for the generator of PU-GAN and retrained it with its generative loss $\mathcal{L}_G$. 
\end{itemize}

\subsubsection{Evaluation protocols} We combined the 13 testing models in ~\cite{yifan2019patch} and the 26 testing models in ~\cite{li2019pu} to form a larger 
testing benchmark. Same as the inference settings of previous works~\cite{yifan2018patch,li2019pu}, we adopted FPS to sample anchor points on the input point cloud and $k$-NN to extract patches with point number $N=256$. Upsampled patches were further combined to $MR$ points by FPS.  We employed four commonly used metrics, i.e., Chamfer distance (CD)~\cite{thayananthan2003shape}, Hausdorff distance (HD)~\cite{berger2013benchmark}, point-to-surface distance (P2F), and Jensen-Shannon divergence (JSD)~\cite{achlioptas2018learning}, to evaluate different methods quantitatively. The P2F distance measures the difference between the upsampled point clouds and the corresponding ground truth 3D mesh models, while the other three metrics evaluate the difference between the upsampled point clouds and the corresponding ground truth dense point clouds. We also examined the uniformity of the upsampled points by using the normalized uniformity coefficient (NUC)~\cite{yu2018pu} under different disk area percentage $p$. We performed theses metrics on a whole point cloud normalized into a unit sphere for all compared methods. For all the metrics, the lower the value, the better the quality. 

\begin{table*}[th]
\centering
\caption{Performance and efficiency comparisons of our method with other deep learning-based methods that are enabled to be flexible ($R=4$).  
We measured the efficiency using the average patch processing time in seconds.}
\begin{tabular}{c||c c c c c |c|c c c}\Xhline{5\arrayrulewidth}
  Method  & \makebox[3em]{CD } & \makebox[3em]{HD }  &\makebox[3em]{JSD } & \makebox[3.5em]{P2F mean} & \makebox[3.5em]{P2F std} & \makebox[3em]{Network} & \makebox[3em]{Upsample} &\makebox[3em]{FPS}& \makebox[3em]{Total} \\
       & \makebox[4em]{$(10^{-2})$} & \makebox[4em]{$(10^{-2})$}  &\makebox[4em]{$(10^{-2})$} & \makebox[4em]{$(10^{-3})$} & \makebox[4em]{$(10^{-3})$}& \makebox[3em]{size} & \makebox[3em]{time} & \makebox[3em]{time} & \makebox[3em]{time}   \\
\Xhline{2\arrayrulewidth}
 PU-Net \cite{yu2018pu} & 2.169&	5.045	&2.332&	4.009	&3.741& 23.0MB&	0.036&	0.353&	0.390 \\
 PU-GAN-G \cite{li2019pu}& 2.008&	4.141&	2.025	&2.912	&3.083&	11.5MB &0.031&	0.353&	0.385 \\
 PUGeo-Net \cite{qian2020pugeo}& 2.024&	\bf{2.876}&	2.000	&2.832	&3.100&	27.1MB & 0.045 &	0.353&	0.398 \\\Xhline{\arrayrulewidth}
 Proposed (flexible) & \bf{1.814}&	3.234&	\bf{1.903}	&\bf{1.524}	&\bf{1.490}&	\bf{4.7MB} &0.046&	0.000	&\bf{0.046} \\\Xhline{5\arrayrulewidth}
\end{tabular}
\label{table:compare_fps} 
\end{table*}

\begin{figure*}[th]
     \centering
        \subfigure[]{\includegraphics[width=1.3in]{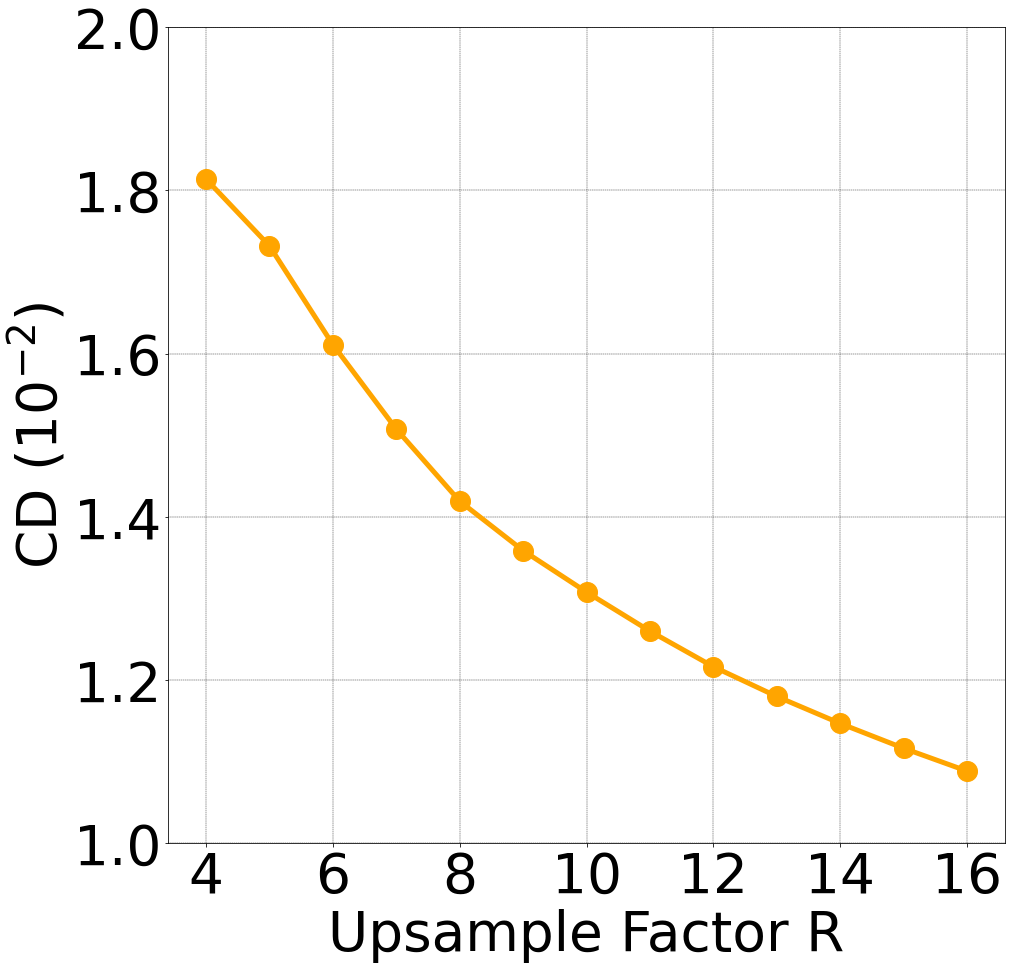}}
        \subfigure[]{\includegraphics[width=1.3in]{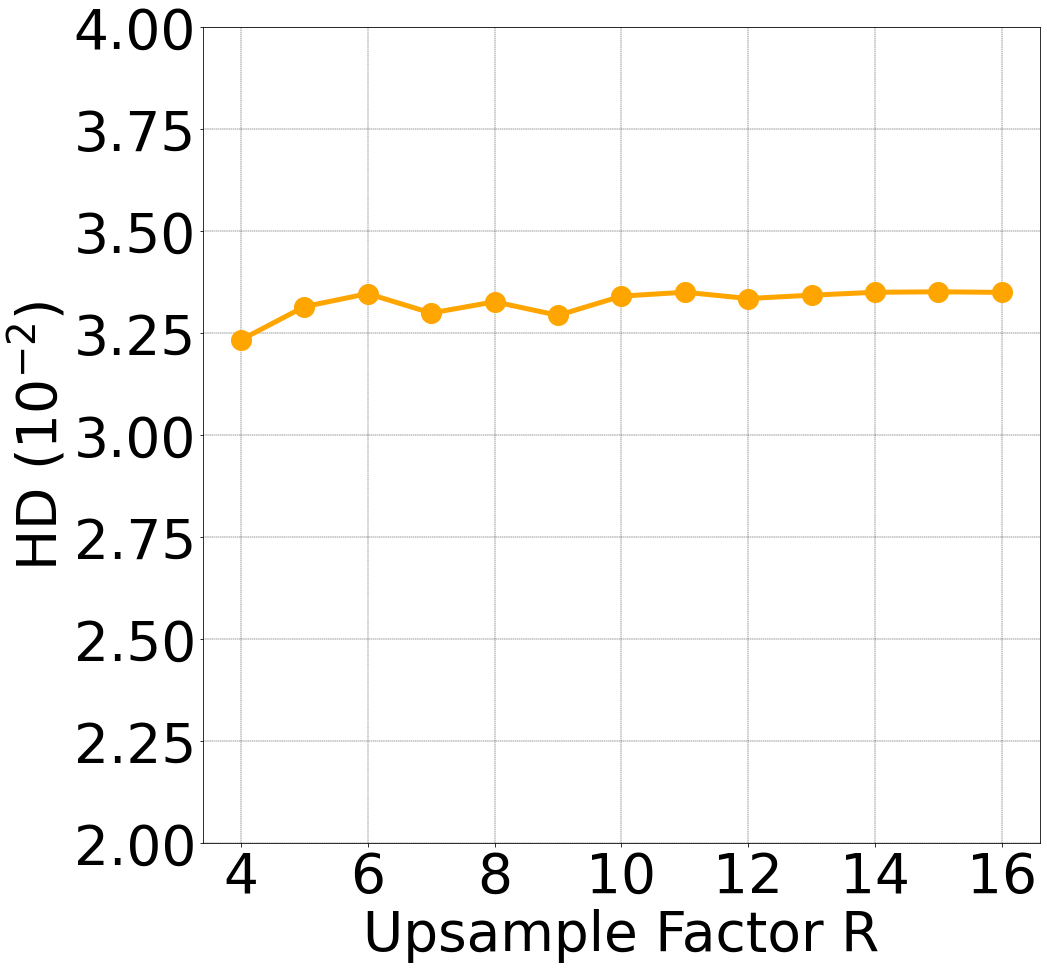}}
        \subfigure[]{\includegraphics[width=1.3in]{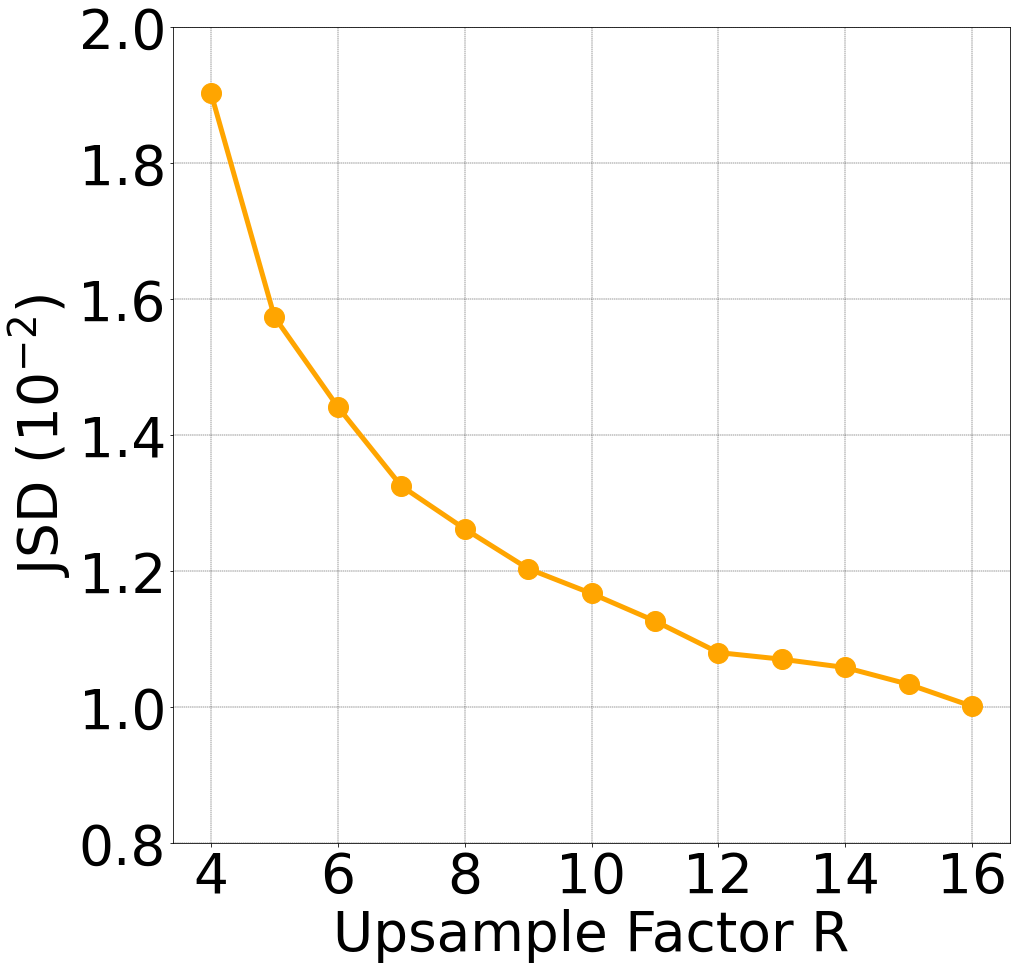}}
        \subfigure[]{\includegraphics[width=1.3in]{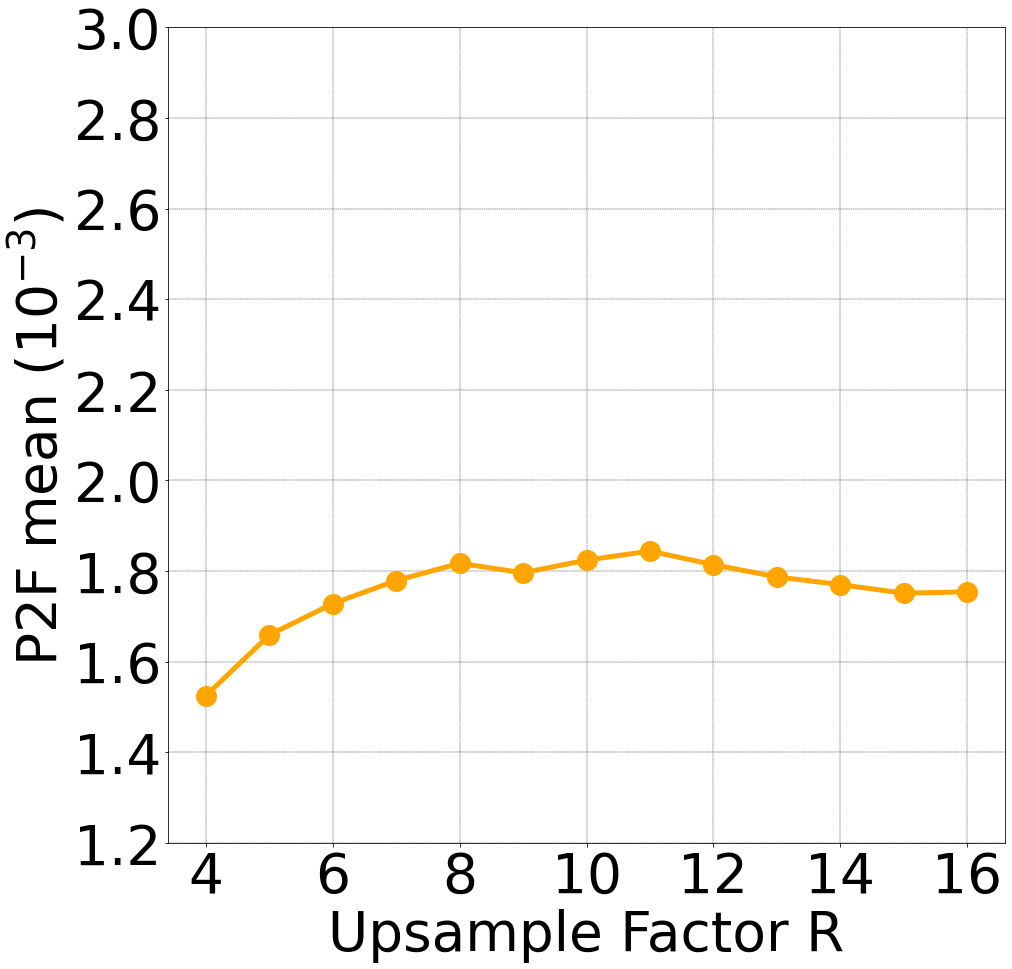}}
        \subfigure[]{\includegraphics[width=1.3in]{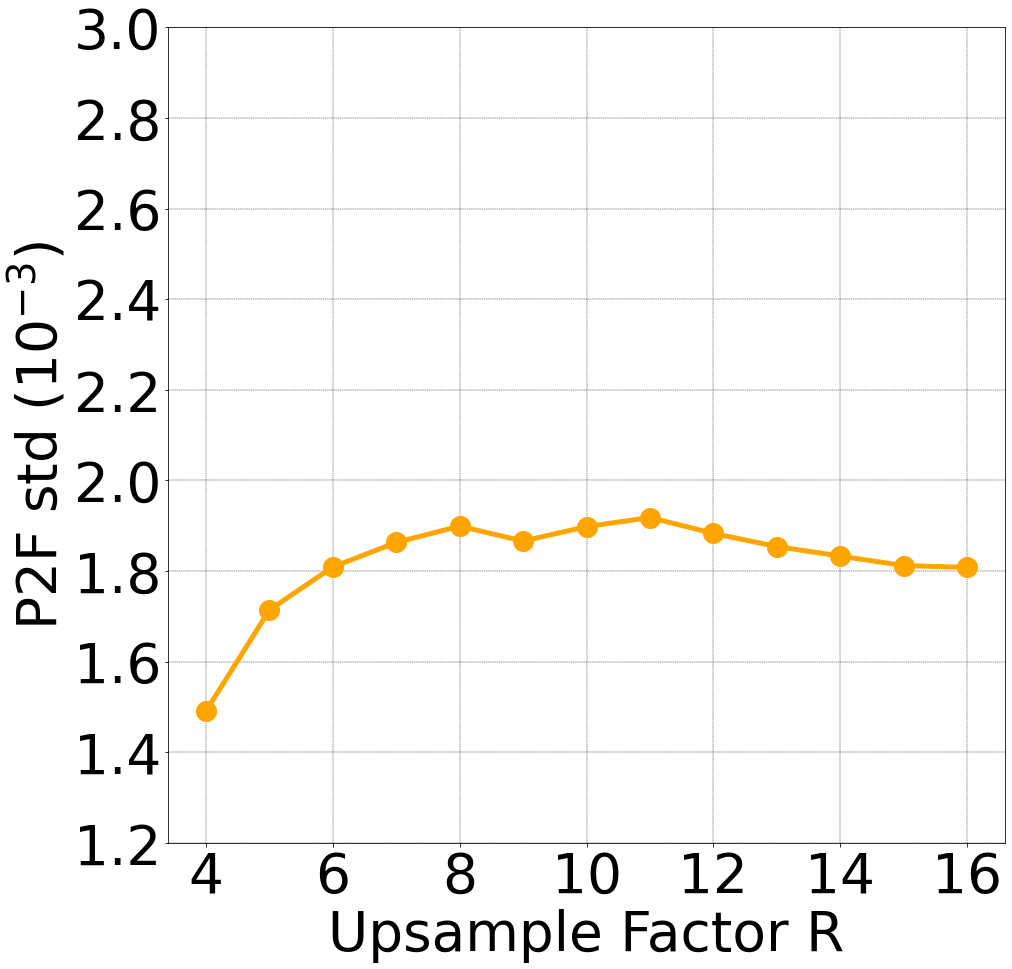}}
        \caption{Quantitative performance of the proposed method for all  the  integer  factors  between  4  and  16. (a) CD; (b) HD; (c) JSD (d) P2F mean; (e) P2F std.}
        \label{fig:arbitrary_quant}
\end{figure*}

\subsection{Evaluation on Uniform Data}
To evaluate the performance of different methods for upsampling uniform point clouds, we applied Poisson disk sampling to the 39 testing mesh models to generate testing point clouds with 2,048 points each. 

\subsubsection{Quantitative comparison} 
Note that we normalized all the testing data to unit spheres to perform the quantitative comparison. Table \ref{table:compare} lists the average results of 39 testing point clouds, where it can be seen that the proposed method trained for single upsampling factors achieves the best performance almost for all upsampling factors in terms of all the four metrics. 
The proposed method for flexible upsampling factors is slightly worse than the individual models; but for most metrics, it still outperforms the other state-of-the-art methods which have to be separately trained for each upsampling factor. 

To evaluate the memory-efficiency of different deep learning-based methods, we also compared their network sizes in Table~\ref{table:compare}, where it can be observed that the proposed method has the smallest network size. Due to the progressive upsampling manner, 3PU-Net can only deal with upsampling factors in powers of 2. 3PU-Net suffers from large memory consumption, which results in model size 20 times as much as ours.
PU-Net suffers from the linearly increasing network size since it adopts the independent multi-branch design for feature expansion. 

Intuitively, the compared deep learning-based methods, i.e., PU-Net, PU-GAN-G, and PUGeo-Net that have to be separately trained for each upsampling factor, can also achieve flexibility in the following simple manner: first upsampling input data using those methods trained with a relatively large factor, and then downsampling the results with a typical downsampling method to generate point clouds corresponding to smaller upsampling factors. To compare with our method, we downsampled the upsampled point clouds by PU-Net, PU-GAN-G, and PUGeo-Net with FPS from $16\times$ to $4\times$. 
As listed in Table~\ref{table:compare_fps},  their performance 
is worse than that of both our method and the corresponding models directly trained for $4\times$ upsampling in Table~\ref{table:compare}. Moreover, such a flexible manner
consumes more memory and inference time. 

In addition, to demonstrate the flexibility of the proposed method, Fig.~\ref{fig:arbitrary_quant} shows the performance of the proposed method
for all the integer factors between 4 and 16. As mentioned in Section \ref{sec:setting}, we only used factors 4, 8, 12, and 16 during training. From Fig.~\ref{fig:arbitrary_quant}, we can see  that  the  values  of  metrics  form smooth lines  with  the  factor  varying,  and  such a  smoothness  observation  validates  the  effectiveness  of  the flexibility of the proposed framework. Specifically, as the upsampling factor increases,  the  values  of  CD  and  JSD  consistently  decrease because these two metrics evaluate the distribution difference between  upsampled  point  clouds  and  corresponding  ground-truth  dense  point  clouds,  and  a  larger  factor  means  more generated points, mitigating the overall distribution difference.  

\begin{figure}[bh]
     \centering
        \subfigure[]{\includegraphics[width=0.6in]{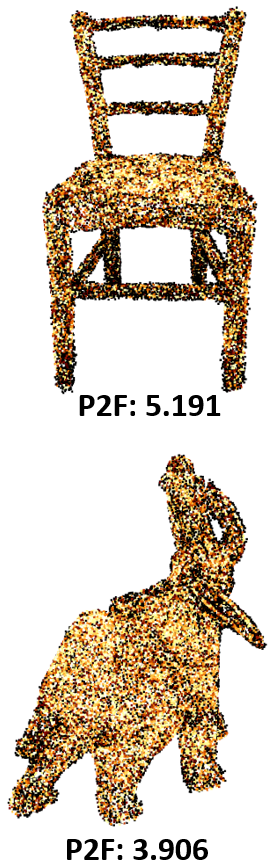}}
        \subfigure[]{\includegraphics[width=0.6in]{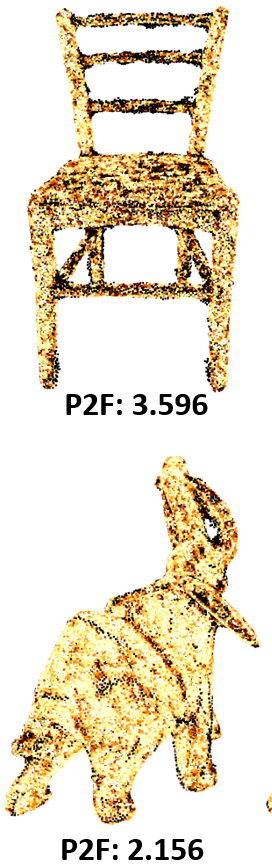}}
        \subfigure[]{\includegraphics[width=0.6in]{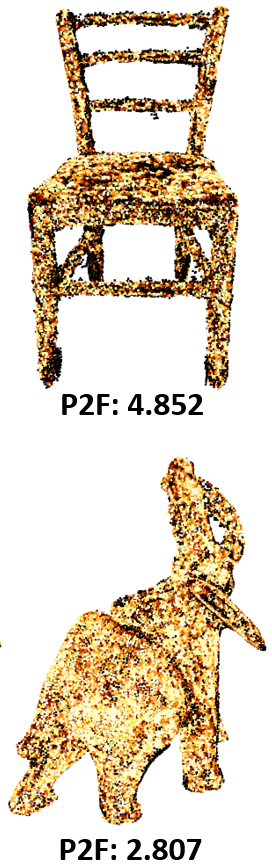}}
        \subfigure[]{\includegraphics[width=0.6in]{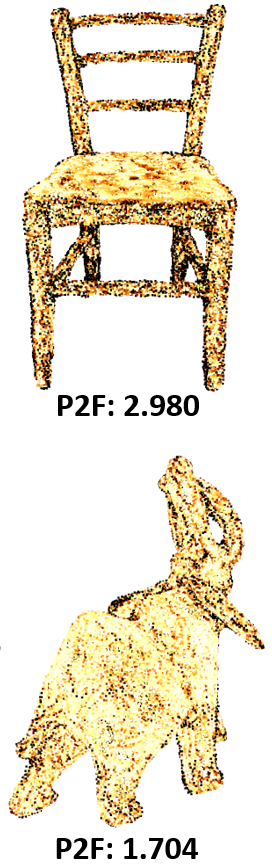}}
        \subfigure[]{\includegraphics[width=0.6in]{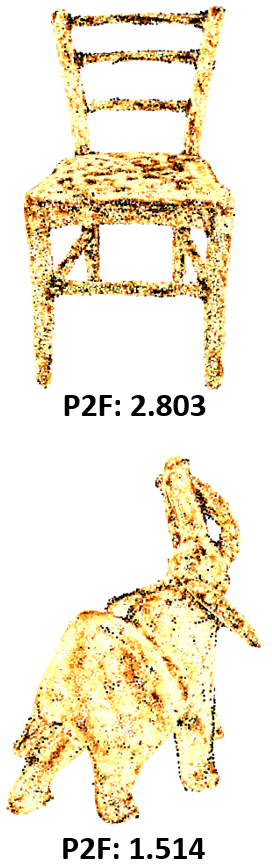}}
        \subfigure{\includegraphics[width=0.2in]{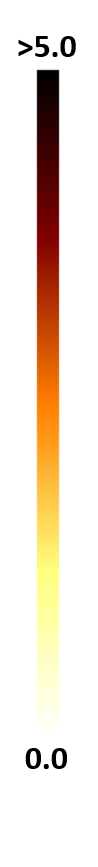}}
        \caption{Visual comparisons of the P2F errors between upsampled point clouds by different methods and ground-truth ones 
($R=16$). We visualized the P2F errors using colors. 
(a) PU-Net, (b) 3PU-Net, (c) PU-GAN-G, (d) PUGeo-Net, and (e) Proposed.}
        \label{fig:p2f}
\end{figure}

\subsubsection{Visual comparison} First, we visualized the point-wise P2F errors between the $16\times$ upsampled point clouds by different methods and corresponding ground truth 3D mesh models in Fig. \ref{fig:p2f}, where
we can observe that the proposed method produces smaller errors than the other compared methods. 

\begin{figure*}[bh]
     \centering
        \subfigure[]{\includegraphics[width=0.9in]{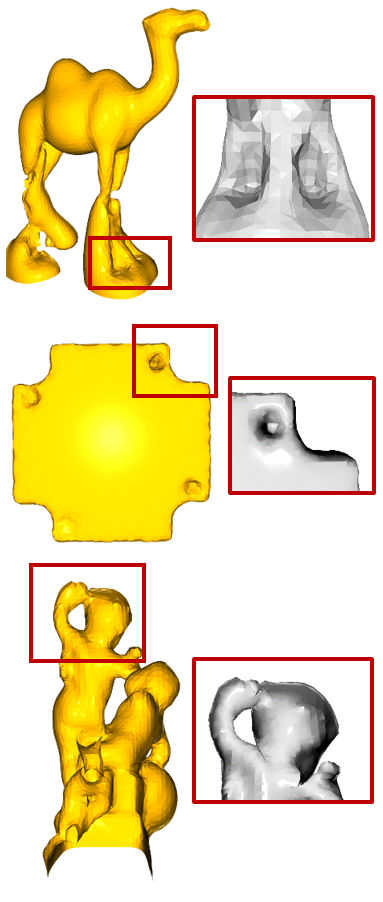}}
        \subfigure[]{\includegraphics[width=0.9in]{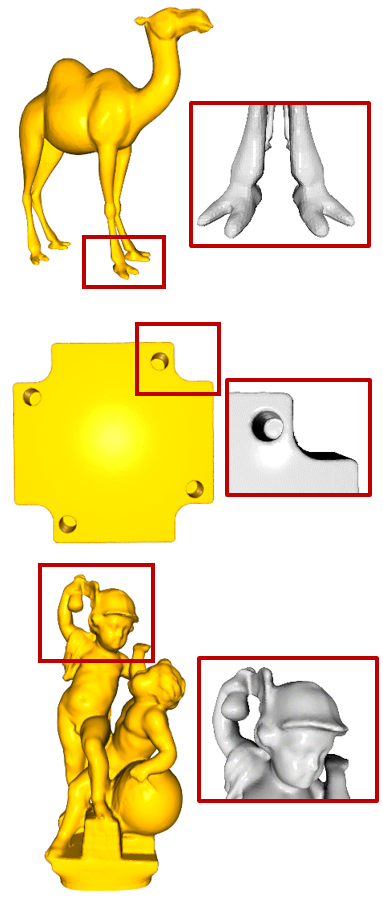}}
        \subfigure[]{\includegraphics[width=0.9in]{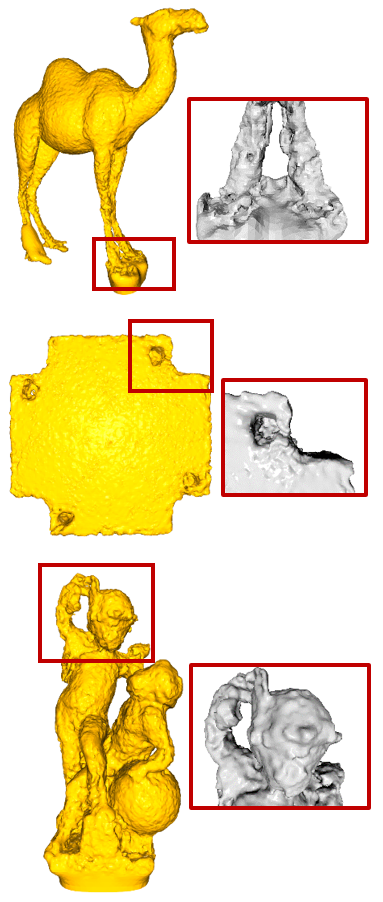}}
        \subfigure[]{\includegraphics[width=0.9in]{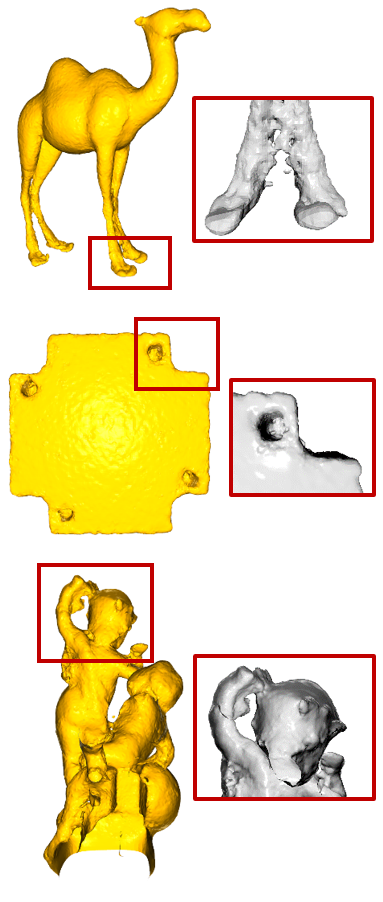}}
        \subfigure[]{\includegraphics[width=0.9in]{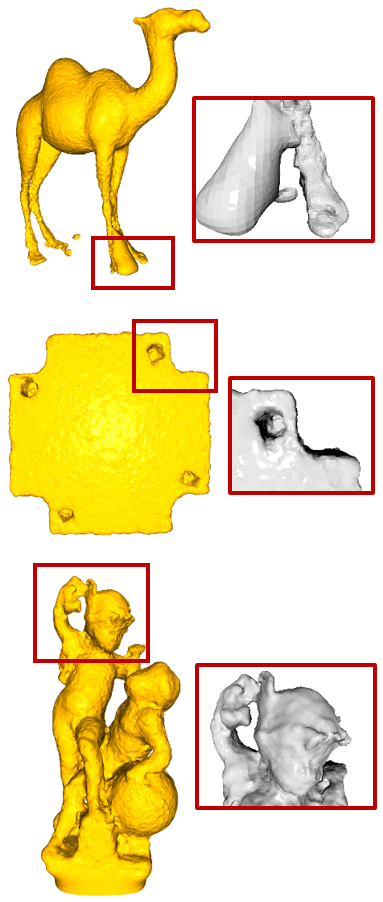}}
        \subfigure[]{\includegraphics[width=0.9in]{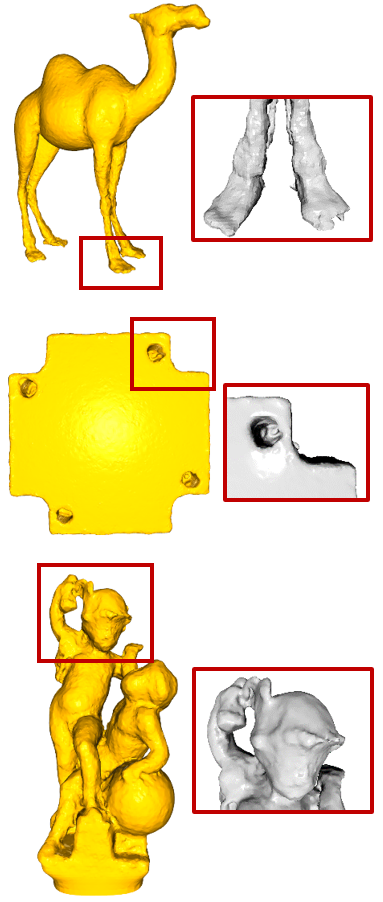}}
        \subfigure[]{\includegraphics[width=0.9in]{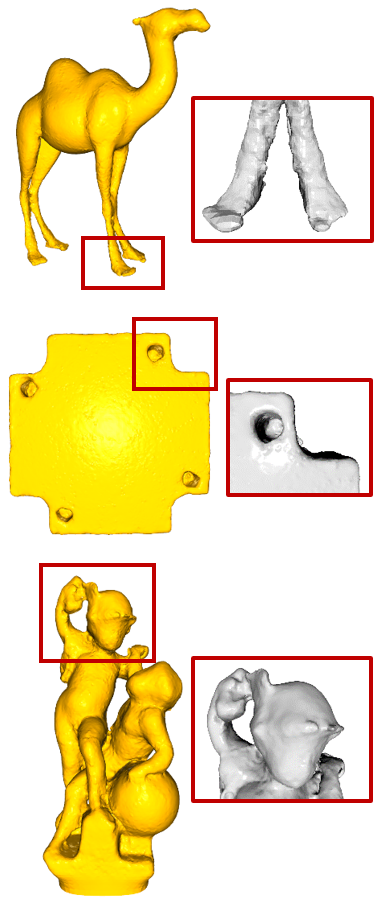}}
        \caption{Visual comparisons of the surfaces reconstructed from (a) sparse point clouds with 2,048 points each, (b) the ground-truth dense point clouds with 32,768 points each, 16$\times$ upsampling results of (c) PU-Net, (d) 3PU-Net, (e) PU-GAN-G, (f) PUGeo-Net, and (g) Proposed. }
        \label{fig:reconstruct}
\end{figure*}

Second, we demonstrated the effectiveness of the proposed method by surface reconstruction in Fig.~\ref{fig:reconstruct}. 
Specifically, we reconstructed surfaces from the 16$\times$ densified point clouds by different methods using Screened Poisson Sampling Reconstruction (SPSR)~\cite{kazhdan2013screened}, where point normals were computed by PCA with a neighborhood of 16 points. 
The identical parameters of SPSR were applied to all point clouds for fair comparisons. From Fig.~\ref{fig:reconstruct}, it can be observed that the surfaces directly reconstructed from the input sparse point clouds are deficient, while those from the upsampled point clouds exhibit richer geometry details. 
Compared with other methods, the reconstructed surfaces by the proposed method are closer to the ground truth surfaces. Especially, the proposed method can recover more details and better preserve the smoothness of smooth regions (see the closed-up regions), which are consistent with the observations from Fig.~\ref{fig:p2f}. 
\begin{table*}[th]
\centering
\caption{Distribution of point-wise CD values of upsampled 39 test shapes ($R=16$).}
\begin{tabular}{c||c c c c c c c}\Xhline{5\arrayrulewidth}
 Method &\makebox[4em]{$[0,1e^{-4}]$}   &\makebox[4em]{$[1e^{-4}, 5e^{-4}]$} &\makebox[4em]{$[5e^{-4}, 1e^{-3}]$}
 &\makebox[4em]{$[1e^{-3}, 5e^{-3}]$}
 &\makebox[4em]{$[5e^{-3}, 1e^{-2}]$} &\makebox[4em]{$[1e^{-2}, 5e^{-2}]$}
 &\makebox[4em]{$[5e^{-2}, 1e^{-1}]$}
\\\Xhline{2\arrayrulewidth}
PUGeo-Net   & 0.0004\% &  0.08\% &  0.52\% & 43.80\% &\bf{48.61\%} & \bf{6.98\%} & 0.00\% \\
Proposed  & \bf{0.001\%}  & \bf{0.12\%} & \bf{0.77\%}  & \bf{49.97\%} & 44.73\% & 4.43\% &\bf{0.0004\%} \\
\Xhline{5\arrayrulewidth} 
\end{tabular}
\label{table:pointwise} 
\end{table*}

\subsubsection{Analysis of the HD performance} From Table \ref{table:compare}, it is observed that that 
PUGeo-Net achieves slightly smaller HD values than the proposed method for all cases. In addition to the evaluation in terms of CD and HD which analyze the average and maximum point-wise errors, respectively, we also provided the distribution of the point-wise error in Table~\ref{table:pointwise} to help better understand the performance. From Table~\ref{table:pointwise}, it can be seen that the proposed method can generate more points with smaller point-wise errors than PUGeo-Net. However, compared with PUGeo-Net, the proposed method has 0.0004\% points with relatively large errors, which finally contribute to the relatively large HD value of the proposed method.

\begin{figure}[H]
\centering
        \subfigure[]{\includegraphics[width=1.1in]{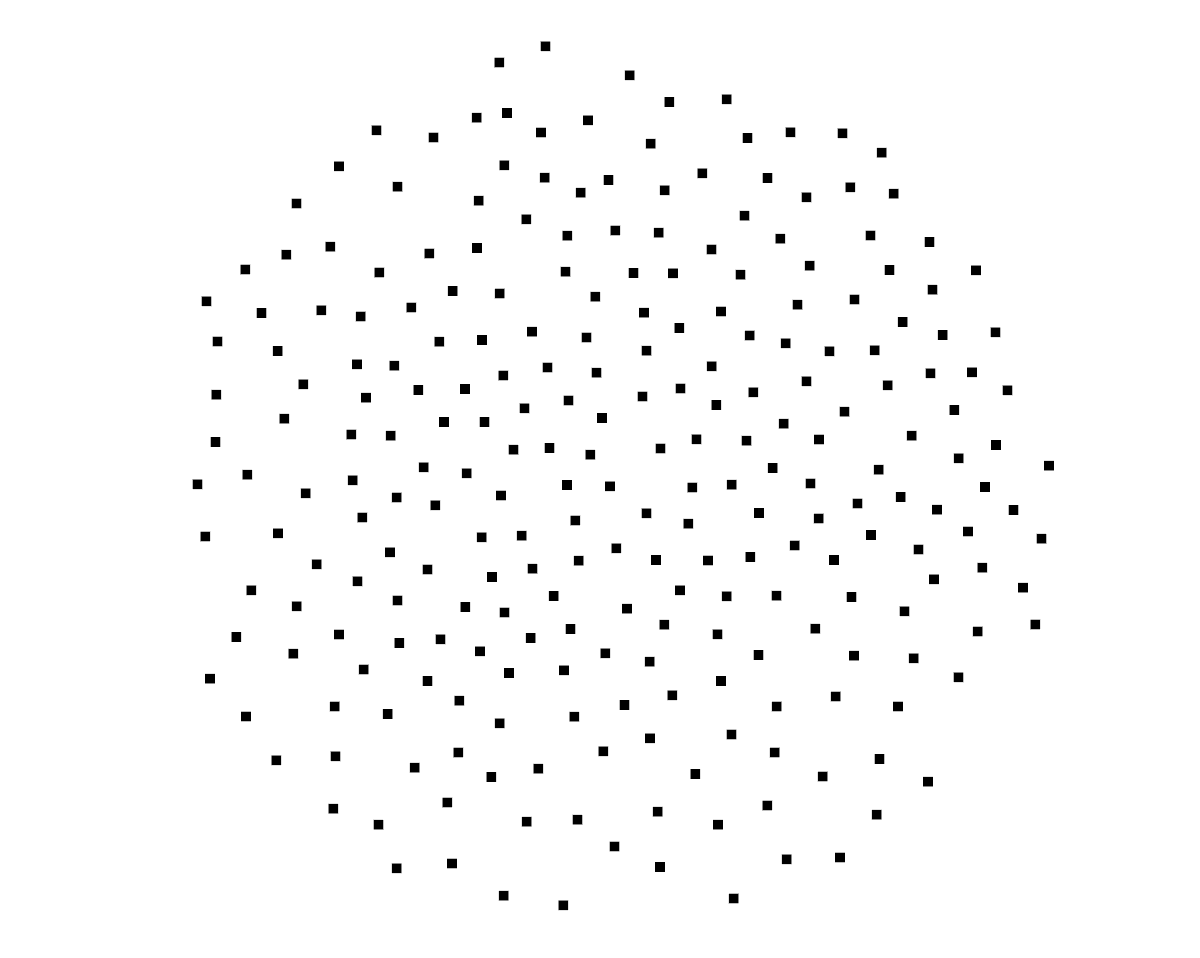}}
        \subfigure[]{\includegraphics[width=1.1in]{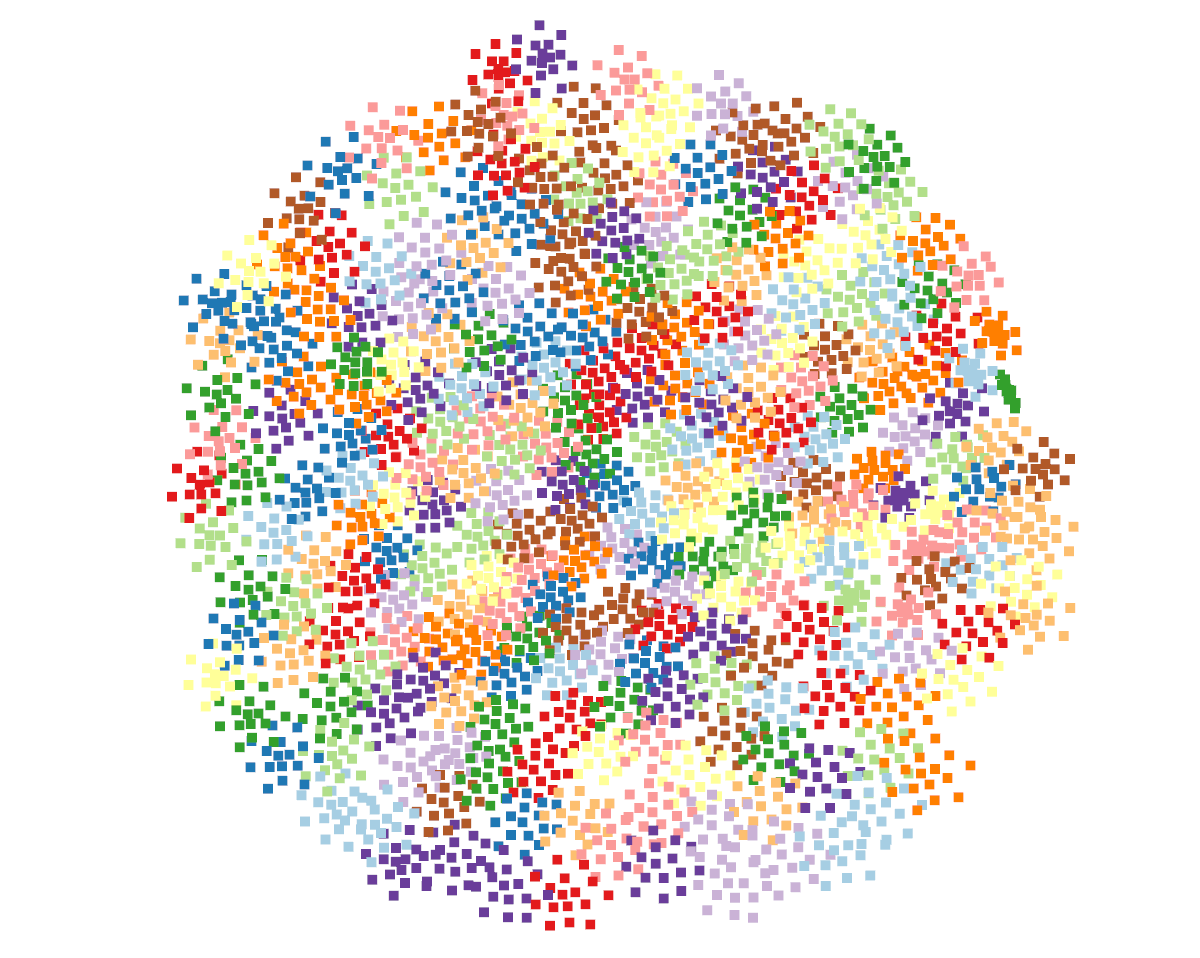}}
        \subfigure[]{\includegraphics[width=1.1in]{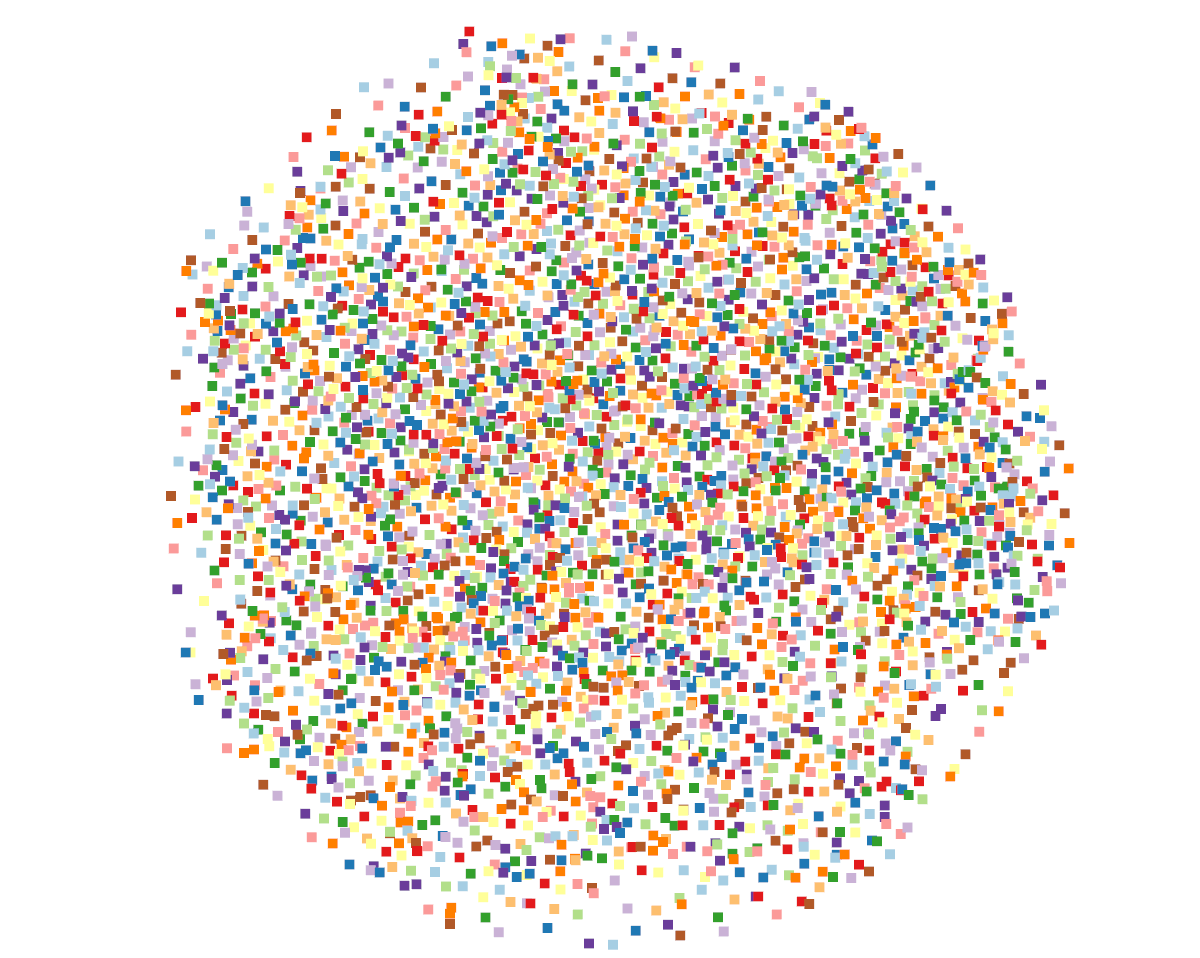}}
\caption{Visual comparison of the distribution of generated 3D points ($R = 16$). (a) Sparse input points; Upsampled results  by (b) PUGeo-Net and (c) Proposed.}
\label{fig:point_dist}
\end{figure}

\begin{figure*}[b!]
     \centering
    \subfigure[]{\includegraphics[width=1.2in]{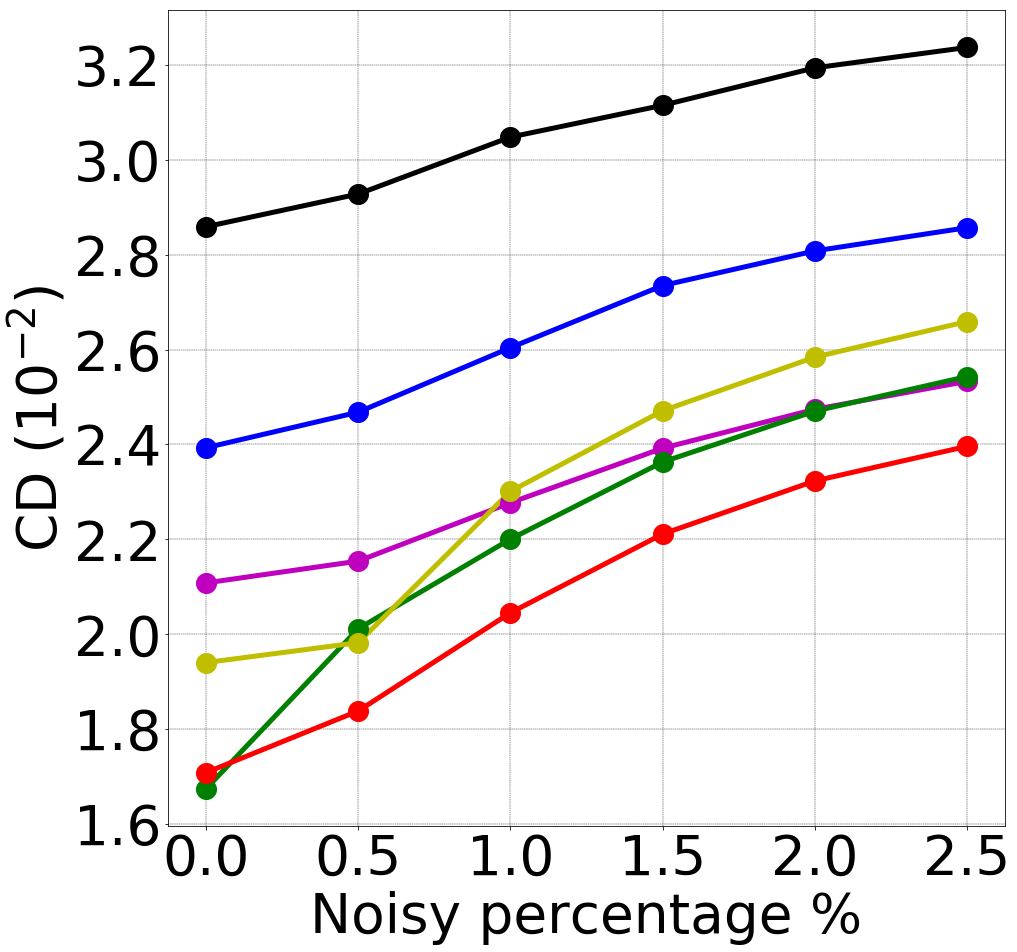}}
    \subfigure[]{\includegraphics[width=1.2in]{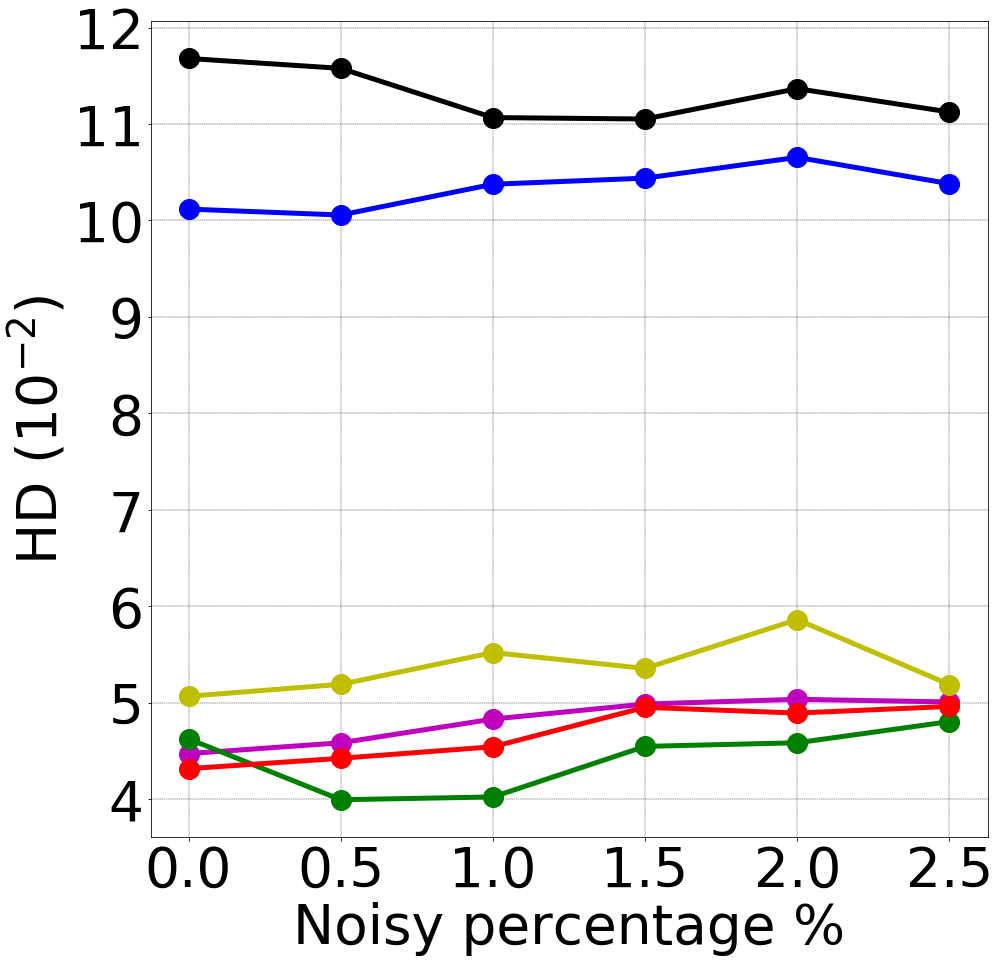}}
    \subfigure[]{\includegraphics[width=1.2in]{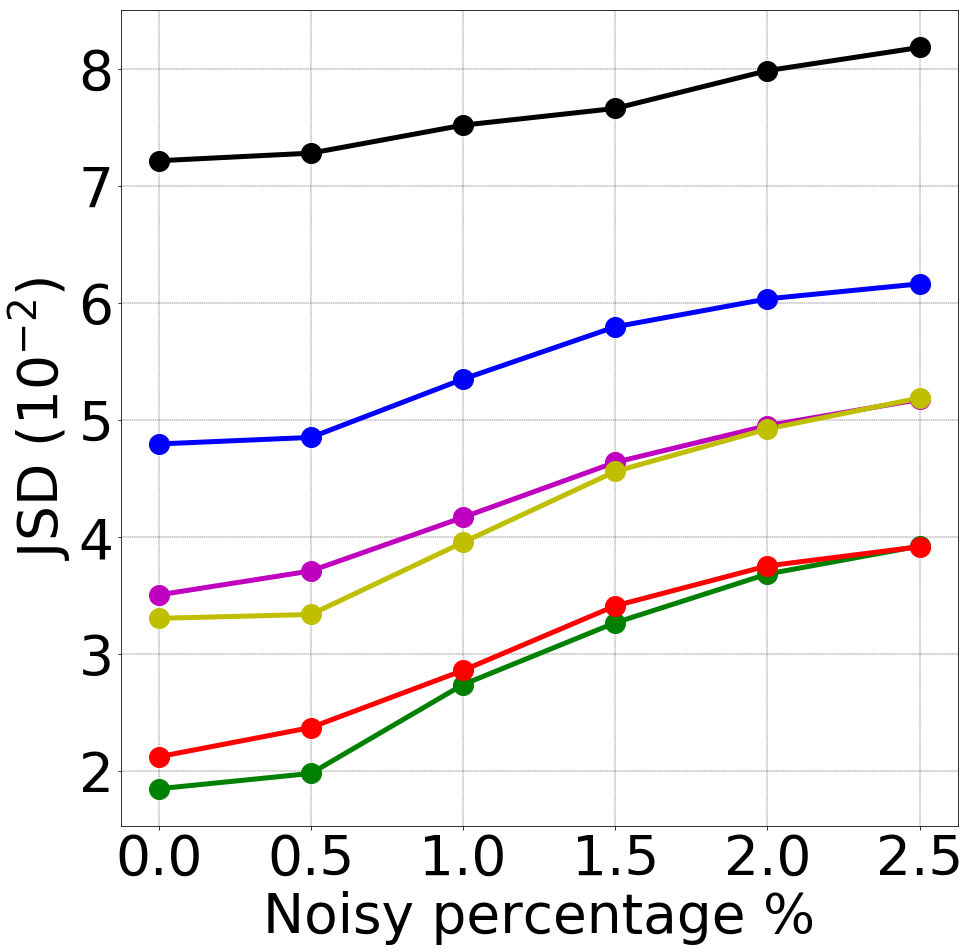}}
    \subfigure[]{\includegraphics[width=1.2in]{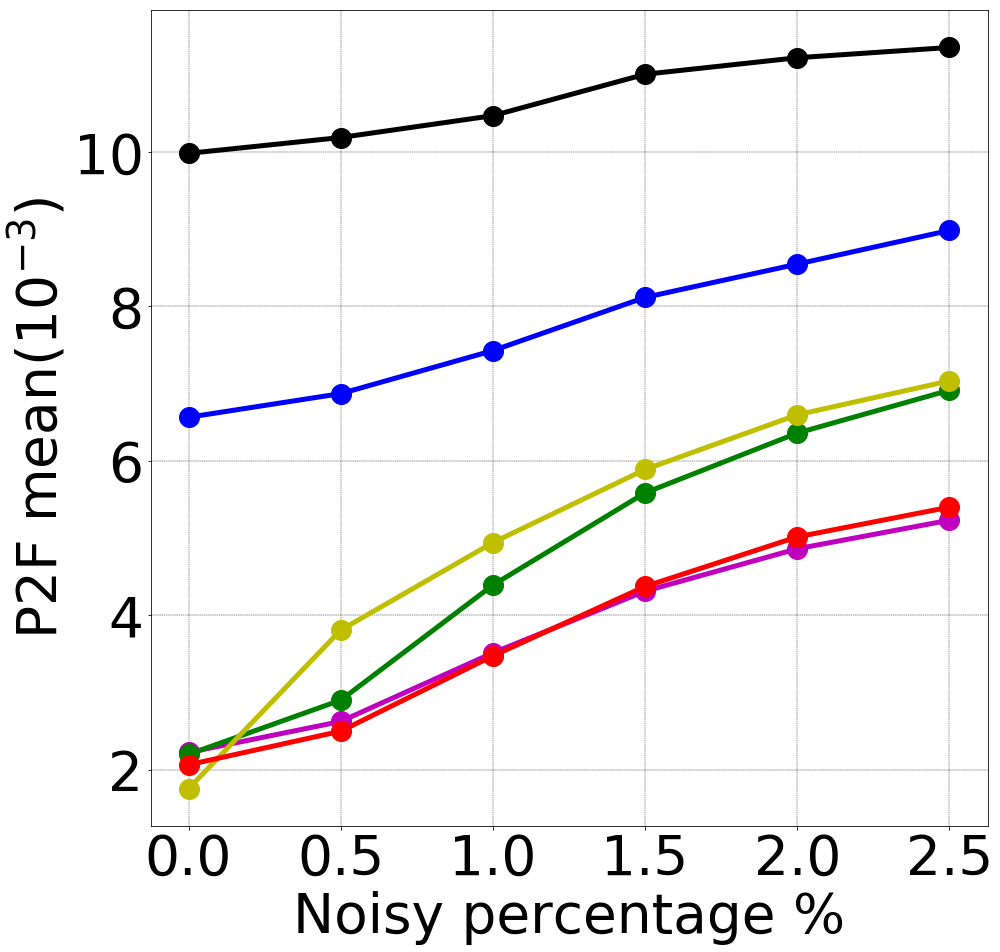}}
    \subfigure[]{\includegraphics[width=1.2in]{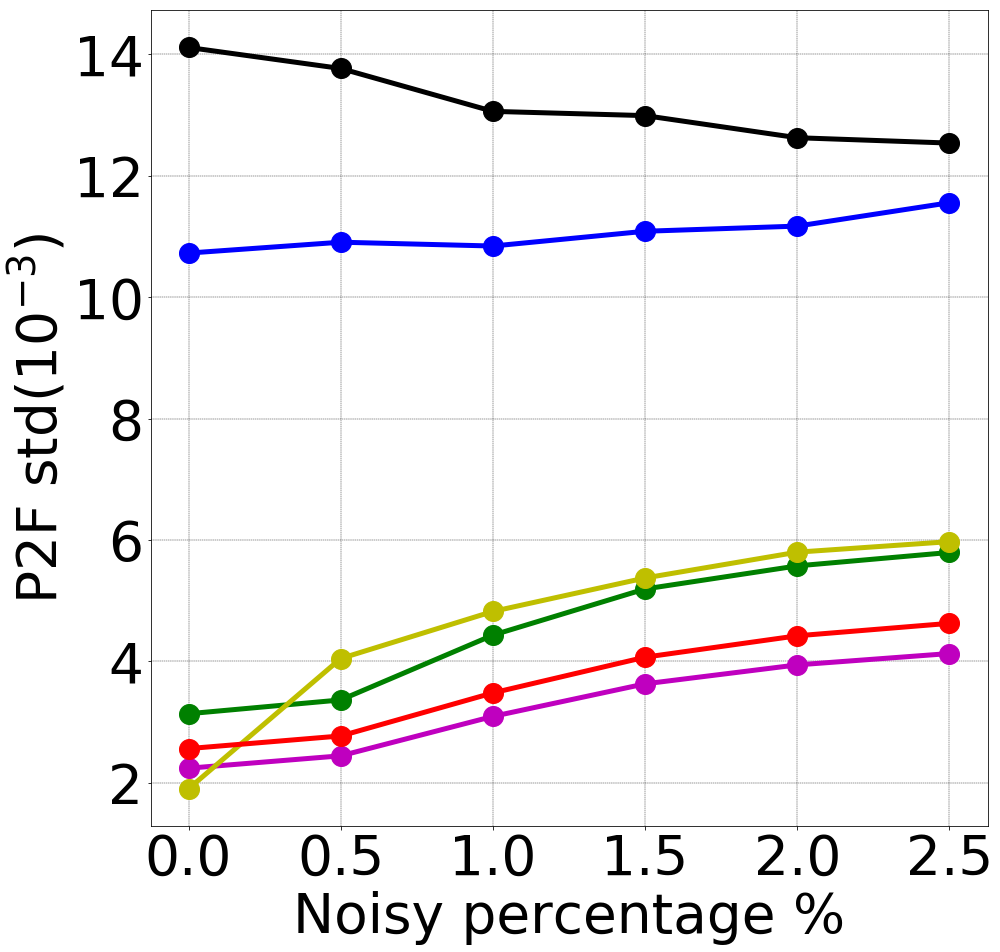}}
    \subfigure{\includegraphics[width=0.8in]{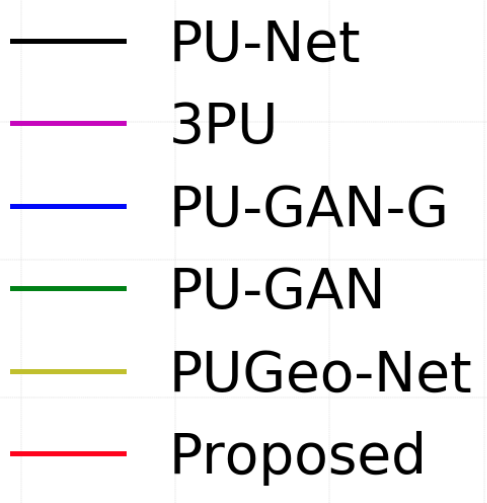}}
\caption{Quantitative comparisons on data with various levels of noise ($R=4$). (a) CD, (b) HD, (c) JSD, (d) P2F mean, (e) P2F std.}
\label{fig:noise_quant}
\end{figure*}

The advantage of PUGeo-Net over the proposed method in terms of the HD performance is credited to the 
different receptive fields used for generating dense points. Fig. \ref{fig:point_dist} depicts the distribution of generated dense points. Because PUGeo-Net predicts the local tangent plane for each sparse point, it tends to produce points around the original sparse input points. By contrast, the proposed method 
is based on the interpolation of $K$ local sparse points. Thus, the proposed method has a larger receptive field, which potentially results in a larger HD value, as HD measures the maximum point-wise error. 
The larger receptive field of the proposed method, on the other hand, helps to improve the algorithm's robustness. As illustrated in Fig. \ref{fig:noise_quant}, our method can retain more stable performance when the noise level increases compared to the other approaches.

\begin{figure*}[b!]
     \centering
    \subfigure[]{\includegraphics[width=0.8in]{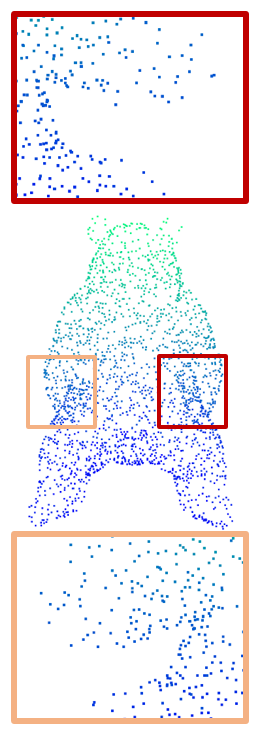}}
    \subfigure[]{\includegraphics[width=0.8in]{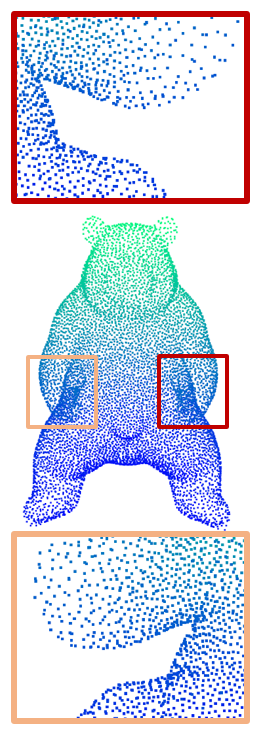}}
    \subfigure[]{\includegraphics[width=0.8in]{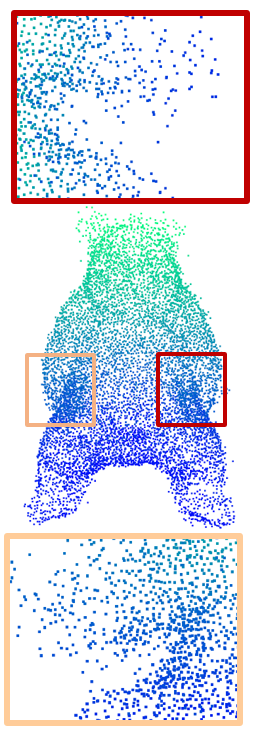}}
    \subfigure[]{\includegraphics[width=0.8in]{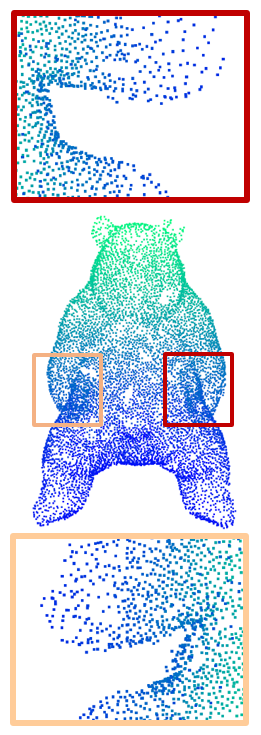}}
    \subfigure[]{\includegraphics[width=0.8in]{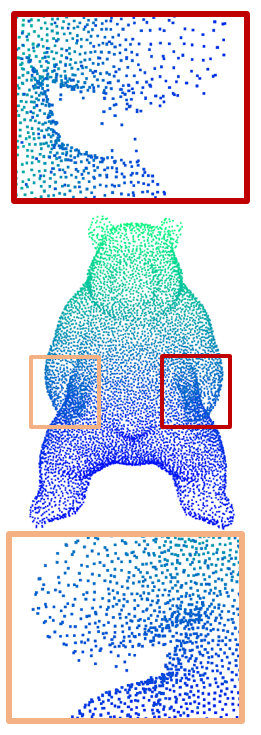}}
    \subfigure[]{\includegraphics[width=0.8in]{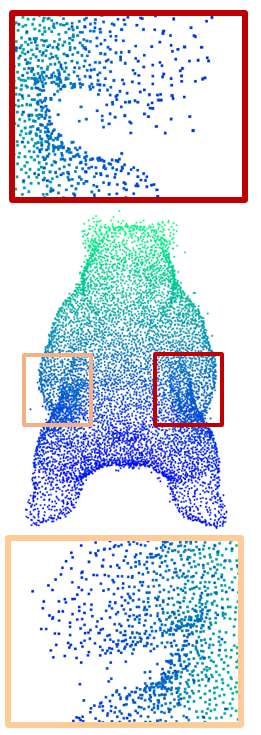}}
    \subfigure[]{\includegraphics[width=0.8in]{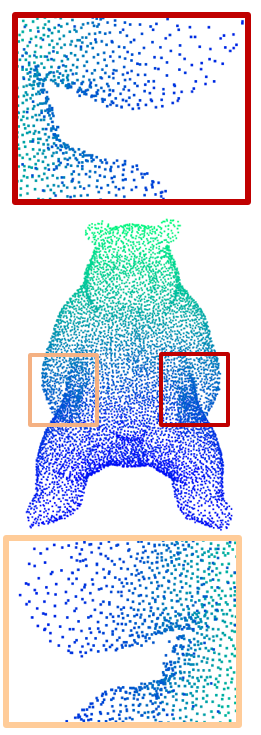}}
    \subfigure[]{\includegraphics[width=0.8in]{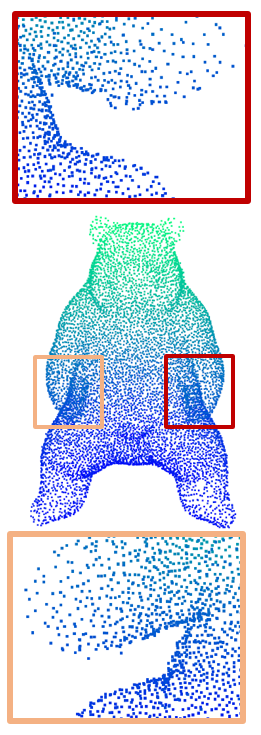}}
\caption{Visual comparisons on non-uniform inputs ($R=4$). 
(a) Non-uniform sparse point clouds with 2,048 points each; (b) Ground-truth dense point clouds with 8,192 points each; The upsampled results yielded by (c) PU-Net, (d) 3PU-Net, (e) PU-GAN, (f) PU-GAN-G, (g) PUGeo-Net, and (h) Proposed.
}
\label{fig:non-uniform}
\end{figure*}

\begin{table}[t!]
\centering
\caption{Quantitative comparisons of different methods applied to non-uniform point clouds ($R=4$).}
\begin{tabular}{c||c c c c c}\Xhline{5\arrayrulewidth}
 Method  & \makebox[3em]{CD} & \makebox[3em]{HD}  &\makebox[3em]{JSD} & \makebox[3em]{P2F mean} & \makebox[3em]{P2F std} \\
        & \makebox[3em]{$(10^{-2})$} & \makebox[3em]{$(10^{-2})$}  &\makebox[3em]{$(10^{-2})$} & \makebox[3em]{$(10^{-3})$} & \makebox[3em]{$(10^{-3})$}  \\
\Xhline{2\arrayrulewidth}
PU-Net \cite{yu2018pu} & 2.859 & 11.680 & 7.214 & 9.983 & 14.109  \\
3PU-Net\cite{yifan2019patch} &  2.108 & 4.471  & 3.504 & 2.226 & 2.243\\
PU-GAN \cite{li2019pu}  & \bf{1.674} & 4.624 & \bf{1.848} & 2.201 & 3.140  \\
PU-GAN-G \cite{li2019pu}  & 2.393 & 10.118 & 4.794 & 6.566  & 10.724  \\
PUGeo-Net~\cite{qian2020pugeo}  & 1.940 & 5.065 & 3.304 & \bf{1.756}  & \bf{1.905}  \\\Xhline{\arrayrulewidth}
Proposed &   1.708 & \bf{4.315} & 2.121 & 2.066 & 2.566  \\
\Xhline{5\arrayrulewidth} 
\end{tabular}
\label{table:compare_nonuni} 
\end{table}

\subsection{Evaluation on Non-uniform Data}
We further evaluated different methods for upsampling non-uniform point clouds. 
During the testing phase, we applied Poisson disk sampling to 39 testing mesh models to generate point clouds with 8192 points each as ground-truth data, then randomly sampled the ground-truth data to generate point clouds with 2,048 points each as the non-uniform input.

\subsubsection{Quantitative comparison} Table~\ref{table:compare_nonuni} quantitatively compares different methods under  
$4\times$ upsampling, where it can be seen that 
the proposed method achieves the best performance in terms of CD, HD and JSD among the non-GAN-based methods, including PU-Net, 3PU-Net, PU-GAN-G, and PUGeo-Net. Meanwhile, the GAN-based approach PU-GAN has superior performance in terms of CD and JSD, but has relatively poor performance in terms of P2F.
In addition, the comparison of the NUC in Table~\ref{table:uniformity}  
demonstrates that the proposed method can generate more uniform point clouds when compared with the non-GAN-based methods, 
 especially for relatively small disk area percentages. These observations demonstrate the advantages and generality of the proposed method on non-uniform data.

\begin{table}[t!]
\centering
\caption{Quantitative comparisons of the uniformity of upsampled point clouds from non-uniform data by different methods ($R=4$) under various disk areas $p$. 
The lower the NUC value is, the points are more uniformly distributed in mesh.}
\begin{tabular}{c||c c c c c}\Xhline{5\arrayrulewidth}
 Method &\makebox[3em]{$0.4\%$}  & \makebox[3em]{$0.6\%$} & \makebox[3em]{$0.8\%$ }  &\makebox[3em]{$1.0\%$ } & \makebox[3em]{$1.2\%$} \\\Xhline{2\arrayrulewidth}
Input &1.020 &	0.963&	0.865&	0.772&	0.699\\
PU-Net \cite{yu2018pu} &0.223&	0.199&	0.189&	0.185&	0.181\\
3PU-Net\cite{yifan2019patch} & 0.186&	0.159&	0.145&	0.132&	0.127\\
PU-GAN \cite{li2019pu}  & 0.104 & 0.096 & 0.092 & 0.090 & 0.090 \\
PU-GAN-G \cite{li2019pu}  &0.193&	0.177&	0.166&	0.160&	0.156\\
PUGeo-Net \cite{qian2020pugeo}  &0.124&	0.111&	0.103&	0.101&	0.100\\\Xhline{\arrayrulewidth} 
Proposed &0.118&	0.108	&0.105&	0.103&	0.103\\
\Xhline{5\arrayrulewidth} 
\end{tabular}
\label{table:uniformity} 
\end{table}
\subsubsection{Visual comparison} Fig.~\ref{fig:non-uniform} visualizes the upsampled results by different methods, where it can be observed that PU-Net, PU-GAN and PU-GAN-G tend to generate outliers, 
and 3PU-Net fails to maintain the uniformity of the upsampled point clouds. By contrast, the points of upsampled point clouds by our method are more uniformly distributed 
closer to the ground-truth ones. 

\subsection{Evaluation on Noisy Data 
}

\begin{figure*}[th]
     \centering
    \subfigure{\includegraphics[width=0.8in]{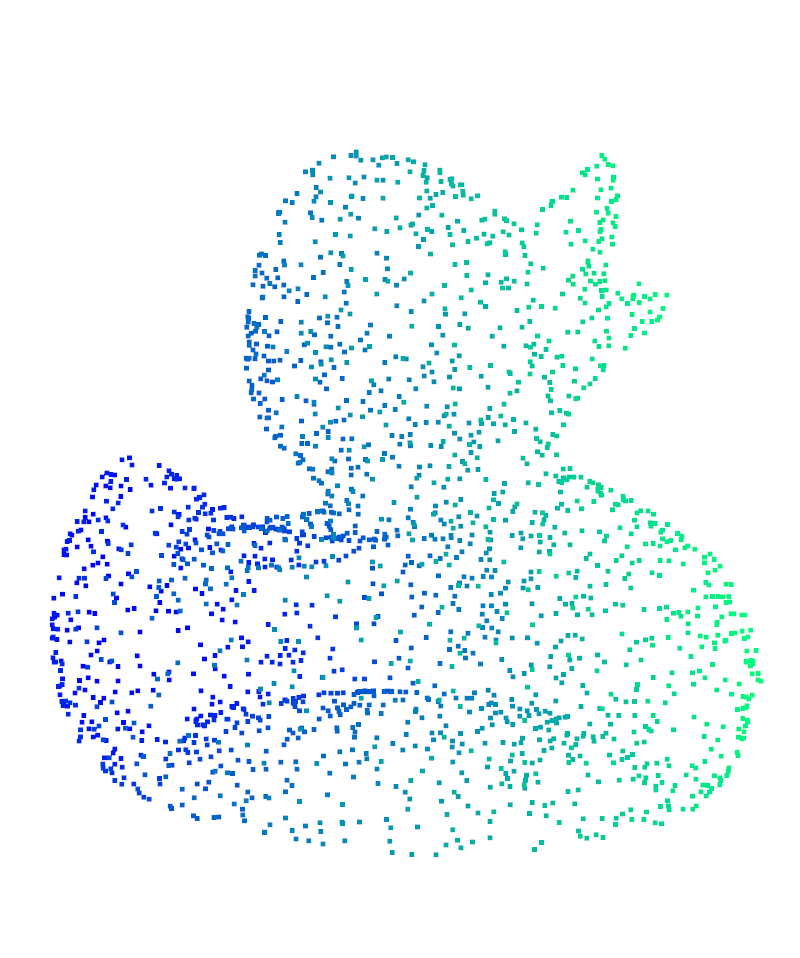}}
    \subfigure{\includegraphics[width=0.8in]{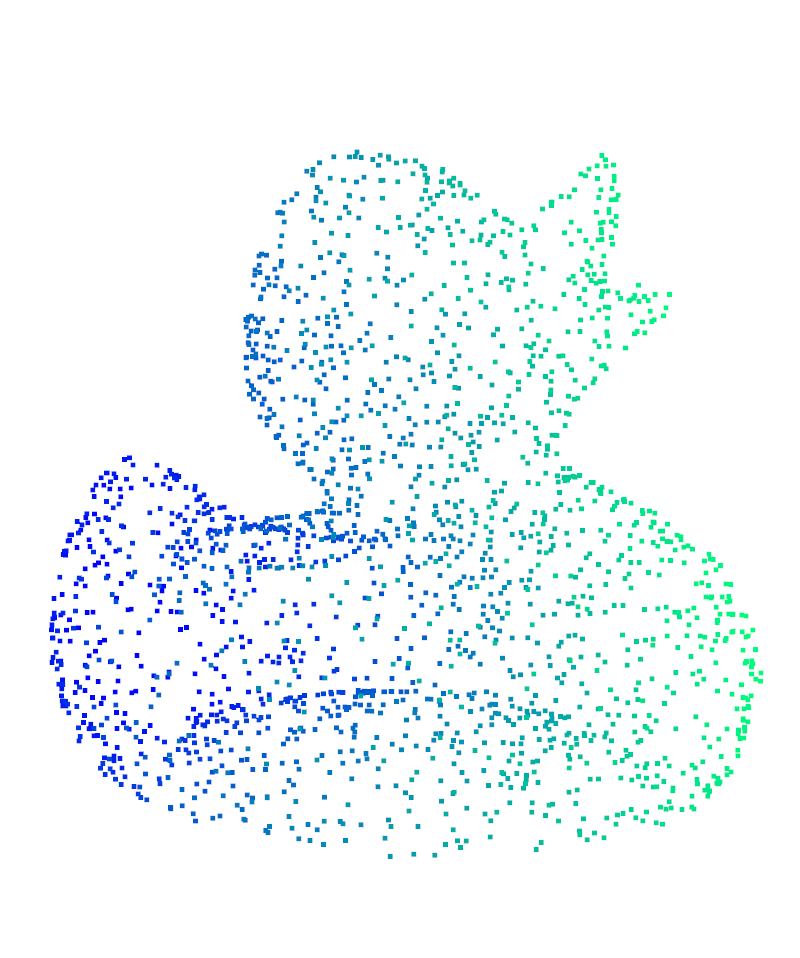}}
    \subfigure{\includegraphics[width=0.8in]{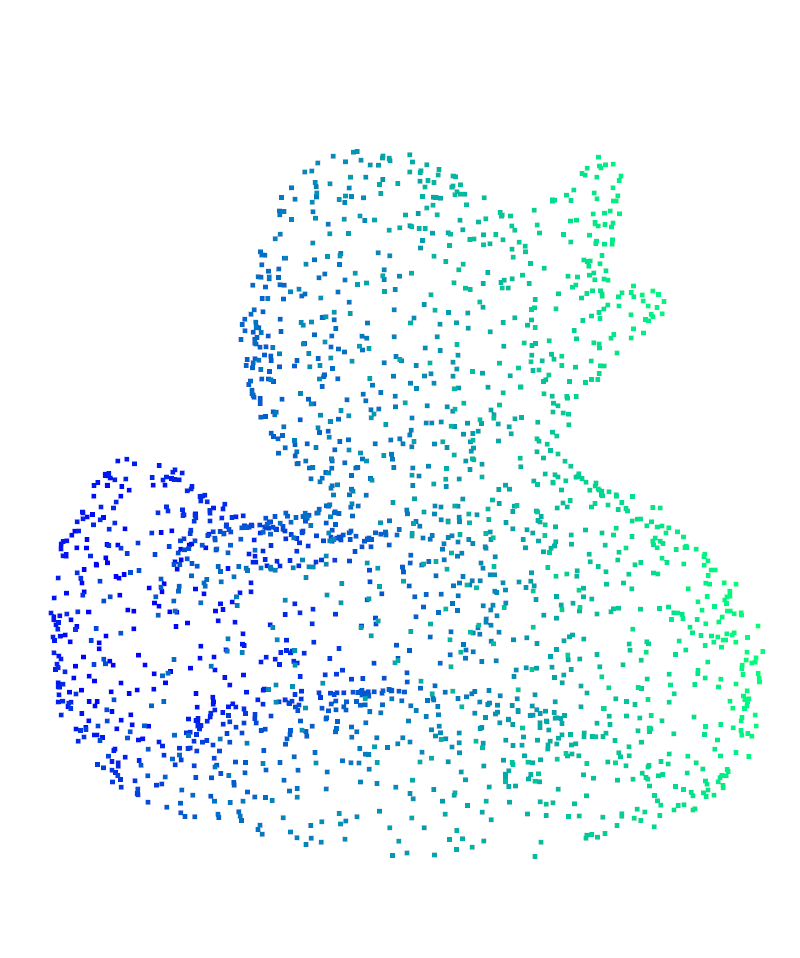}}
    \subfigure{\includegraphics[width=0.8in]{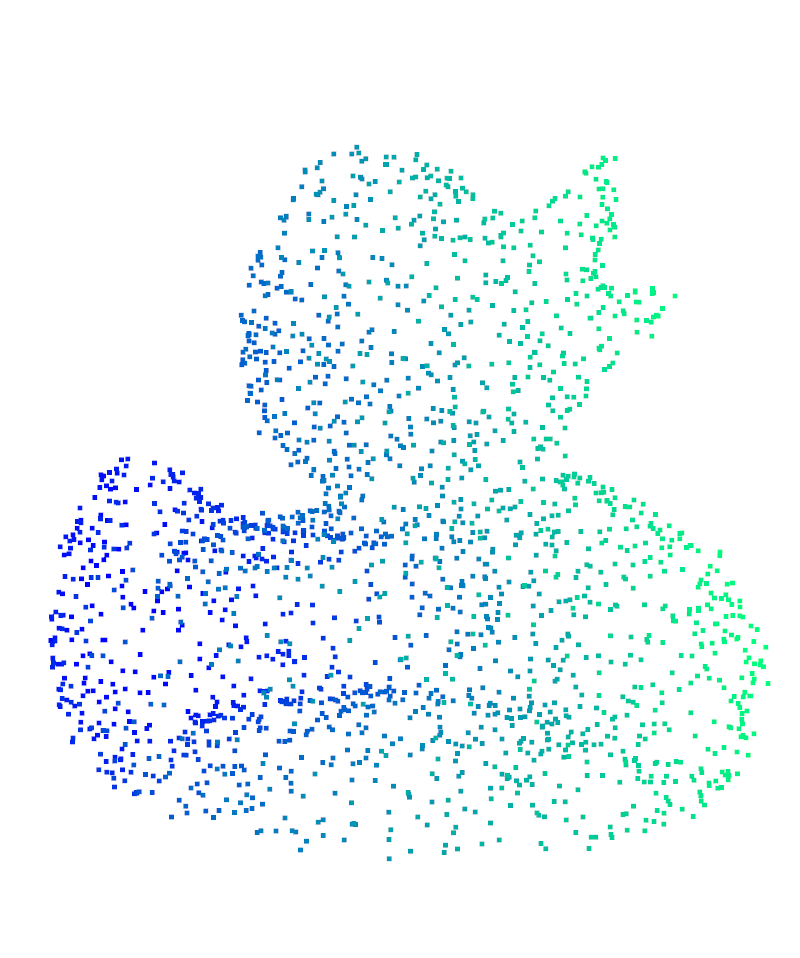}}
        \subfigure{\includegraphics[width=0.8in]{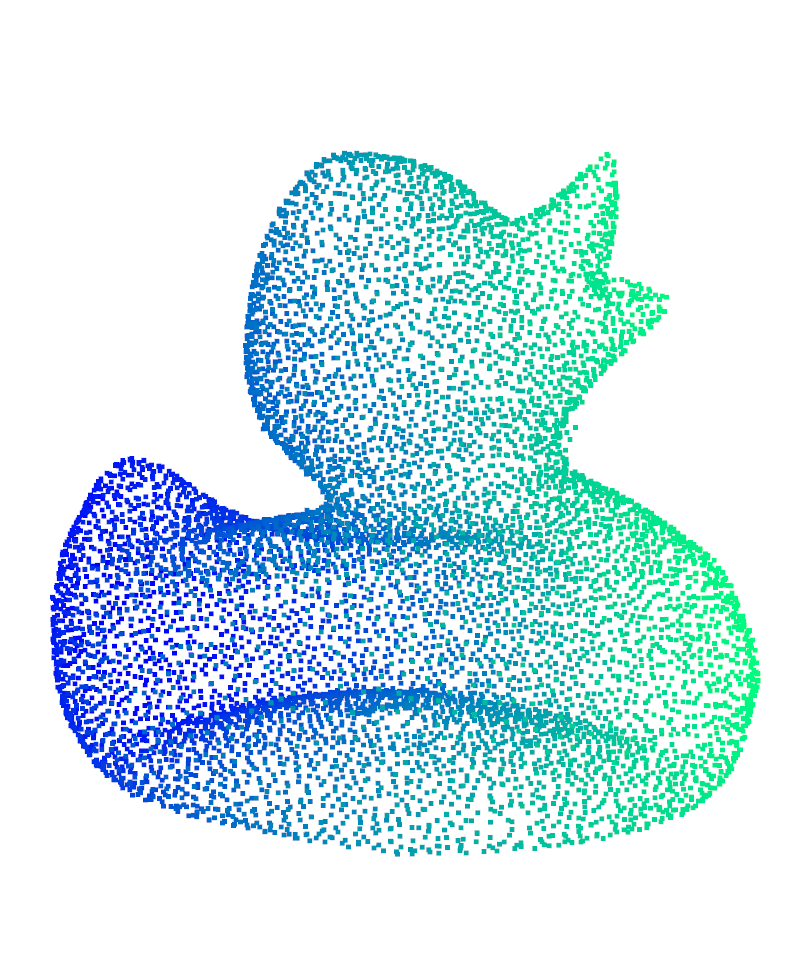}}
    \subfigure{\includegraphics[width=0.8in]{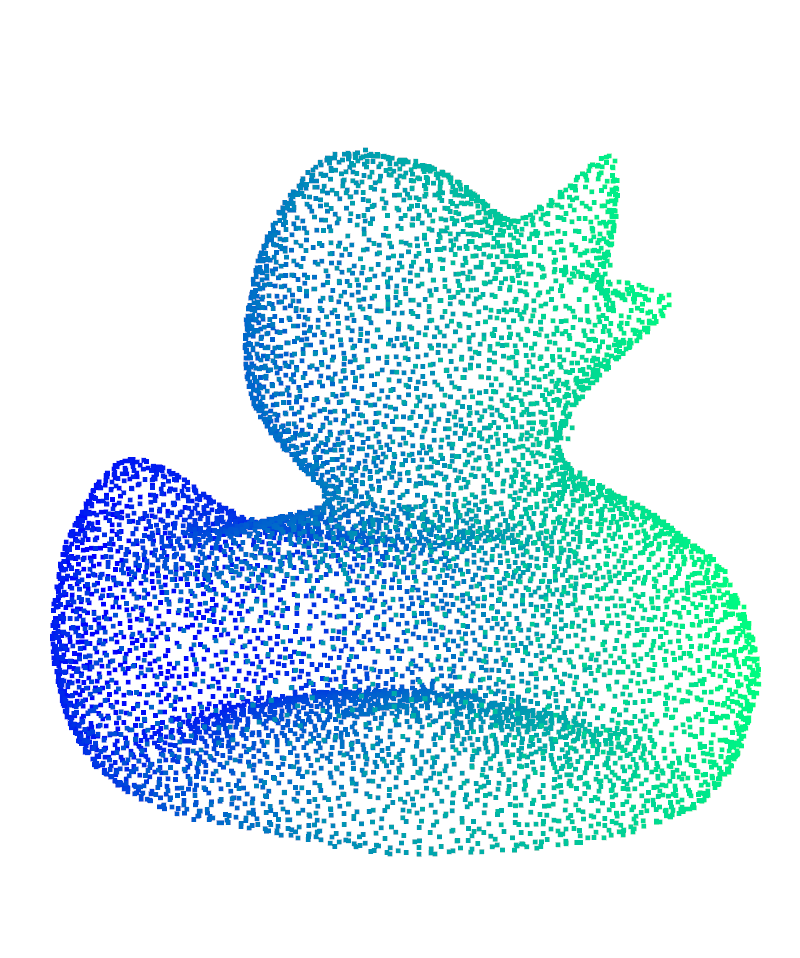}}
    \subfigure{\includegraphics[width=0.8in]{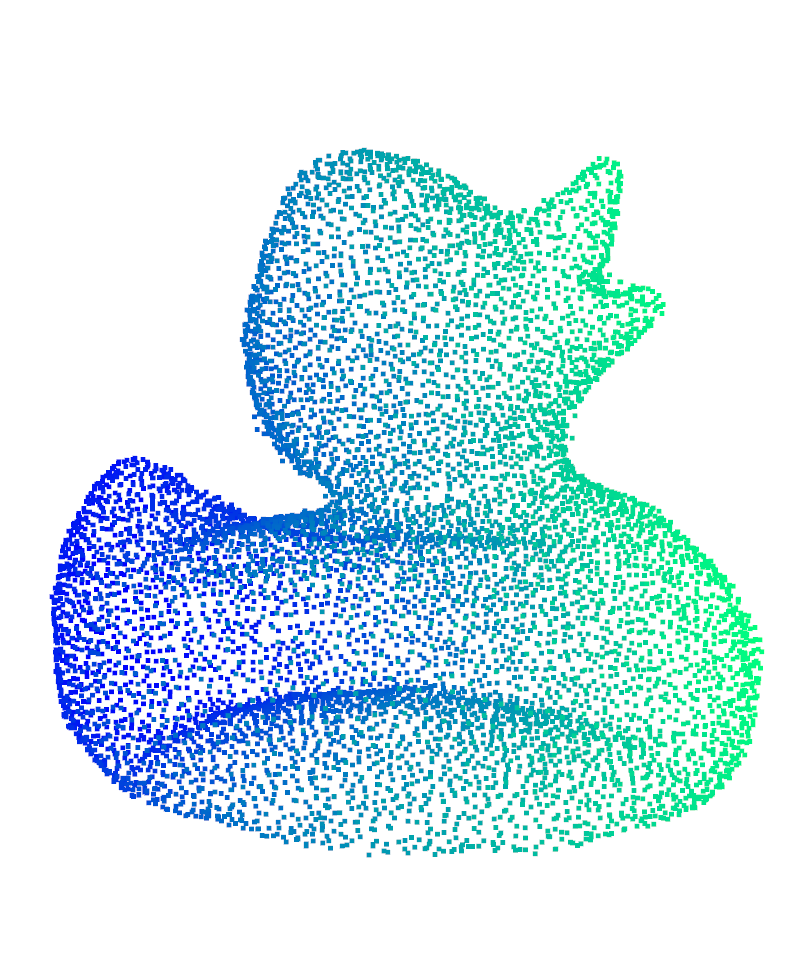}}
    \subfigure{\includegraphics[width=0.8in]{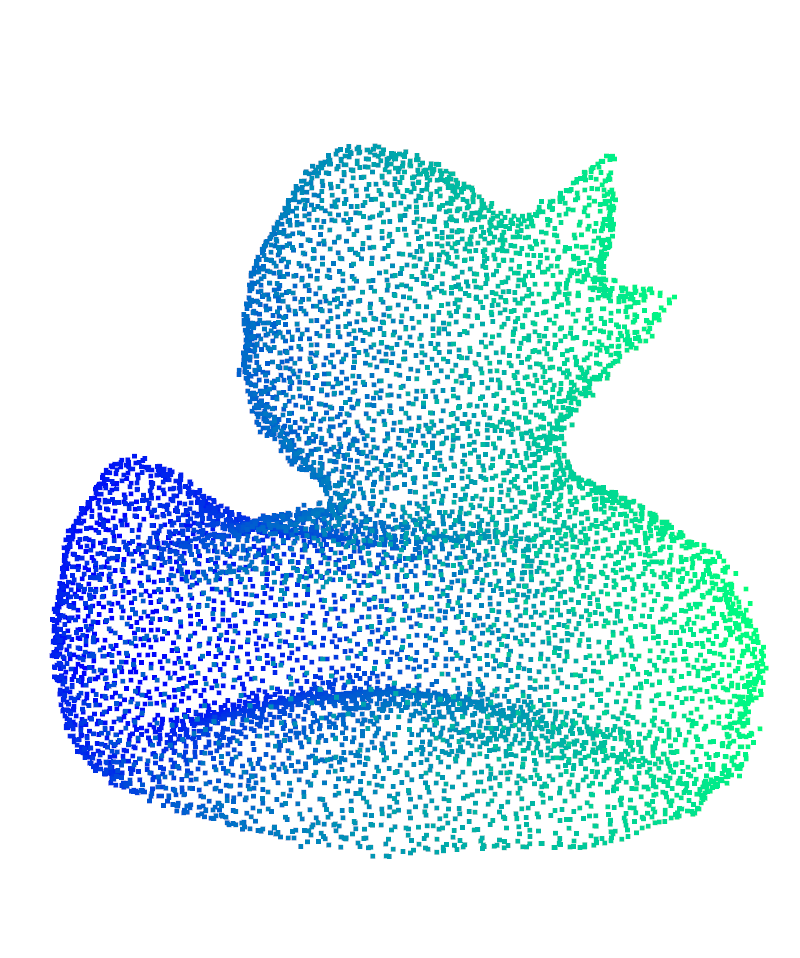}}
                  \renewcommand{\thesubfigure}{{\scriptsize (a1)}}
    \subfigure[]{\includegraphics[width=0.8in]{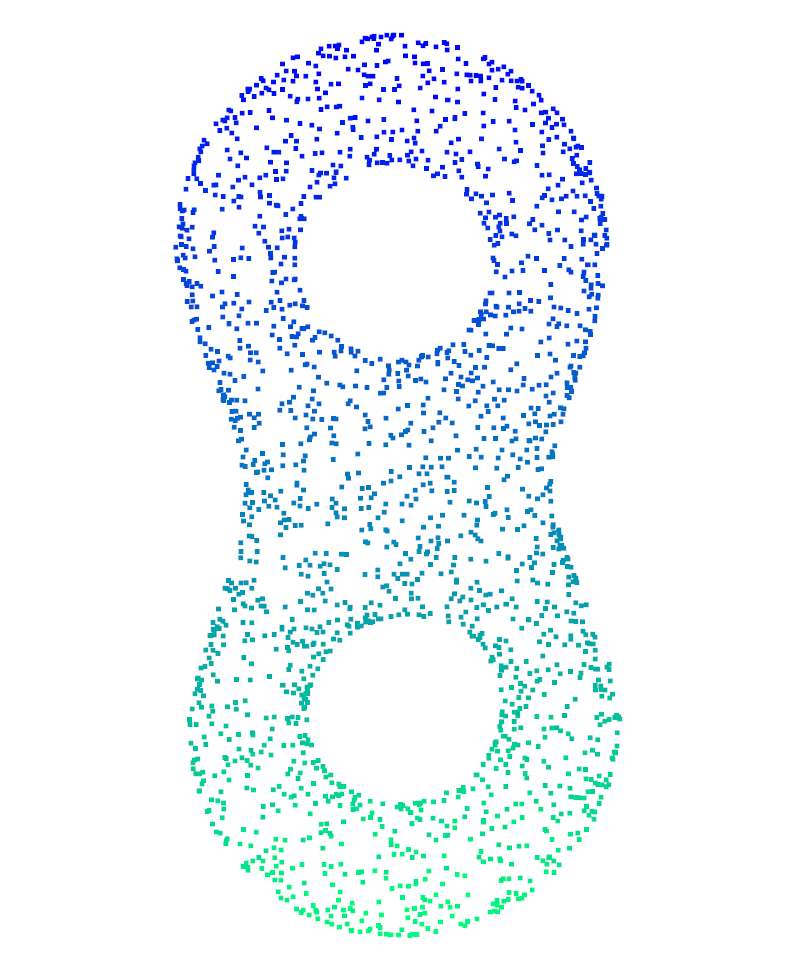}}
                  \renewcommand{\thesubfigure}{{\scriptsize (b1)}}
    \subfigure[]{\includegraphics[width=0.8in]{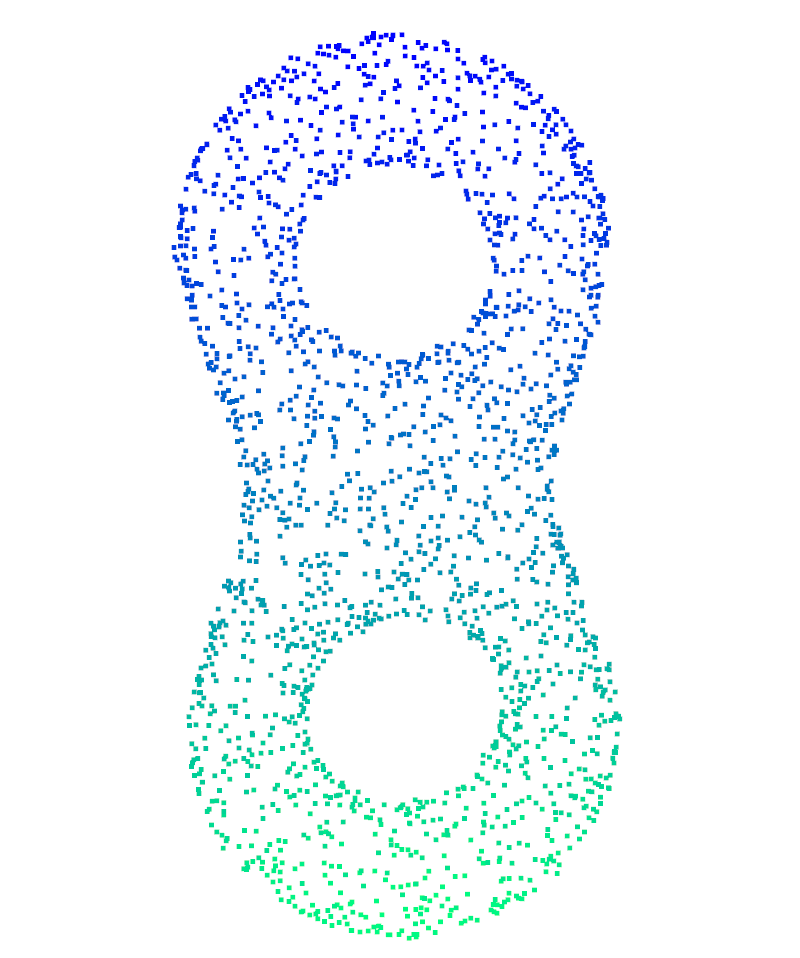}}
                  \renewcommand{\thesubfigure}{{\scriptsize (c1)}}
    \subfigure[]{\includegraphics[width=0.8in]{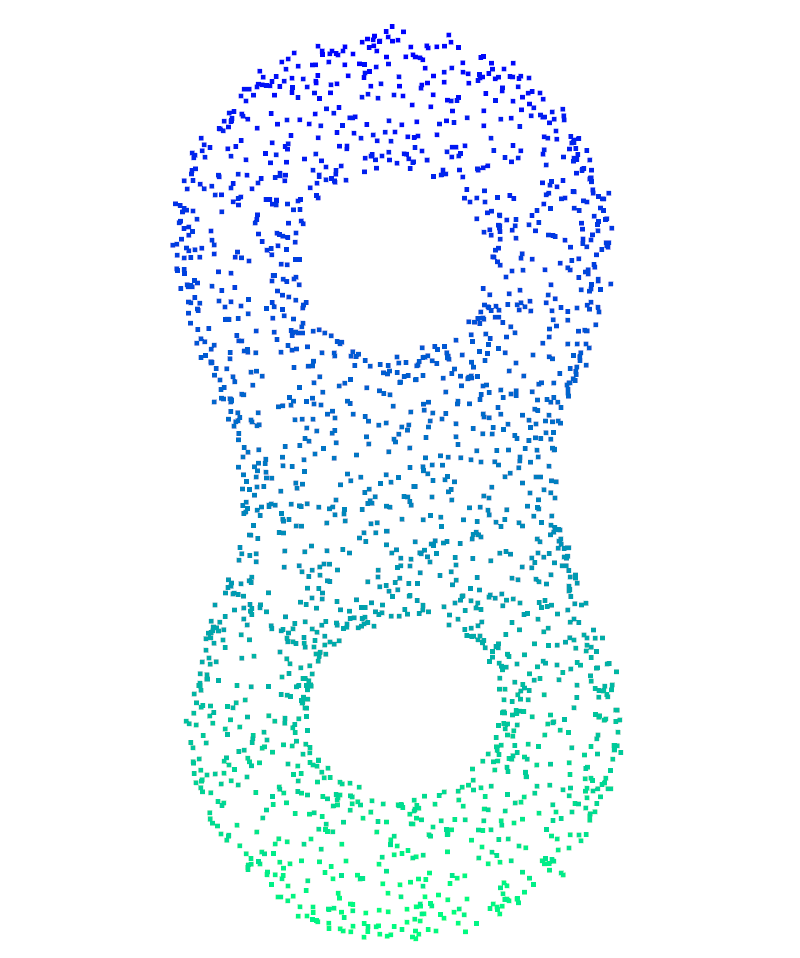}}
                  \renewcommand{\thesubfigure}{{\scriptsize (d1)}}
    \subfigure[]{\includegraphics[width=0.8in]{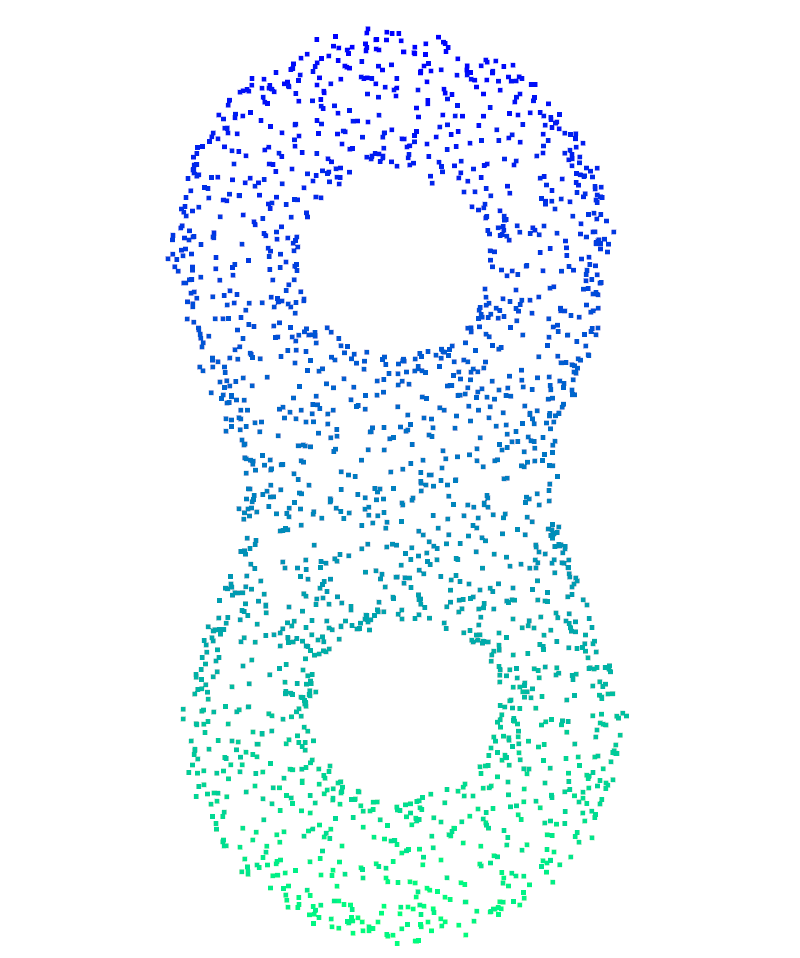}}
                  \renewcommand{\thesubfigure}{{\scriptsize (a2)}}
        \subfigure[]{\includegraphics[width=0.8in]{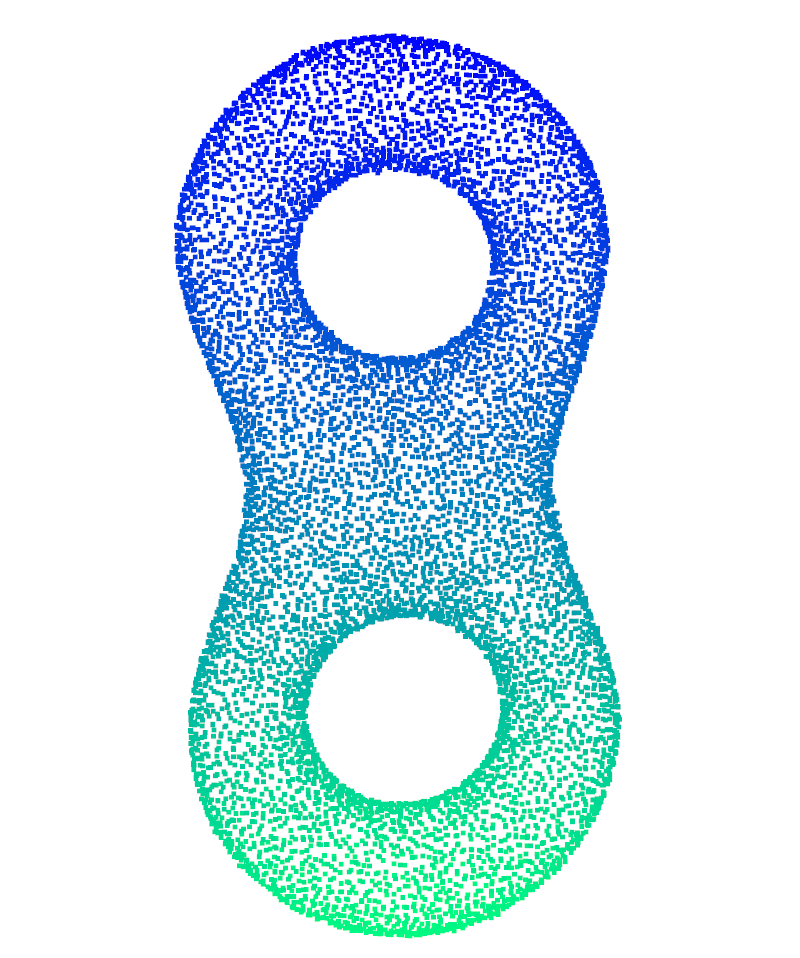}}
                      \renewcommand{\thesubfigure}{{\scriptsize (b2)}}
    \subfigure[]{\includegraphics[width=0.8in]{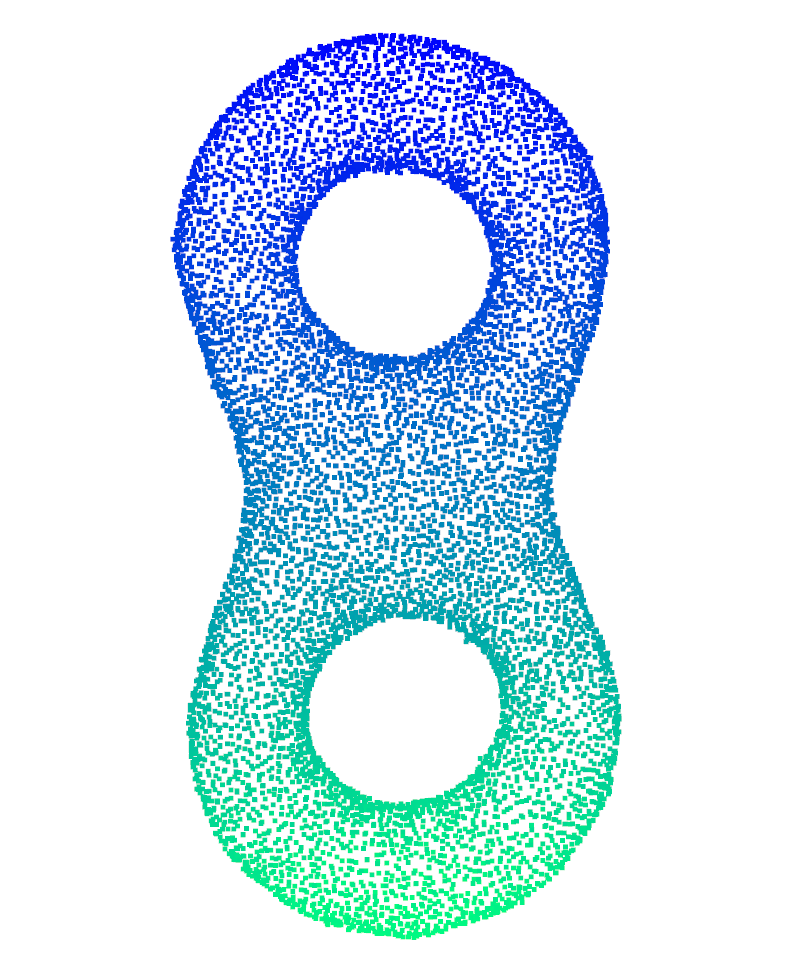}}
                  \renewcommand{\thesubfigure}{{\scriptsize (c2)}}
    \subfigure[]{\includegraphics[width=0.8in]{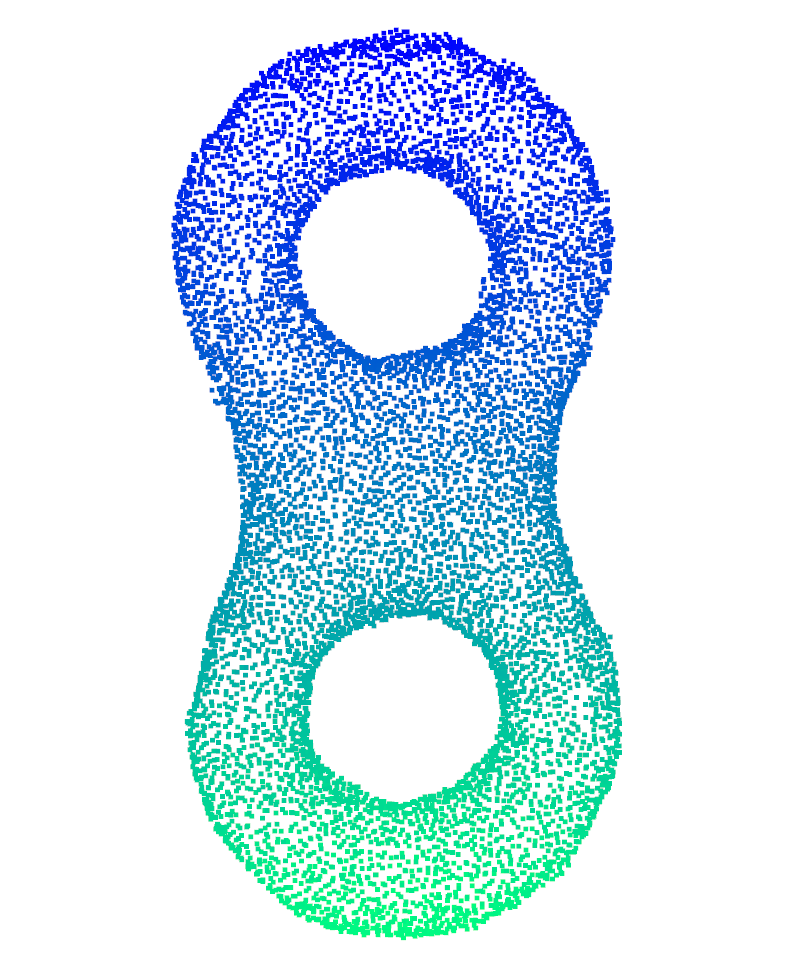}}
                  \renewcommand{\thesubfigure}{{\scriptsize (d2)}}
    \subfigure[]{\includegraphics[width=0.8in]{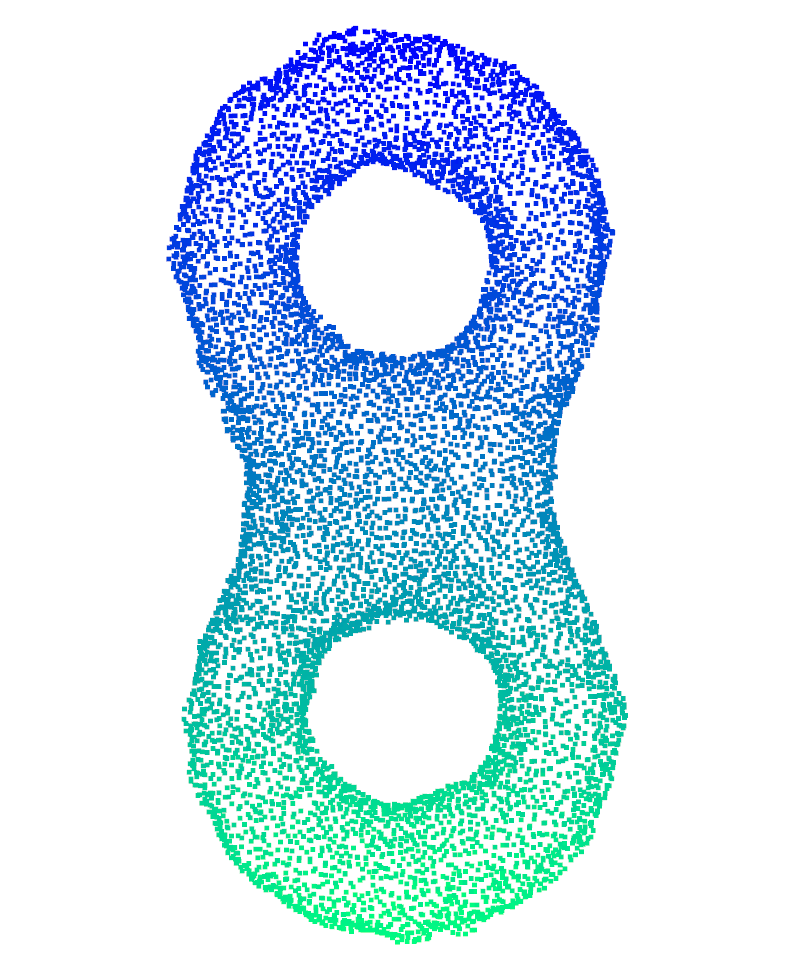}}
\caption{Visual results of the proposed method on noisy data. Left: (a1), (b1), (c1), and (d1) are the sparse inputs with 0\%, 0.5\%, 1.5\%, and 2.5\% Gaussian noise, respectively. 
Right: (a2), (b2), (c2), and (d2) are the 4$\times$ upsampled results from (a1), (b1), (c1), and (d1), respectively.
}
\label{fig:noise_vis}
\end{figure*}

\begin{figure}[b!]
     \centering
      \renewcommand{\thesubfigure}{{\scriptsize (a1)}}
        \subfigure[]{\includegraphics[width=0.65in]{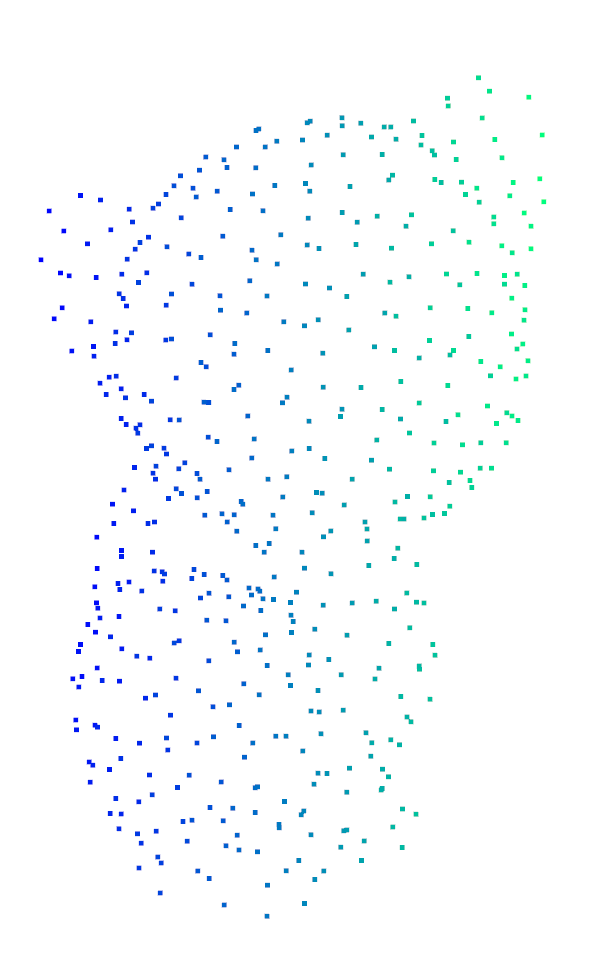}}
              \renewcommand{\thesubfigure}{{\scriptsize (b1)}}
        \subfigure[]{\includegraphics[width=0.65in]{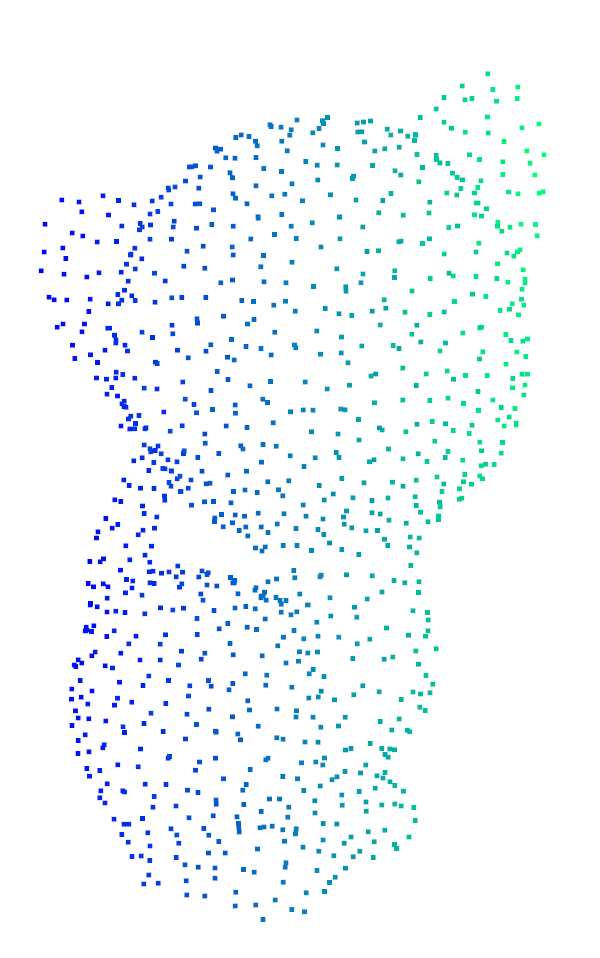}}
              \renewcommand{\thesubfigure}{{\scriptsize (c1)}}
        \subfigure[]{\includegraphics[width=0.65in]{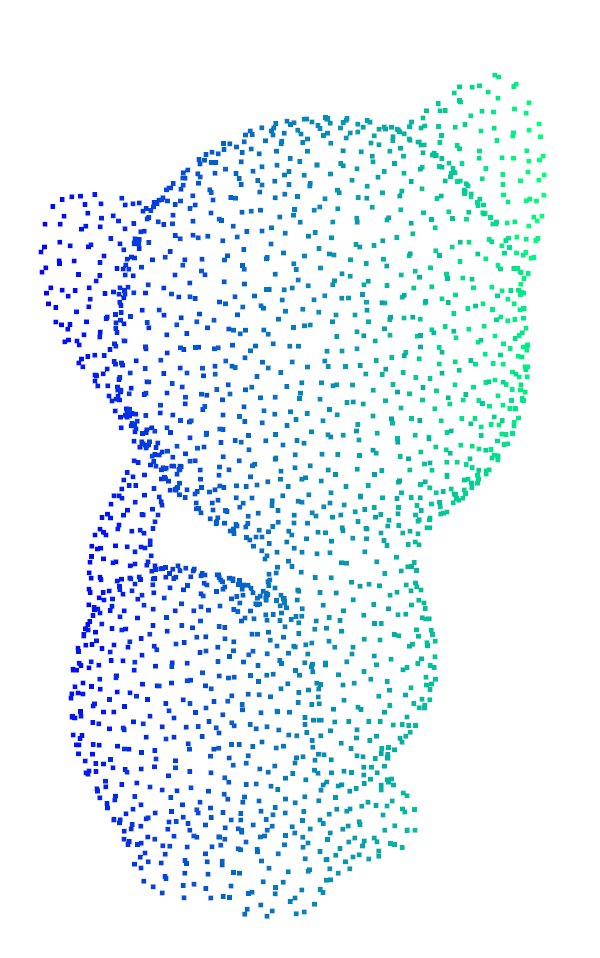}}
              \renewcommand{\thesubfigure}{{\scriptsize (d1)}}
        \subfigure[]{\includegraphics[width=0.65in]{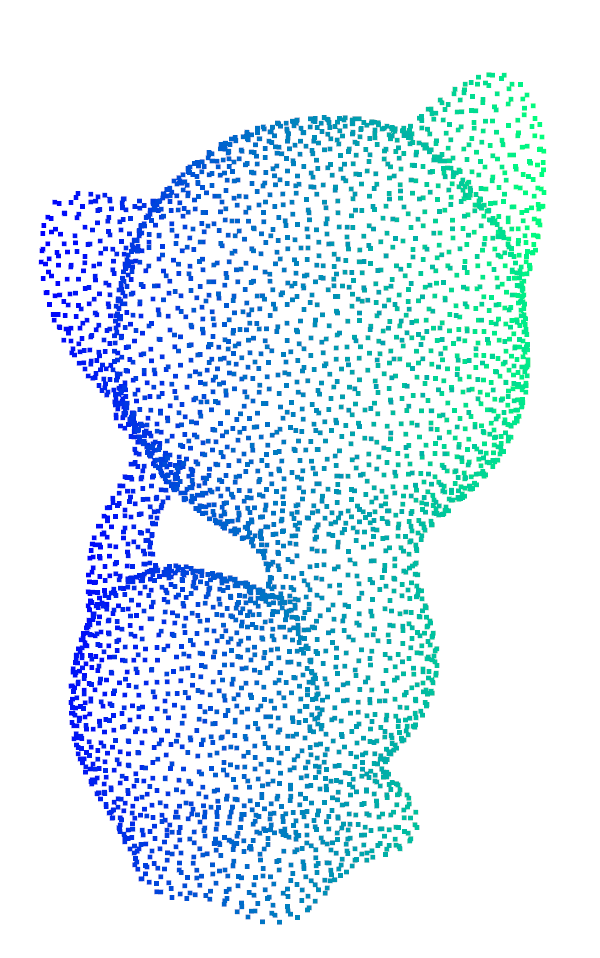}}
              \renewcommand{\thesubfigure}{{\scriptsize (a2)}}
       \subfigure[]{\includegraphics[width=0.65in]{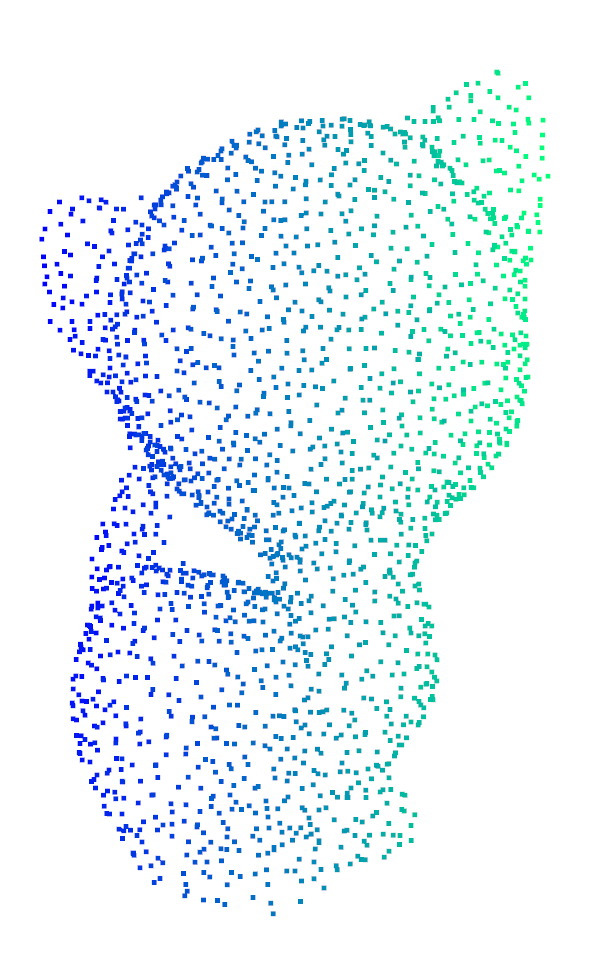}}
             \renewcommand{\thesubfigure}{{\scriptsize (b2)}}
        \subfigure[]{\includegraphics[width=0.65in]{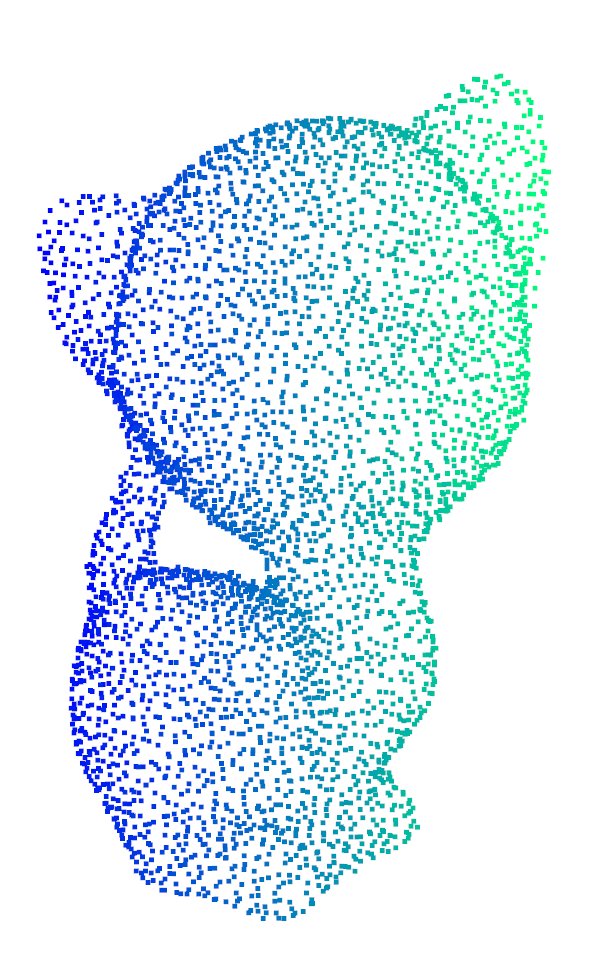}}
              \renewcommand{\thesubfigure}{{\scriptsize (c2)}}
        \subfigure[]{\includegraphics[width=0.65in]{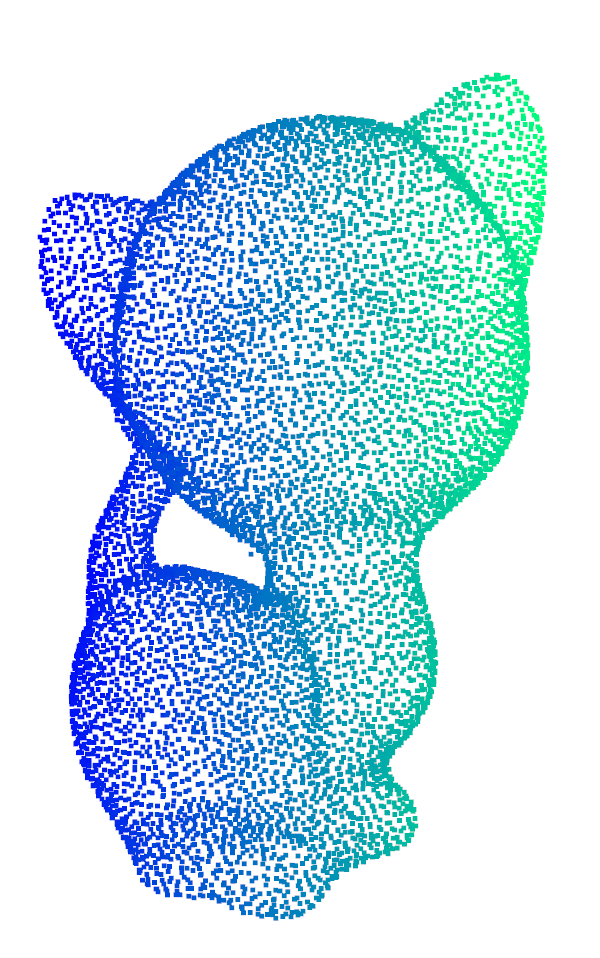}}
              \renewcommand{\thesubfigure}{{\scriptsize (d2)}}
        \subfigure[]{\includegraphics[width=0.65in]{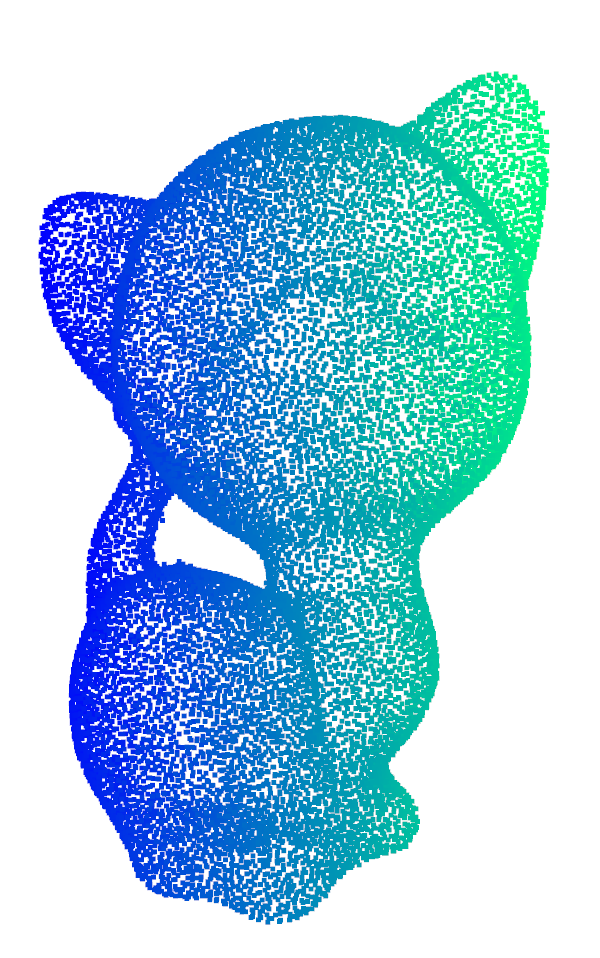}}
\caption{Visual results of the proposed method for upsampling point clouds with various sparsities. Top row shows the input point clouds with (a1) 512, (a2) 1024, (a3) 2048, and (a4) 4096  points, respectively. Bottom row shows the corresponding $4\times$ upsampled point clouds.}
\label{fig:sparse}
\end{figure}

Here we evaluated the robustness of different methods 
to noise. 
Fig.~\ref{fig:noise_quant}  quantitatively compares different methods under 5 levels of Gaussian noise, including 0.5\%, 1.0\%, 1.5\%, 2.0\%, and 2.5\%, where it can be seen that the performance of  all methods decreases 
with the noise level increasing. Nevertheless, the proposed method consistently achieves the best performance under each noise level. Although 3PU-Net performs well for noisy data in terms of the quantitative metrics, it fails to generate uniformly distributed point clouds even for clean input, as illustrated in Fig. \ref{fig:non-uniform}.
Besides, Fig.~\ref{fig:noise_vis} visualizes the upsampled results by the proposed method from various noisy point clouds.
We observe that the results from the noisy inputs are close to those from noise-free inputs, demonstrating the robustness of the proposed method to noise.

Besides, 
the results illustrated in Fig.~\ref{fig:sparse} demonstrate the robustness of the proposed method to data with varying degrees of sparseness.
\begin{figure}[t!]
\centering
         \includegraphics[width=0.4\textwidth]{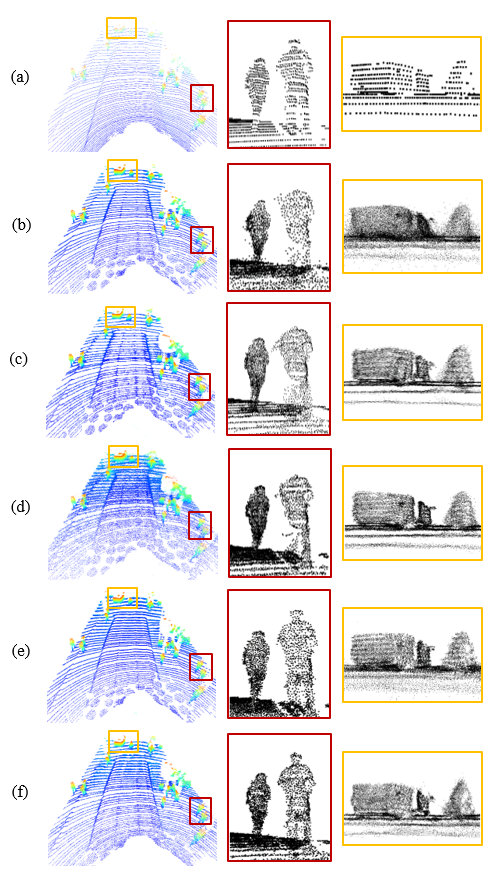}
\caption{Visual comparisons on real-world LiDAR data ($R=4$). 
(a) shows the input point cloud. The upsampled results yielded by (b) 3PU-Net, (c) PU-GAN, (d) PU-GAN-G, (e) PUGeo-Net, and (f) Proposed.}
\label{fig:lidar}
\end{figure}

\subsection{Evaluation on Real-World Data}
We also examined the performance of the proposed method on real-world point cloud data 
i.e.,  one street scene from KITTI \cite{Geiger2013IJRR} captured by LiDAR for autonomous driving, objects with 2048 points from ScanObjectNN~\cite{uy2019revisiting}, and two voxelized full human bodies from 8iVFB~\cite{8ivfb} for immersive communication, which were captured via a typical multi-view technique and voxelization. 
\begin{figure}[b!]
     \centering
    \subfigure[]{\includegraphics[width=0.5in]{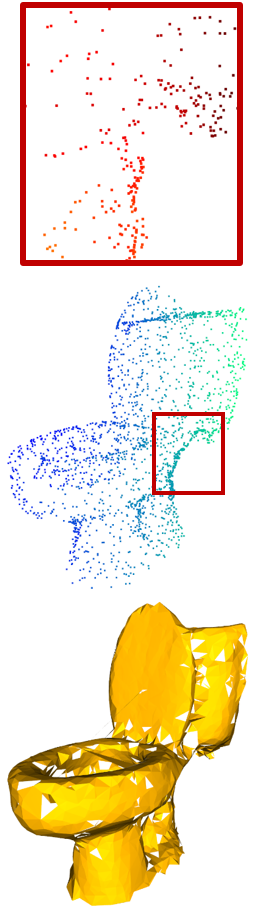}}
    \subfigure[]{\includegraphics[width=0.5in]{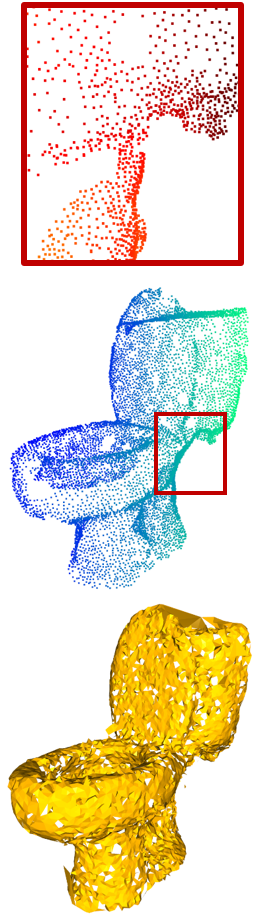}}
        \subfigure[]{\includegraphics[width=0.508in]{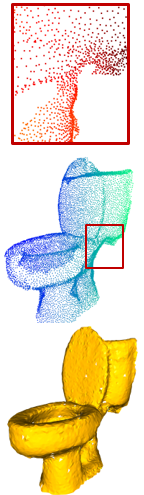}}
    \subfigure[]{\includegraphics[width=0.5in]{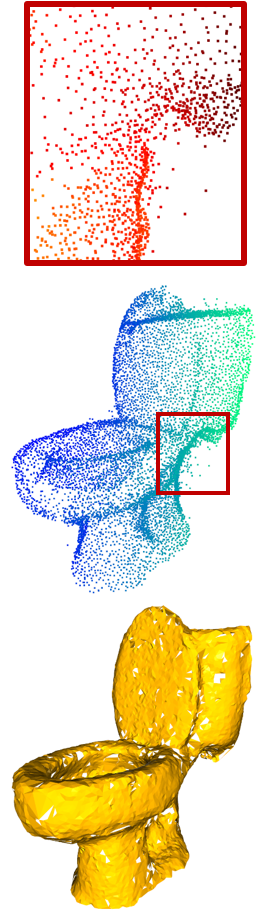}}
    \subfigure[]{\includegraphics[width=0.5in]{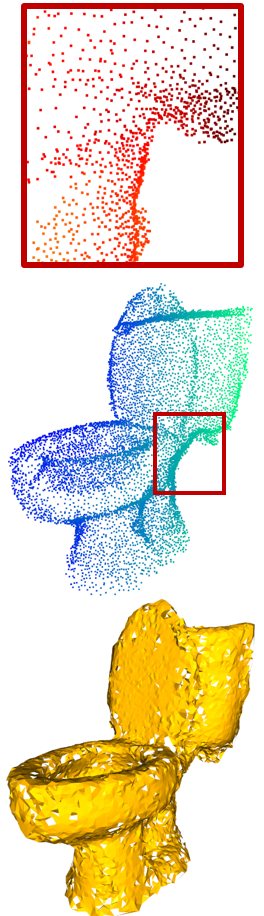}}
    \subfigure[]{\includegraphics[width=0.5in]{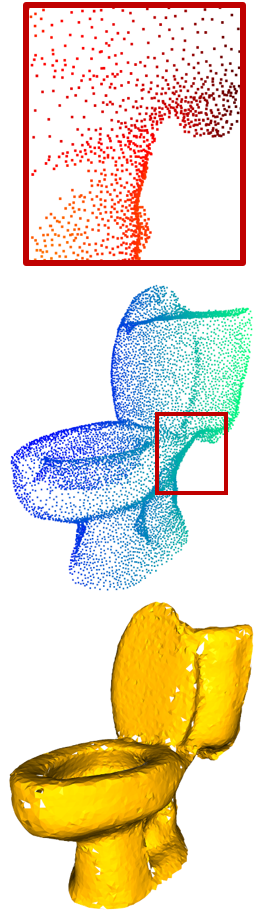}}
\caption{Visual comparisons on ScanObjectNN ($R = 4$). (a) Real-scanned point clouds with 2,048 points.
 The $4\times$ upsampled results by (b) 3PU-Net, (c) PU-GAN, (d) PU-GAN-G, (e) PUGeo-Net, and (f) Proposed. The top row shows the zoomed-in regions indicated by the red frames, and the bottom row shows the reconstructed surfaces from corresponding point clouds via ball-pivoting reconstruction.}
\label{fig:scanobjnn}
\end{figure}
\begin{figure}[b!]
     \centering
    \subfigure[]{\includegraphics[width=0.5in]{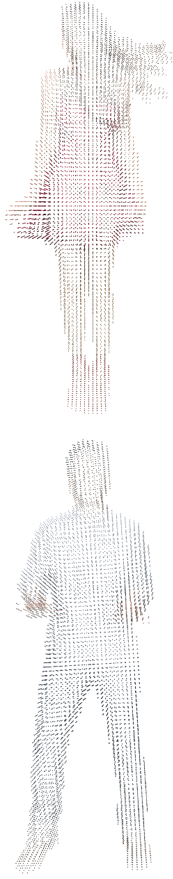}}
    \subfigure[]{\includegraphics[width=0.5in]{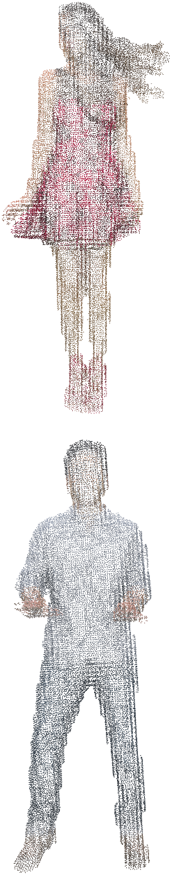}}
    \subfigure[]{\includegraphics[width=0.5in]{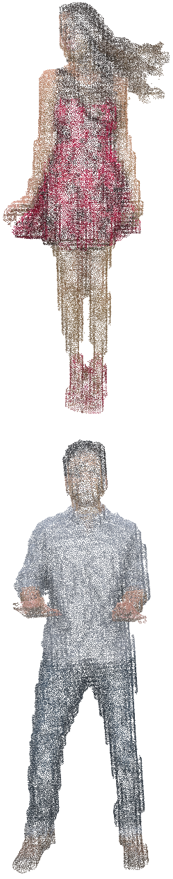}}
    \subfigure[]{\includegraphics[width=0.5in]{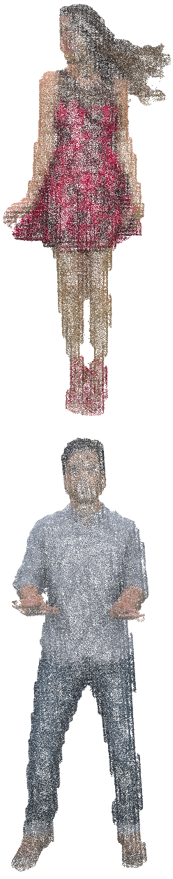}}
    \subfigure[]{\includegraphics[width=0.5in]{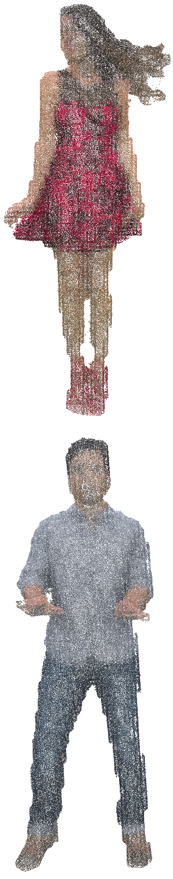}}
\caption{Visual results of our method used for upsampling (a) real-world data 
with (b) $4\times$, (c) $8\times$, (d) $12\times$, and (e) $16\times$. For better visualization, we set the colors of the newly interpolated points as those of the nearest input points.}
\label{fig:scan}
\end{figure}
\begin{table*}[b!]
\centering
\caption{Ablation studies towards the effectiveness of several key modules contained in the proposed method ($R=4$).}
\begin{tabular}{c c c c||c c c c c}\Xhline{5\arrayrulewidth}
  Refinement & $L_{coarse}$ & $L_{pro}$ & $L_{uni}$ & \makebox[3em]{CD } & \makebox[3em]{HD }  &\makebox[3em]{JSD } & \makebox[3.5em]{P2F mean} & \makebox[3.5em]{P2F std} \\\Xhline{2\arrayrulewidth}
 \xmark &  &  &  & 1.757&	4.453&	2.770&	2.251&	2.665\\
  & \xmark &  &  & 1.719&	4.560&	2.225&	2.143&	2.680\\
 &  & \xmark  &  & 1.733&	4.610&	2.216&	2.515&	3.201\\
 &  &  &   \xmark & 1.723&	4.451&	2.337&	2.344&	2.890\\\Xhline{\arrayrulewidth}
 \multicolumn{4}{c||}{Complete Model}    & \bf{1.708}&	\bf{4.315}&	\bf{2.121}&	\bf{2.066}&	\bf{2.566}
 \\\Xhline{2\arrayrulewidth}\end{tabular}
\label{table:ablation} 
\end{table*}

As shown in Fig.~\ref{fig:lidar},  
due to the cost of hardware, the original point cloud by LiDAR suffers from sparsity and non-uniformity issues. 
The upsampled point clouds by different methods show more geometry details of objects in comparison with the original one. 
Moreover, compared with 3PU-Net, PU-GAN, PU-GAN-G, and PUGeo-Net, the proposed method can recover more accurate object shapes while introducing fewer 
outliers, which may be beneficial to downstream applications. 

In Fig.~\ref{fig:scanobjnn}, we also visualized the
reconstructed surfaces using the ball-pivoting algorithm for upsampled ScanObjectNN  with the same hyper-parameters. By examining the visual results, we observe that our results (both the upsampled point clouds and the reconstructed surfaces) have comparable quality to PU-GAN and much higher than the other non-GAN-based methods. 

Fig.~\ref{fig:scan} shows more visual results of real world data upsampled by the proposed method.  
Here we also displayed the associated colors of the point clouds for better visualization purposes. Particularly, the color attributes of newly upsampled points are kept identical to the closest points in the sparse input.
From Fig.~\ref{fig:scan}, it can be seen that the quality of upsampled point clouds gradually improves with the upsampling factor increasing, i.e., more geometry details exhibit.

\subsection{Ablation Study}
To deeply understand the proposed method, we conducted extensive ablation studies. 
As listed in Table~\ref{table:ablation}, 
after removing the refinement module (i.e., the $1^{st}$ row), the quantitative performance gets worse,
compared with the complete model, validating the effectiveness of the refinement module.

\begin{figure}[th]
     \centering
    \subfigure{\includegraphics[width=0.8in]{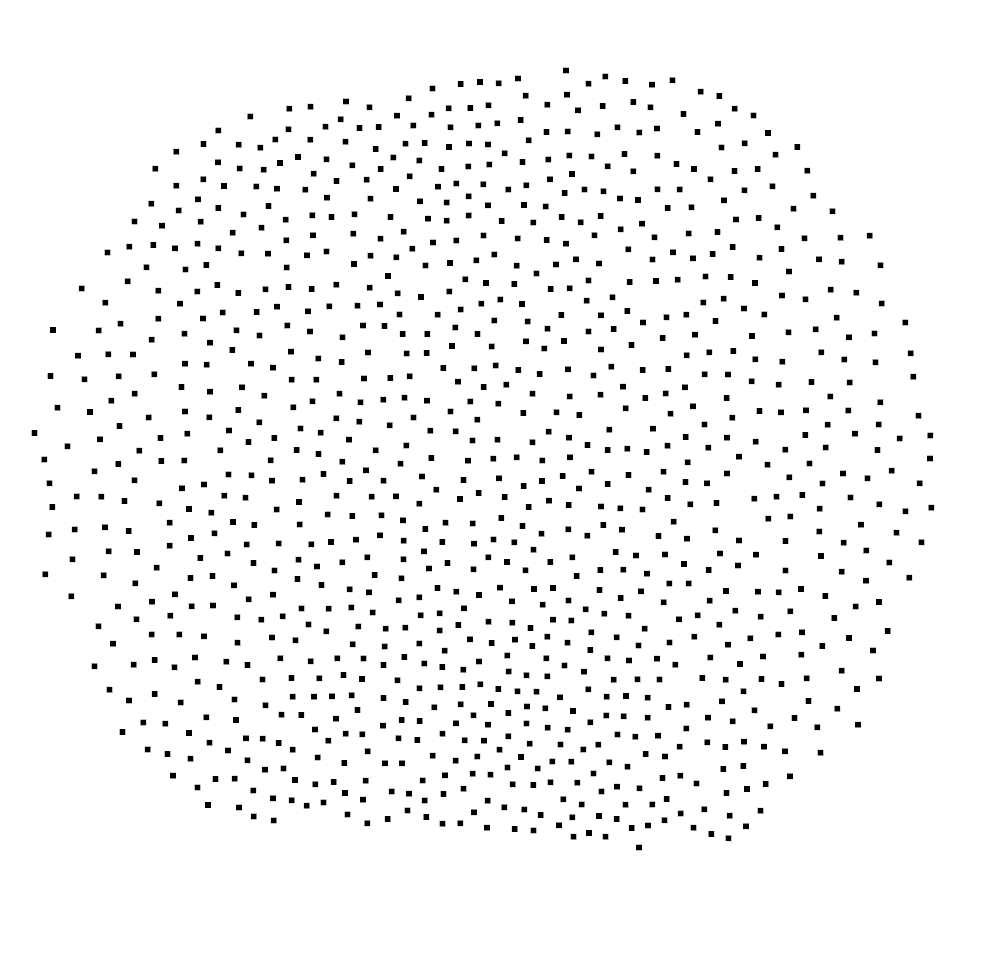}}
    \subfigure{\includegraphics[width=0.8in]{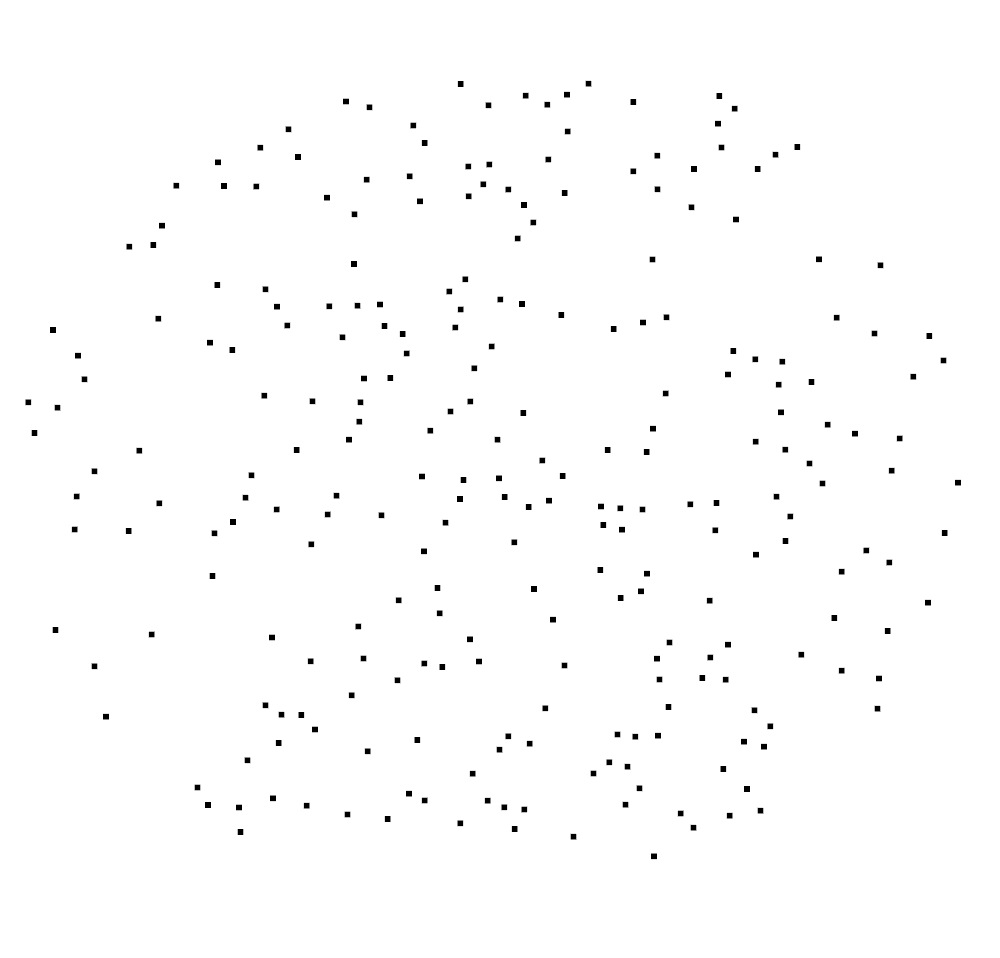}}
    \subfigure{\includegraphics[width=0.8in]{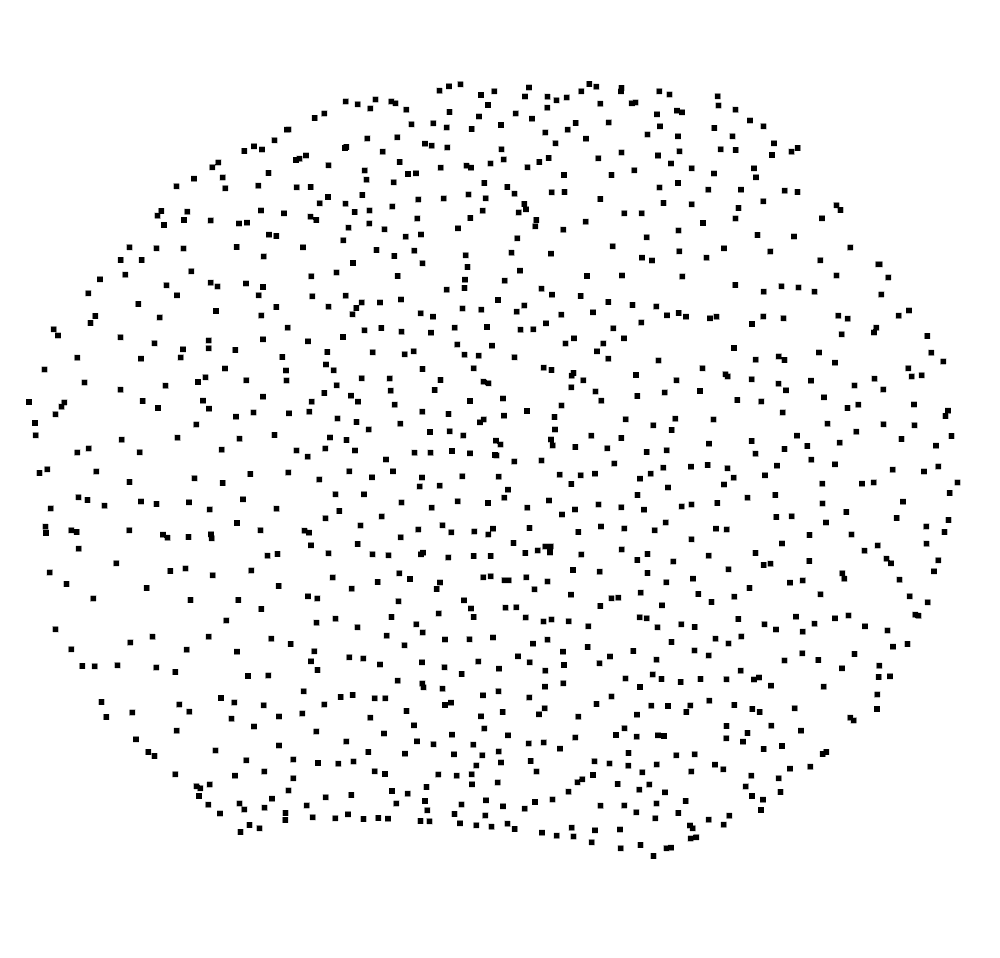}}
    \subfigure{\includegraphics[width=0.8in]{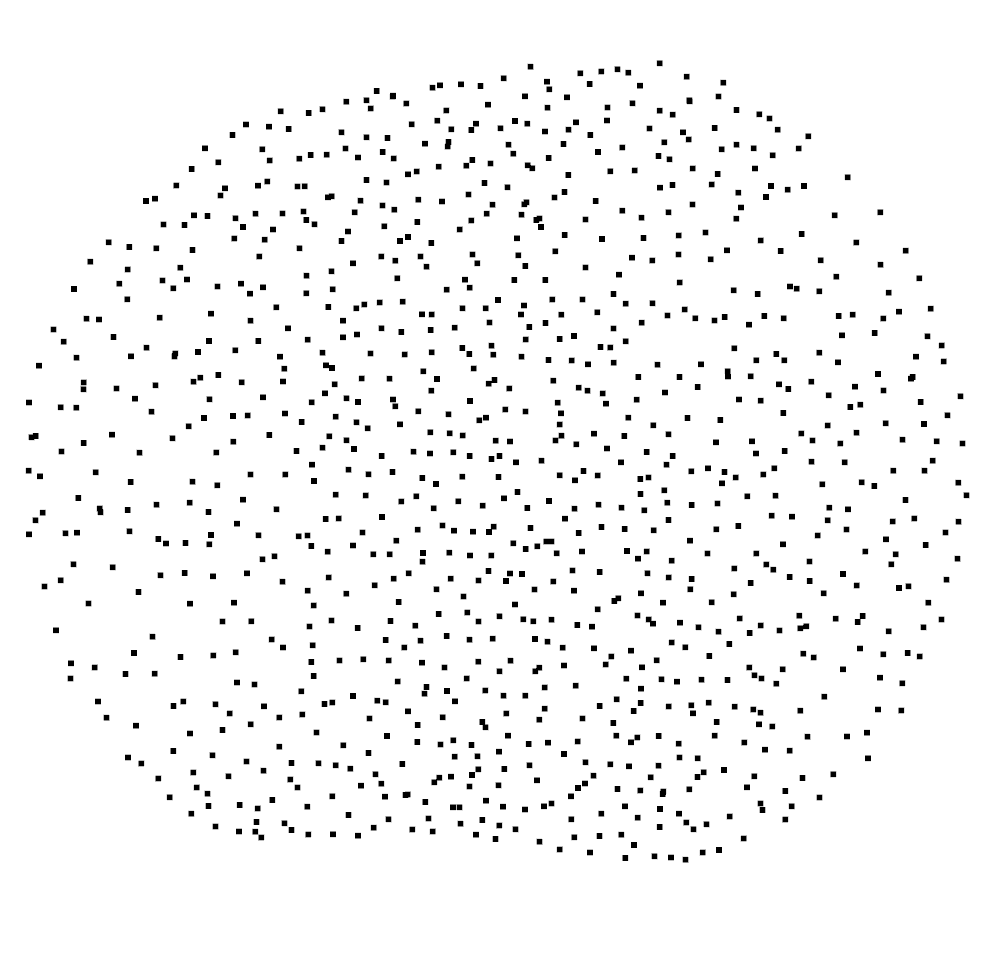}}
 \renewcommand{\thesubfigure}{{\scriptsize (a)}}
    \subfigure[]{\includegraphics[width=0.8in]{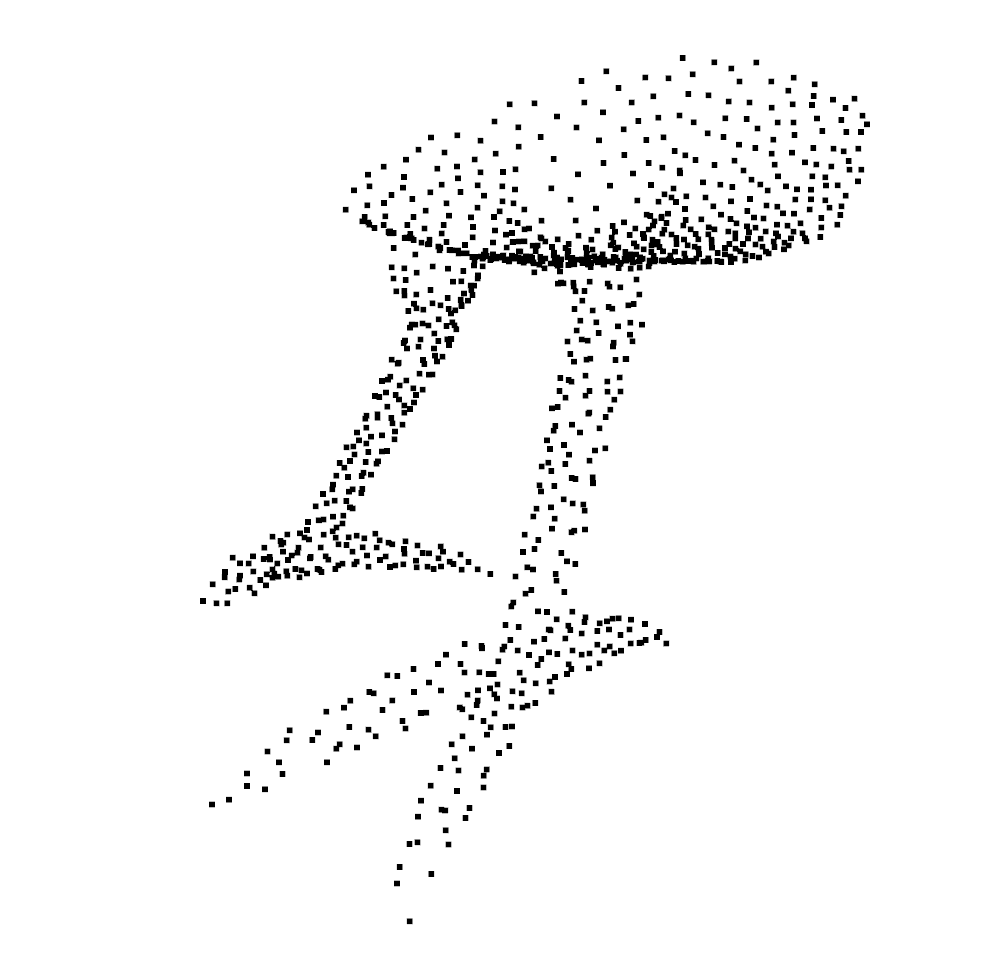}}
     \renewcommand{\thesubfigure}{{\scriptsize (b)}}
    \subfigure[]{\includegraphics[width=0.8in]{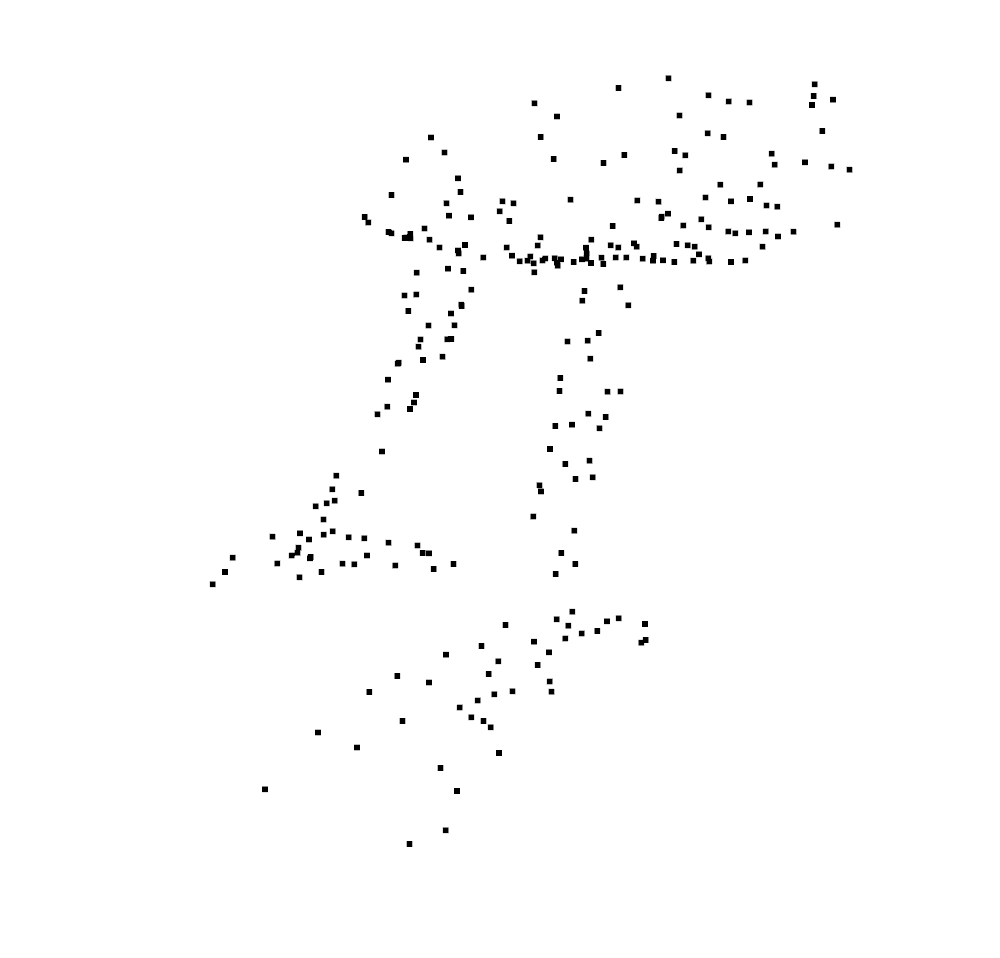}}
     \renewcommand{\thesubfigure}{{\scriptsize (c)}}
    \subfigure[]{\includegraphics[width=0.8in]{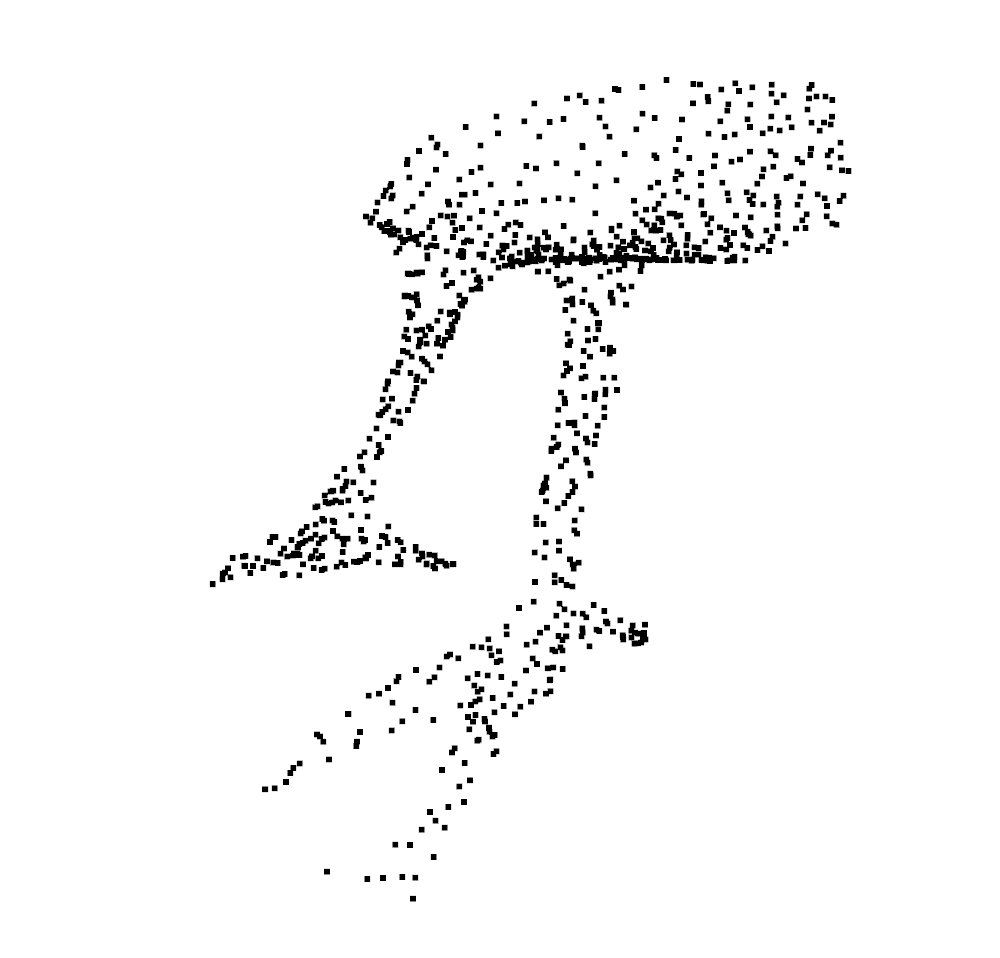}}
 \renewcommand{\thesubfigure}{{\scriptsize (d)}}
    \subfigure[]{\includegraphics[width=0.8in]{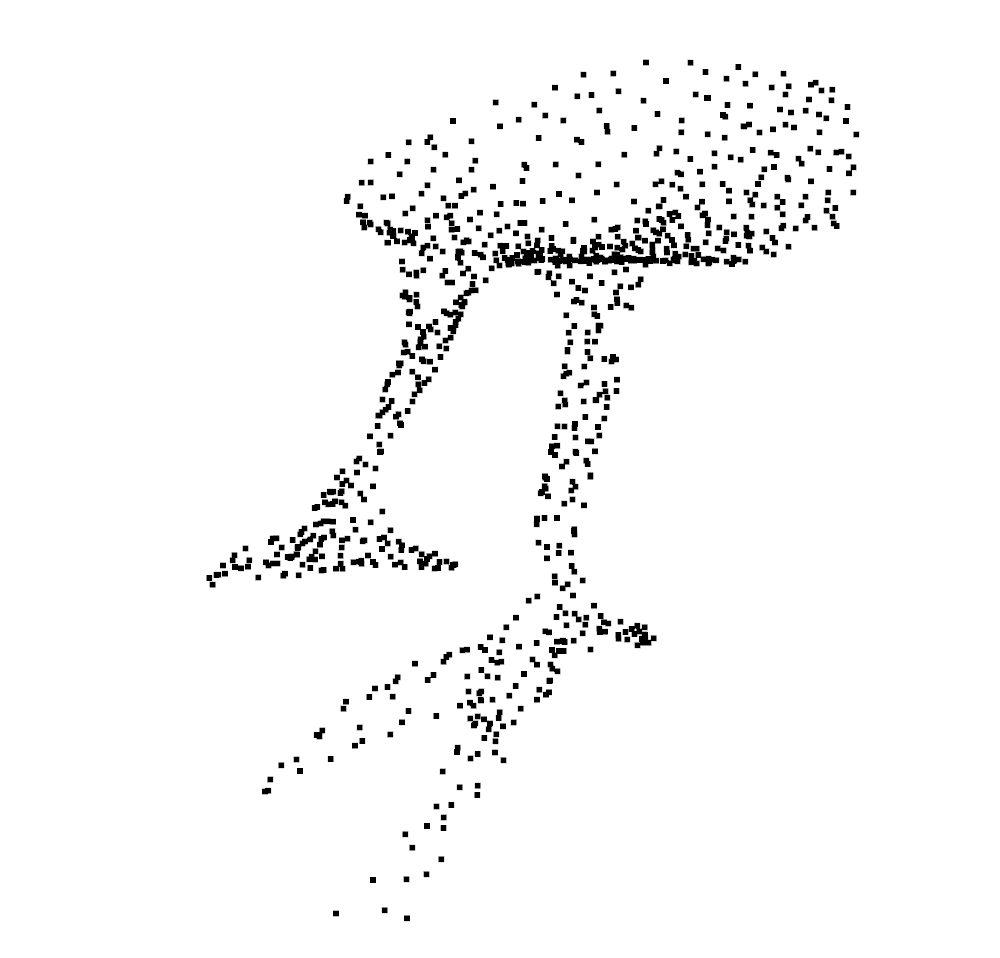}}
\caption{
(a) Non-uniform input point clouds with 256 points; (b)  Ground-truth dense point clouds with 1024 points; (c) and (d) $4\times$ upsampled results by the proposed method without and with the refinement module, respectively. }
\label{fig:patch}
\end{figure}

Besides, Fig.~\ref{fig:patch} illustrates upsampled results by the proposed method with and without the refinement module,  
where it can be seen that the point clouds by the proposed method with the refinement module are more uniform and contain more features. 

The $2^{nd}$, $3^{rd}$, and $4^{th}$ rows of Table~\ref{table:ablation} correspond to the proposed method without $L_{coarse}$, $L_{pro}$, and $L_{uni}$, respectively. By 
observing performance degradation is due to the removal of which component, 
we can conclude that all the three components play their roles.
Particularly, the removal of $L_{pro}$ would cause relatively large degradation for the P2F metrics. Such an  observation is consistent with our analysis in Section~\ref{sec:loss}, as $L_{pro}$ explicitly supervises the projection distance between the upsampled point clouds and the ground-truths.

\section{Conclusion}
\label{sec:con}
We presented a novel end-to-end learning framework for magnification-flexible point cloud upsampling. 
As a neural network built upon an explicit formulation of the upsampling problem using linear approximation, the proposed method is interpretable and compact. It distinguishes itself from the deep learning peers in flexibility since it is able to generate dense point clouds for various upsampling factors after only one-time training. 
Quantitative evaluation on synthetic data shows that the proposed method is more accurate and can produce richer and more meaningful geometric details than state-of-the-art methods. 
We also demonstrated the effectiveness and advantage of the proposed method on non-uniform, noisy point clouds as well as real-world LiDAR data. 

We notice that our performance judging by HD and  NUC for large disk ratios does not exceed the state-of-the-art method. 
A possible improvement is to introduce GAN-based structure, which we will investigate in the near future.
Besides, we will extend our method to simultaneously increase the resolution of geometry and the associated attributes (e.g., colors) of point clouds. Moreover, we will investigate the potential of the proposed method in point cloud compression, which is highly demanded for efficient storage and transmission.

\end{document}